\documentclass{alt2027}  % arXiv; retain the existing template

\usepackage{times}
\usepackage{booktabs}    % table rules and spacing
\usepackage{longtable,array} % multipage notation tables with wrapped columns
\usepackage{microtype}   % better line breaking; reduces overfull boxes
\usepackage{xurl}        % allow long URLs in the bibliography to break
\usepackage{subfiles}    % compile the Lean supplement in the same document
\usepackage{needspace}
\usepackage{etoc}        % separate paper and supplement tables of contents

\jmlrproceedings{}{}
\jmlrvolume{}
\jmlryear{}
\jmlrpages{}
\jmlrworkshop{}
\editor{}
\makeatletter
\let\ps@jmlrtps\ps@plain
\renewcommand{\jmlrpostauthor}{%
  \@endauthor\egroup
  \par\vskip 12pt
}
\makeatother
\usepackage{amsmath,amsfonts,bm,dsfont}

\def\eqref#1{equation~\ref{#1}}
\def\plaineqref#1{\ref{#1}}
\def\1{\mathds{1}}

\def\mD{{\bm{D}}}

\def\mI{{\bm{I}}}

\def\mV{{\bm{V}}}

\def\mOmega{{\bm{\Omega}}}

\DeclareMathAlphabet{\mathsfit}{\encodingdefault}{\sfdefault}{m}{sl}
\SetMathAlphabet{\mathsfit}{bold}{\encodingdefault}{\sfdefault}{bx}{n}

\def\gA{{\mathcal{A}}}
\def\gB{{\mathcal{B}}}
\def\gC{{\mathcal{C}}}

\def\gE{{\mathcal{E}}}
\def\gF{{\mathcal{F}}}

\def\gH{{\mathcal{H}}}
\def\gI{{\mathcal{I}}}
\def\gJ{{\mathcal{J}}}
\def\gK{{\mathcal{K}}}

\def\gO{{\mathcal{O}}}
\def\gP{{\mathcal{P}}}
\def\gQ{{\mathcal{Q}}}
\def\gR{{\mathcal{R}}}
\def\gS{{\mathcal{S}}}
\def\gT{{\mathcal{T}}}

\def\gV{{\mathcal{V}}}
\def\gW{{\mathcal{W}}}

\def\gZ{{\mathcal{Z}}}

\def\sN{{\mathbb{N}}}

\def\sP{{\mathbb{P}}}
\def\sQ{{\mathbb{Q}}}

\newcommand{\E}{\mathbb{E}}

\newcommand{\R}{\mathbb{R}}

\newcommand{\KL}{D_{\mathrm{KL}}}

\makeatletter
\newcommand{\stepby@refs}[1]{%
  \text{eq.}\,%
  \let\stepby@sep\@empty
  \@for\stepby@lab:=#1\do{%
    \stepby@sep\plaineqref{\stepby@lab}\def\stepby@sep{,\,}}%
}
\newcommand{\stepby@render}[2]{\mathrel{\stackrel{\stepby@refs{#1}}{#2}}}
\newcommand{\eqby}[1]{\stepby@render{#1}{=}}
\newcommand{\leby}[1]{\stepby@render{#1}{\leq}}
\newcommand{\geby}[1]{\stepby@render{#1}{\geq}}
\newcommand{\ltby}[1]{\stepby@render{#1}{<}}
\makeatother

\newcommand{\Ber}{\mathrm{Ber}}

\usepackage{listings}
\lstdefinelanguage{lean4}{
  morekeywords={def,noncomputable,theorem,lemma,example,by,fun,let,have,show,
    match,with,structure,instance,variable,variables,open,namespace,end,section,
    if,then,else,do,return,abbrev,class,where,calc,exact,intro,at,this},
  sensitive=true,
  morecomment=[l]{--},
  morecomment=[s]{/-}{-/},
  morestring=[b]",
}

\newcommand{\leanfont}{\ttfamily}

\lstdefinestyle{lean}{
  language=lean4,
  basicstyle=\leanfont\footnotesize,
  keywordstyle=\bfseries,
  commentstyle=\itshape,
  columns=fullflexible,
  keepspaces=true,
  showstringspaces=false,
  extendedchars=true,
  inputencoding=utf8,
  breaklines=true,
  breakatwhitespace=true,
  breakindent=0pt,
  frame=leftline,
  framesep=6pt,
  xleftmargin=8pt,
  aboveskip=2pt,
  belowskip=4pt,
  literate=%
    {ℝ}{{$\mathbb{R}$}}1 {ℕ}{{$\mathbb{N}$}}1 {₀}{{$_0$}}1 {₁}{{$_1$}}1
    {₂}{{$_2$}}1 {ᵢ}{{$_i$}}1 {ⁿ}{{$^n$}}1 {∀}{{$\forall$}}1
    {∃}{{$\exists$}}1 {→}{{$\rightarrow$}}1 {↦}{{$\mapsto$}}1
    {≤}{{$\leq$}}1 {≥}{{$\geq$}}1 {≠}{{$\neq$}}1 {∈}{{$\in$}}1
    {∧}{{$\wedge$}}1 {∨}{{$\vee$}}1 {¬}{{$\neg$}}1 {·}{{$\cdot$}}1
    {λ}{{$\lambda$}}1 {α}{{$\alpha$}}1 {π}{{$\pi$}}1 {∑}{{$\sum$}}1
    {∏}{{$\prod$}}1 {⟨}{{$\langle$}}1 {⟩}{{$\rangle$}}1
    {Ω}{{$\Omega$}}1 {ω}{{$\omega$}}1 {×}{{$\times$}}1,
}

\makeatletter
\newcommand{\leanhead}[1]{%
  \@@par\addvspace{\smallskipamount}%
  \noindent{\footnotesize\textsc{Lean\,4}\quad #1}%
  \@@par\nobreak\vspace{2pt}}
\makeatother
\lstnewenvironment{leancode}[1]{\leanhead{#1}\lstset{style=lean}}{}

\newtheorem{assumption}[theorem]{Assumption}

\makeatletter
\let\lebet@orig@begintheorem\@begintheorem
\renewcommand*{\@begintheorem}[2]{%
  \lebet@orig@begintheorem{#1}{#2}%
  \@ifundefined{lebet@bodyfont@\@currenvir}{}%
    {\csname lebet@bodyfont@\@currenvir\endcsname}%
}
\let\lebet@orig@opargbegintheorem\@opargbegintheorem
\renewcommand*{\@opargbegintheorem}[3]{%
  \lebet@orig@opargbegintheorem{#1}{#2}{#3}%
  \@ifundefined{lebet@bodyfont@\@currenvir}{}%
    {\csname lebet@bodyfont@\@currenvir\endcsname}%
}
\@namedef{lebet@bodyfont@remark}{\normalfont}
\makeatother

\newcounter{squishnum}

\newcommand{\squishenum}{
\begin{list}{\arabic{squishnum}.}
  {
    \usecounter{squishnum}
    \setlength{\itemsep}{2pt}
    \setlength{\parsep}{0pt}
    \setlength{\topsep}{3pt}
    \setlength{\partopsep}{0pt}
    \setlength{\leftmargin}{1.8em}
    \setlength{\labelwidth}{1.2em}
    \setlength{\labelsep}{0.4em}
  }
}

\newcommand{\squishenumend}{
\end{list}
}

\title[Learn then Bet]{Learn then Bet: Learning Interpretable Rules for \\Reliable Agent Intervention}

\altauthor{%
 \Name{You-Lin Chen} \Email{cyoulin@amazon.com}\\
 \addr Amazon
 \AND
 \Name{Xinyi Lisa Chen}\textsuperscript{\textdaggerdbl} \Email{xinyi.lisa.chen@yale.edu}\\
 \addr Yale University
 \AND
 \Name{Pedro Herrero-Vidal} \Email{phvidal@amazon.com}\\
 \addr Amazon
 \AND
 \Name{Xiaojing Yu} \Email{yyxjing@amazon.com}\\
 \addr Amazon
 \AND
 \Name{Bingqing Song} \Email{bingqiso@amazon.com}\\
 \addr Amazon
 \AND
 \Name{Bin Xu} \Email{binxu@amazon.com}\\
 \addr Amazon
}
\renewcommand{\jmlrmaketitlehook}{%
  \begingroup
  \renewcommand{\thefootnote}{\fnsymbol{footnote}}%
  \footnotetext[3]{Work done while an intern at Amazon.}%
  \endgroup
}
\hypersetup{%
  pdftitle={Learn then Bet: Learning Interpretable Rules for Reliable Agent Intervention},
  pdfauthor={You-Lin Chen, Xinyi Lisa Chen, Pedro Herrero-Vidal, Xiaojing Yu, Bingqing Song, Bin Xu}
}

\newcommand{\paperheading}[3]{%
  \texorpdfstring{\hyperref[#1]{#2~\ref*{#1}: #3}}%
    {#2 \getrefnumber{#1}: #3}}
\newcommand{\paperlink}[2]{\texorpdfstring{\hyperref[#1]{#2}}{#2}}

\lstdefinestyle{leansupplement}{
  style=lean,
  numbers=none,
  xleftmargin=8pt,
  literate=%
    {ℝ}{{$\mathbb{R}$}}1 {ℕ}{{$\mathbb{N}$}}1 {ℤ}{{$\mathbb{Z}$}}1
    {𝓝}{{$\mathcal{N}$}}1 {‖}{{$\Vert$}}1
    {𝔼}{{$\mathbb{E}$}}1 {ℱ}{{$\mathcal{F}$}}1 {𝒢}{{$\mathcal{G}$}}1 {Θ}{{$\Theta$}}1
    {Ω}{{$\Omega$}}1 {ω}{{$\omega$}}1 {ι}{{$\iota$}}1
    {α}{{$\alpha$}}1 {λ}{{$\lambda$}}1 {μ}{{$\mu$}}1 {ν}{{$\nu$}}1
    {β}{{$\beta$}}1 {γ}{{$\gamma$}}1 {δ}{{$\delta$}}1 {ρ}{{$\rho$}}1
    {ε}{{$\varepsilon$}}1 {ζ}{{$\zeta$}}1 {η}{{$\eta$}}1
    {θ}{{$\theta$}}1 {κ}{{$\kappa$}}1 {Ξ}{{$\Xi$}}1 {ξ}{{$\xi$}}1
    {«}{{{\fontencoding{T1}\selectfont\guillemotleft}}}1
    {»}{{{\fontencoding{T1}\selectfont\guillemotright}}}1
    {π}{{$\pi$}}1 {σ}{{$\sigma$}}1 {τ}{{$\tau$}}1 {ψ}{{$\psi$}}1 {φ}{{$\varphi$}}1
    {₀}{{$_0$}}1 {₁}{{$_1$}}1 {₂}{{$_2$}}1 {₃}{{$_3$}}1
    {ᵐ}{{$^m$}}1 {ᶜ}{{$^c$}}1 {⁻}{{$^-$}}1 {¹}{{$^1$}}1 {ˢ}{{$^s$}}1
    {∀}{{$\forall$}}1 {∃}{{$\exists$}}1
    {→}{{$\rightarrow$}}1 {←}{{$\leftarrow$}}1 {↦}{{$\mapsto$}}1
    {↔}{{$\leftrightarrow$}}1 {↓}{{$\downarrow$}}1
    {≤}{{$\leq$}}1 {≥}{{$\geq$}}1 {≠}{{$\neq$}}1
    {∈}{{$\in$}}1 {∉}{{$\notin$}}1 {∅}{{$\emptyset$}}1
    {⊆}{{$\subseteq$}}1 {⊂}{{$\subsetneq$}}1
    {∩}{{$\cap$}}1 {∪}{{$\cup$}}1 {⋃}{{$\bigcup$}}1 {⋂}{{$\bigcap$}}1
    {∧}{{$\wedge$}}1 {∨}{{$\vee$}}1 {¬}{{$\neg$}}1
    {×}{{$\times$}}1 {∘}{{$\circ$}}1 {·}{{$\cdot$}}1 {•}{{$\bullet$}}1
    {∑}{{$\sum$}}1 {∏}{{$\prod$}}1 {∫}{{$\int$}}1
    {∂}{{$\partial$}}1 {≃}{{$\simeq$}}1
    {⨆}{{$\bigsqcup$}}1 {⊤}{{$\top$}}1 {⊔}{{$\sqcup$}}1 {⊢}{{$\vdash$}}1 {▸}{{$\blacktriangleright$}}1
    {∞}{{$\infty$}}1
    {⟨}{{$\langle$}}1 {⟩}{{$\rangle$}}1
    {q̂}{{\^{q}}}1 {p̂}{{\^{p}}}1 {Ŝ}{{\^{S}}}1 {X̂}{{\^{X}}}1
    {Û}{{\^{U}}}1 {L̂}{{\^{L}}}1 {τ̂}{{$\hat\tau$}}1
}
\newcommand{\reviewcode}[1]{%
  \par\smallskip
  \Needspace{6\baselineskip}%
  \lstinputlisting[style=leansupplement]{supplement/lean/#1.lean}%
  \par\smallskip
}
\newcommand{\proofcode}[2][]{%
  \Needspace{6\baselineskip}%
  \lstinputlisting[style=leansupplement,emptylines=1,#1]{lean/#2.lean}%
}
\newcommand{\proofpart}[1]{%
  \Needspace{6\baselineskip}%
  \lstinputlisting[style=leansupplement,emptylines=1]{supplement/lean/#1.lean}%
}
\newcommand{\leanname}[1]{\texttt{\detokenize{#1}}}

\newenvironment{reviewunit}{%
  \par\medskip\noindent\begin{minipage}{\linewidth}%
}{\end{minipage}\par\medskip}

\begin{document}

% Fixed spacing confined to the first-page title and author block.
\begingroup
\vspace*{-42pt}
\setlength{\interauthorskip}{6pt}
\setlength{\parskip}{0pt}
\renewcommand{\jmlrpretitle}{\begingroup\centering\Large\bfseries}
\renewcommand{\jmlrposttitle}{\par\endgroup\vskip 16pt}
\pdfbookmark[0]{Learn then Bet}{paper}
\maketitle
\endgroup
\etocdepthtag.toc{paper}

% ===========================================================================
%  Main text.  One file per section; the input order fixes reading order.
%  Comment out an \input while drafting to shorten the compile (cross-
%  references into the skipped file will render as "??" until restored).
% ===========================================================================
% ==========================================================================
% Abstract and keywords.  Input from main.tex.
% ==========================================================================
\begin{abstract}%
Autonomous agents now run for hundreds of steps with little or no human supervision. In practice, practitioners use runtime monitors such as turn limits and keyword filters to decide when a run should be stopped, upgraded, or assisted. 
However, these heuristics may stop runs that would otherwise succeed or allow failing runs to continue.
To address this limitation, we introduce LeBet (Learn then Bet), a statistical learning framework that derives its learning objective from sufficient conditions for valid and powerful intervention. In particular, LeBet learns an interpretable OR-of-AND monitor and accumulates evidence through a fixed betting statistic, intervening when the evidence reaches a prescribed threshold. 
{Under finite within-class exchangeability and leave-one-out stability, we derive finite-sample guarantees for false-positive control (validity) and detection power conditional on the training data, using the same trajectories for learning and calibration without assuming independent and identically distributed trajectories.}
To address the intractable combinatorial optimization problem in our learning formulation, we use stochastic local search that leverages the monitor's Boolean structure to propose rule edits aimed at improving empirical separation.
On SWE-bench Lite with Ministral 3 8B, the learned monitor saves {$34.46\%$} of input-plus-output tokens and increases output precision from $18.33\%$ to {$21.59\%$}. {The empirical false-positive rate is $13.64\%$.
Our results show how to formalize the learning problem over interpretable rule sets with statistical conditions and establish training-conditional guarantees for reliable intervention based on partial histories.
}
\end{abstract}


% ==========================================================================
% \section{Introduction}
% Input from main.tex; see the \input list in the body.
% ==========================================================================
\section{Introduction}

Large language model (LLM) agents are increasingly used to solve complex tasks through multi-step reasoning, tool use, and interaction with external environments. Early agent systems such as ReAct~\citep{yao_react_2023} interleave reasoning and actions, while later work introduces explicit workflow states, memory management, and search over alternative decisions to improve long-horizon execution \citep{wu_stateflow_2024,zhang_memory_2026,wei_long-horizon_2026}. Recent work also studies how to allocate models, tools, and computation across agent pipelines, reflecting the growing cost and complexity of long-running agents \citep{hua_agentopt_2026,gao_more_2025}. As agents operate for longer horizons with less human supervision, failures may arise at intermediate steps and propagate through the remaining trajectory. It is therefore important to have a monitoring system that observes the evolving trajectory and decides when intervention is needed~\citep{gao_more_2025,rabanser_towards_2026,mittapalli_trace_2026,modgil_saturation_2026}.

Existing work on runtime monitoring follows two main directions. 
One line takes a verifier or monitoring output as given and converts it into an online decision rule with statistical guarantees~\citep{sadhuka_e-valuator_2026,schirmer_online_2026,cho_when_2026,prinster_conformal_2026}. 
These guarantees specify conditions under which the supplied signal can support valid intervention and control false alarm risk.
Some of these methods use calibration procedures similar to conformal methods~\citep{angelopoulos_theoretical_2027} to provide finite-sample risk control conditional on the calibration set.
Another line learns a monitor or a model of agent behavior from trajectory data and uses it to predict failures during execution~\citep{wang_probguard_2026,wang_agentspec_2026,xiang_guardagent_2025,chen_shieldagent_2025}. 
The training objective shapes the signal supplied to the decision rule, including its behavior on trajectories where intervention is unnecessary or needed, and the resulting signal can then be calibrated on a separate calibration set to control the false-positive probability.
Together, these directions suggest incorporating the requirements for false-alarm control and detection power into the monitor's learning objective, motivating the following question:
%\begin{quote}
\emph{Can the statistical requirements of reliable intervention guide the learning process?}
%\end{quote}

In this paper, we answer this question through LeBet (Learn then Bet). We formulate agent intervention as a statistical learning problem over complete trajectories while requiring the learned monitor to operate only on the evolving trajectory prefix. Our contributions are as follows:
\squishenum
\item \textbf{Learning formulation and algorithm design (Algorithm~\ref{alg:lebet}).} Rather than first training a general verifier and then calibrating it, we use testing by betting~\citep{ramdas_hypothesis_2025} as an online decision framework to identify statistical conditions sufficient for reliable intervention. We formulate the search for a rule set satisfying these conditions as a learning problem. For interpretability, we restrict the learning class to OR-of-AND rule sets~\citep{xin_exploring_2022,perreault_neural_2026}. Due to the intractable combinatorial optimization problem for learning rule sets~\citep{malioutov_exact_2013}, we adapt FastSRS~\citep{molero-rio_fast_2026} to optimize the empirical separation underlying our intervention guarantees. The key insight is that Boolean identities and  monotonicity guide the choice of rule modifications~\citep{yang_learning_2021}. 

\item \textbf{Training-conditional guarantees (Theorem~\ref{thm:learned-intervention-guarantees}).} Standard sample-splitting approaches use separate training and calibration sets~\citep{vovk_conditional_2013}. Following the approach of \citet{liang_algorithmic_2025}, we extend their stability argument to calibration within intervention classes, replacing independent and identically distributed (i.i.d.) sampling with a finite within-class exchangeable extension and {leave-one-out stability averaged over training-sample exchanges}. We prove finite-sample bounds of a Probably Approximately Correct (PAC) form for the conditional false-positive {and missed-detection probabilities, using the same trajectories for learning and calibration.} These bounds quantify the intuitive tradeoff between separation, instability, and the evidence required for detection.
{In particular, betting allows us to introduce an error budget that controls the trade-off between false intervention and detection.}
The underlying argument may be of independent interest for training-conditional inference under finite exchangeability.

\item \textbf{Empirical illustration (Section~\ref{sec:empirical-illustration}).} On SWE-bench Lite~\citep{jimenez_swe-bench_2024} with a Ministral~3 8B~\citep{liu2026ministral3} coding agent, the learned monitor saves {$34.46\%$} of input-plus-output tokens and increases output precision from $18.33\%$ to {$21.59\%$ with the false-positive rate $13.64\%$.}

\item \textbf{Lean 4 verification (\hyperref[supp:start]{Supplement}).} The formal proofs are machine-checked by Lean 4~\citep{de_moura_lean_2021, de_moura_lean_2015}. We encode the paper's definitions and theorem statements as Challenges and use Comparator~\citep{lean_fro_comparator} to check that Solutions do not weaken the claims or introduce additional assumptions.
\squishenumend

% ==========================================================================
% \section{Preliminaries}
% Input from main.tex; see the \input list in the body.
% ==========================================================================
\section{Preliminaries}
\label{sec:preliminaries}

Throughout, $\sN:=\{0,1,\ldots\}$, $[n]:=\{1,\ldots,n\}$, and $[a]_+:=\max\{a,0\}$. We write $|A|$ for the cardinality of a finite set $A$, and $\lfloor a\rfloor$ and $\lceil a\rceil$ for the floor and ceiling of $a$. We use standard probabilistic notation~\citep{chung2000course}. The symbol $\overset{d}{=}$ denotes equality in distribution. Let $(\Omega,\gF_\infty,\sP)$ be a probability space. We write $\sigma(Z)$ for the $\sigma$-algebra generated by $Z$. A \emph{filtration} ${\gF:=}(\gF_t)_{t\in\sN}$ is an increasing sequence of sub-$\sigma$-algebras of $\gF_\infty$. We write $\E^\sP$ and $\E^\sP[\cdot\mid\gF_t]$ for expectation and conditional expectation under $\sP$, respectively. A process $(X_t)_{t\in\sN}$ is \emph{adapted} if $X_t$ is $\gF_t$-measurable for every $t$. When $X_t$ is binary, $\E^\sP[X_t\mid\gF_{t-1}]=\sP(X_t=1\mid\gF_{t-1})$ for $t\ge1$. A random variable $\tau$ taking values in $\sN\cup\{\infty\}$ is a \emph{stopping time} if $\{\tau\le t\}\in\gF_t$ for every $t$. For $p\in[0,1]$, $\Ber(p)$ denotes the Bernoulli distribution with mean $p$. For $p,q\in(0,1)$, its Kullback–Leibler divergence is $\KL(\Ber(p)\|\Ber(q)):=p\log(p/q)+(1-p)\log((1-p)/(1-q))$. We write $\vee$ and $\wedge$ for logical OR and AND, respectively. An empty disjunction equals zero. We write $\1\{A\}$ for the indicator of an event $A$ and use the convention $\inf\emptyset=\infty$. All logarithms are natural.
{We use the endpoint conventions $0\log(0/q)=0$, $p\log(p/0)=+\infty$ for $p>0$, and $e^{-\infty}=0$.}
For a finite family $(v_i)_{i\in\gI}$ of real values and $k\in[|\gI|]$, $v_{(k:\gI)}$ denotes the $k$th smallest value, counting repetitions.
We write vectors in bold. For a vector $\mV=(v_1,\ldots,v_N)$ and an index vector $\mI=(i_1,\ldots,i_r)$ with distinct entries in $[N]$, write $\mV_{\mI}:=(v_{i_1},\ldots,v_{i_r})$ for the subvector ordered by $\mI$. We use $\gI:=\{i_1,\ldots,i_r\}$ for the corresponding index set, ignoring order.
We summarize additional notation in Appendix~\ref{app:notation}.
Appendix~\ref{app:auxiliary-lemmas} collects auxiliary lemmas used in the proofs.

% ==========================================================================
% \section{Problem Statement: Statistical Modeling of LLM Agents}
% Input from main.tex; see the \input list in the body.
% ==========================================================================
\section{Problem Statement: Statistical Modeling of LLM Agents}
\label{sec:problem-statement}

An LLM agent is specified by a system prompt and a tool set (where skills, MCP servers, and other capabilities are special cases of tools) and interacts with an environment through a ReAct loop \citep{yao_react_2023}. Formally, let $\gZ_t$ denote the space of interaction histories of length $t$, and let
$
\gZ:=\bigcup_{t\ge 0}\gZ_t 
$
denote the space of all finite interaction histories, $\gR_{\mathrm{trace}}$ the space of reasoning traces, $\gA_{\mathrm{tool\_call}}$ the space of tool-call actions, and $\gO$ the space of observations. The agent policy may be viewed as a map $f:\gZ\to\gR_{\mathrm{trace}}\times\gA_{\mathrm{tool\_call}}$, and executing an action produces an outcome in $\gO$. Both the agent policy $f$ and environment outputs may be deterministic. However, determinism of the underlying mechanism does not preclude statistical modeling. This perspective is standard in deterministic computer experiments~\citep{sacks_design_1989, vaart_asymptotic_1998, blackwell_comparison_1951}, where repeated evaluation at the same input produces the same output, but the unknown response function is nevertheless modeled statistically for prediction and uncertainty quantification. We take the same view of agent--environment interactions. In particular, let $(\Omega,\gF_\infty)$ be a measurable sample space whose elements specify single agent--environment interactions. For $\omega\in\Omega$, let $O_0(\omega)$ denote the initial request. At each step $t\ge1$, the agent produces
$
(R_t(\omega),A_t(\omega))
=
f\!\left(Z_{t-1}(\omega)\right),
$
and executing $A_t(\omega)$ in the environment produces the observation $O_t(\omega)$. The resulting trajectory is
\[
Z_t(\omega)
=
\bigl(
O_0(\omega),R_1(\omega),A_1(\omega),O_1(\omega),
\ldots,
R_t(\omega),A_t(\omega),O_t(\omega)
\bigr).
\]
For a fixed $\omega$, the entire agent--environment interaction is fully deterministic. Since $Z_t$ contains the history up to time $t$, the trajectory process $(Z_t)_{t\in\sN}$ is adapted to its natural filtration $(\sigma(Z_t))_{t\in\sN}$. Let $Z_\infty(\omega)$ denote the complete agent--environment trajectory in the absence of intervention. In hindsight, one can assess whether intervention was necessary for the agent to achieve an acceptable outcome. We encode this application-dependent judgment by the binary label $Y:\Omega\to\{0,1\}$, determined by $Z_\infty(\omega)$, with $Y(\omega)=1$ indicating that intervention was needed and $Y(\omega)=0$ indicating that the agent could have proceeded without intervention. All trajectory and label maps are assumed measurable.

We then formalize the finite-sample setting. For $y\in\{0,1\}$, let $\Omega_y:=\{\omega\in\Omega:Y(\omega)=y\}$ and fix integers $n_y\ge1$ and $m_y\ge1$. Put $n:=n_0+n_1$ and $m:=m_0+m_1$. We use a stratified sampling design on the space
$
\mOmega
:=
\Omega_0^{n_0+m_0}\times\Omega_1^{n_1+m_1}\times\Omega,
$
equipped with its product $\sigma$-algebra and a joint probability law $\sP$. A sample is
$
\boldsymbol{\omega}
=
\bigl((\omega_{i,0})_{i=1}^{n_0+m_0},
(\omega_{i,1})_{i=1}^{n_1+m_1},\omega_\star\bigr).
$
Thus $Y(\omega_{i,y})=y$ by construction. The first $n_y$ samples in each class are observed, the remaining $m_y$ form a finite extension, and $\omega_\star$ denotes the next trajectory. Reindex the class samples in a fixed order as $(\omega_1,\ldots,\omega_{n+m+1})$, placing the observed trajectories first and $\omega_\star = \omega_{n+m+1}$ last.
Without loss of generality, $\gI_\star:=[n+m+1]$. Write $\mD:=(\omega_1,\ldots,\omega_{n+m+1})$ for the full sample vector. For an index vector $\mI=(i_1,\ldots,i_r)$ with entries in $\gI_\star$, write $\mD_{\mI}$ for the corresponding indexed sample. For $\gI\subseteq\gI_\star$, let $\Pi_{\gI}$ denote the group of permutations $\pi:\gI_\star\to\gI_\star$ such that $\pi(i)=i$ for every $i\in\gI_\star\setminus\gI$. By abuse of notation, for any permutation $\pi$ of $\gI_\star$, write $\pi(\mD):=(\omega_{\pi(1)},\ldots,\omega_{\pi(n+m+1)})$.
{Write $\mI_{\mathrm{in}}:=(1,\ldots,n)$, so $\mD_{\mathrm{in}}=\mD_{\mI_{\mathrm{in}}}$.}
Write $Z_{t,i}:=Z_t(\omega_i)$ and $Y_i:=Y(\omega_i)$ for $i\in\gI_\star$. Hereafter, we treat the trajectory, label, and associated quantities as random variables under $\sP$ and suppress their sample arguments. When the trajectory index is clear from context, we suppress it; in particular, $Y=Y_\star$ and $Z_t=Z_{t,\star}$ for the next trajectory.
{Write $\gI_{\text{in},y}:=\{i\in[n]:Y_i=y\}$, $\gI_{\text{out},y}:=\{i\in[n+m]\setminus[n]:Y_i=y\}$, and $\gI_{\mathrm{in/out},y}:=\gI_{\text{in},y}\cup\gI_{\text{out},y}$.}
{For any indexed sample $\mD$, write $\mD_y$ for its class-$y$ subvector, preserving order and occurrences. We also write $v_{(k:\mD)}$ for the order statistic over the occurrences in $\mD$.}
For each $y$ with $\sP(Y_\star=y)>0$, write $\sP_y:=\sP(\cdot\mid Y_\star=y)$ and $\E_y:=\E^{\sP_y}$. Rather than assuming i.i.d. trajectories, we impose exchangeability~\citep{chow_probability_1997} under $\sP_y$ within each intervention class, including $\omega_\star$ in class $y$. This assumption rules out systematic distribution shifts between observed and future trajectories within each class while allowing dependence both within and between classes:
\begin{assumption}[Finite exchangeable extension]
\label{assump:finite-exchangeable-extension}
For each $y$ with $\sP(Y_\star=y)>0$, under $\sP_y$,
\[
\begin{gathered}
\pi(\mD)
\overset{d}{=}
\mD,\ \
\forall\,\pi\in
\Pi_{{\gI_{\mathrm{in/out},y}}\cup\{\star\}}
\cup\Pi_{{\gI_{\mathrm{in/out},1-y}}}.
\end{gathered}
\]
\end{assumption}

Finite exchangeable extensions are discussed in \citet[Section~6.2.1]{angelopoulos_theoretical_2027}. Our classwise formulation is related to label-conditional conformal prediction \citep{vovk_conditional_2013,xie_conformal_2025}, where calibration is performed separately within each class.
See Appendix~\ref{app:exchangeability-stability-discussion} for further discussion.

% ==========================================================================
% \section{Motivation: Learn then Bet}
% Input from main.tex; see the \input list in the body.
% ==========================================================================
\section{Motivation: Learn then Bet}
\label{sec:motivation}

Although $Y$ is assigned from the complete trajectory, an intervention decision at time $t$ must depend only on the observed history $Z_t$~\citep{li_judgment_2025,wang_uncertainty-aware_2026}. Suppose we want to detect whether intervention is needed. Consider two adapted binary predicates:
$
\xi_t^{i}: \gZ_t \rightarrow \{0, 1\}, \ \
\xi_{t}^1
=
\1\{O_t\text{ reports a tool failure}\}, \ \
\xi_{t}^2
=
\1\{\exists k<t:A_t=A_k\}.
$
The first predicate detects whether the current tool call fails, while the second detects whether the agent repeats a previous action. A simple monitor $\xi_{t}^1\wedge \xi_{t}^2$ fires when the agent repeats an action and the resulting tool call fails, which may indicate the final status of $Y$. Neither predicate perfectly determines whether a trajectory should be stopped: A trajectory with $Y=0$ may contain an occasional tool failure or repeated action, while a trajectory with $Y=1$ may not exhibit these events at every step. We therefore do not require a monitor $\xi_{t}^1\wedge \xi_{t}^2$ to identify $Y$ deterministically. 
Appendix~\ref{app:monitor-examples} gives additional examples of monitors and possible interventions.
  
We now generalize this simple construction to a combinatorial selection problem. 
\emph{We aim to learn a monitor from the observed trajectories by combining basic adapted predicates.} 
The monitor should detect when intervention is needed while controlling unnecessary intervention.
We set $\mD_{\text{in}}:=\mD_{(1,\ldots,n)}$.
Formally, let $\Xi=\{\xi_t^j:j\in[d]\}$ be a collection of adapted binary predicates fixed before observing $\mD_{\text{in}}$. A nonempty $\gR\subseteq[d]$ defines a conjunction, and a collection $\gS\subseteq 2^{[d]}$ of such sets defines a disjunction:
$
\xi_t^{\gR}:=\bigwedge_{j\in\gR}\xi_t^j, \ \ X_t (\gS): \gZ_t \rightarrow \{0, 1\}, \ \ X_t(\gS):=\bigvee_{\gR\in\gS}\xi_t^{\gR}.
$
Note that we call each $\xi_t^j$ a \emph{predicate} and each conjunction $\xi_t^{\gR}$ a \emph{rule}.
Intuitively, a good monitor $X_t$ should rarely fire when $Y=0$ and fire persistently when $Y=1$. One could formalize this intuition by requiring the conditional probabilities at each step to be small under $Y=0$ and large under $Y=1$. However, these conditional probabilities depend on the entire trajectory history and are difficult to estimate directly from finitely many trajectories. We therefore consider a weaker requirement based on the average conditional mean.
{For a binary monitor $X$ adapted to $\gF$, define
$
\mu_{t,y}^X(Z_t;\gF)
:=
t^{-1}\sum_{u=1}^t\E_y\!\left[
X_u(Z_u)
\,\middle|\,
\gF_{u-1}
\right].
$
We suppress arguments when clear from context.}
We aim to find $\gS$ whose average conditional mean ${\mu_{t,y}^X}$ is small on most trajectories with $Y=0$ and large on most trajectories with $Y=1$. Fix integers $1 < T_0<T_1$ and write $\gW:=\{T_0,\ldots,T_1\}$ for the monitoring window. 
Here $T_0$ sets the earliest intervention time, while $T_1$ specifies the maximum number of steps the agent may execute.
For error budgets $\epsilon_0,\epsilon_1\in[0,1)$, {take $\gF=(\sigma(Z_s))_{s\ge0}$ and} define
\[
\gP(\gS):=\left\{(p_0,p_1)\in[0,1]^2:
\forall y\in\{0,1\},\ 
\sP_y\!\left(\max_{t\in\gW}(-1)^y\bigl({\mu_{t,y}^{X(\gS)}(Z_t)}-p_y\bigr)\le0\right)\ge1-\epsilon_y
\right\}.
\]
For positive integers $K_{\gR}$ and $K_{\gS}$, we seek to maximize the population separation
\begin{equation}
\label{eq:population-separation}
\begin{aligned}
\Delta
:=
\sup_{(p_0,p_1)\in\gP(\gS), \gS\in\gC}
[p_1-p_0]_+,\
\
\gC:=\left\{\gS\subseteq 2^{[d]}:|\gS|\le K_{\gS},\
1\le|\gR|\le K_{\gR}\ \forall\,\gR\in\gS\right\}.
\end{aligned}
\end{equation}
This formulation leads to several challenges.
\squishenum
\item \textbf{How to construct an online intervention rule.} The conditional means in \eqref{eq:population-separation} are not directly observable. In Section~\ref{sec:intervening-by-betting}, we introduce a betting statistic and relate its evidence directly to observed averages.
\item \textbf{How to ensure validity and detection.} An intervention rule should control unnecessary interventions while detecting trajectories that require intervention. Section~\ref{sec:validity-and-detection} establishes sufficient conditions for both guarantees.
\item \textbf{How to generalize to future trajectories.} Even under exchangeability, a large empirical separation does not by itself guarantee the desired error rate on the next trajectory conditional on the observed data. In Section~\ref{sec:learned-intervention-guarantees}, we use stability to establish training-conditional guarantees without splitting the data into training and calibration sets.
\item \textbf{How to solve the combinatorial optimization problem.} Exhaustive search over rule combinations is costly, while random edits can be inefficient. In Section~\ref{sec:rule-learning}, we use the effects of Boolean rule edits on the separation to guide a stochastic local search.
\squishenumend

Together, these ingredients instantiate the \emph{Learn then Bet (LeBet)} principle: we first {learn a monitor} from historical trajectories, and then deploy the frozen monitor through a sequential betting rule for online intervention. Algorithm~\ref{alg:lebet} summarizes our framework.
In Section~\ref{sec:empirical-illustration}, we evaluate LeBet and demonstrate its effectiveness in stopping failed trajectories early and reducing token use.
Section~\ref{sec:related-work} reviews related work, and Section~\ref{sec:discussion} concludes the paper.

\begin{algorithm2e}[t]
\caption{\textsc{LeBet}}
\label{alg:lebet}
\small
\DontPrintSemicolon
\SetKwProg{Stage}{}{}{}
\KwIn{Observed data $\mD_{\text{in}}$, predicate pool $\Xi$, window $\gW$, parameters $\epsilon_0,\epsilon_1,\alpha$.}
\KwOut{Intervention time $\widehat\tau\in\gW\cup\{\infty\}$.}
\Stage{\textnormal{\textbf{Learn}}}{
Obtain $\widehat{\gS}$ by optimizing \eqref{eq:computable-empirical-separation} over $\gC$ in \eqref{eq:population-separation} using Algorithm~\ref{alg:stochastic-local-search}\;
\leIf{$\widehat\Delta^{\epsilon_0,\epsilon_1}_{\mD_{\text{in}}}(\widehat \gS)>{0}$}
{\Return $(\widehat{\gS},\widehat p,\widehat\lambda)$ {with parameters from \eqref{eq:learned-betting-parameters}}}
{\Return $(\emptyset, 1, 1)$}
}
\BlankLine
\Stage{\textnormal{\textbf{Bet}}$(X(\widehat{\gS}),\widehat p,\widehat\lambda)$}{
$E_0(\widehat\lambda,{\widehat p})\gets1$\;
\For{$t=1,\ldots,T_1$}{
Observe $Z_t$ and update $E_t(\widehat\lambda,{\widehat p})$ by \eqref{eq:monitor-evidence}, using $X(\widehat{\gS})$ and $\widehat p$\;
\lIf{$t\in\gW$ and $E_t(\widehat\lambda,{\widehat p})\ge1/\alpha$}{Intervene and \Return $t$}
}
\Return $\infty$\;
}
\end{algorithm2e}

% ==========================================================================
% \section{Intervening by Betting}
% Input from main.tex; see the \input list in the body.
% ==========================================================================
\section{Intervening by Betting}
\label{sec:intervening-by-betting}

For a binary monitor $X$ {adapted to a filtration $\gF$}, we accumulate evidence for intervention through a betting process. Given a reference level $p_0\in[0,1]$, a constant stake $\lambda>0$, and a threshold parameter $\alpha\in(0,1)$, consider the betting process and the corresponding stopping time: set $E_0(\lambda,{p_0})=1$ and
\begin{equation}
\label{eq:monitor-evidence}
E_t(\lambda,{p_0})
:=
\prod_{s=1}^t
\frac{\exp(\lambda X_s)}
{1-p_0+p_0\exp(\lambda)},
\
\tau(\lambda,{p_0})
:=
\inf\left\{
t \in \gW:
E_t(\lambda,{p_0})\ge\frac{1}{\alpha}
\right\}.
\end{equation}
\begingroup
\makeatletter
\def\@currentlabel{\theequation}
\label{eq:intervention-time}
\makeatother
\endgroup
Here $\inf\emptyset=\infty$, so that $\tau(\lambda,{p_0})=\infty$ means no intervention is triggered in the monitoring window.

Each observation $X_t$ updates the evidence relative to $p_0$. The stake $\lambda$ controls the difference between the log-evidence increments for $X_t=1$ and $X_t=0$. Appendix~\ref{app:betting-interpretation} gives the betting interpretation and, for $0<p_0<1$, the equivalent Bernoulli likelihood-ratio form. 
Define the observed averages and their temporal extrema by
\begin{equation}
\label{eq:trajectory-confidence-bounds}
\widehat\mu_t^X(Z_t):=\frac{1}{t}\sum_{s=1}^t X_s(Z_s),
\quad
\widehat U^X(Z_\infty):=\max_{t\in\gW}\widehat\mu_t^X(Z_t),
\quad
\widehat L^X(Z_\infty):=\min_{t\in\gW}\widehat\mu_t^X(Z_t).
\end{equation}
{The following proposition gives calibration and mean-estimation bounds.}

\begin{proposition}
\label{prop:pathwise-calibration}
{Put $v(p):=1/4-[p-1/2]_+^2$, ${a_t}(\alpha):=\sqrt{2\log(1/\alpha)/t}$.} For $p_0\in[0,1]$, $\lambda>0$,
\squishenum
\csname @beginparpenalty\endcsname=10000\relax
\item[\textnormal{(1)}] For $\alpha\in(0,1)$, the stopping time satisfies $\widehat U^X\le p_0
\ \Longrightarrow\ \tau(\lambda,{p_0})=\infty$ and
\begin{equation}
\label{eq:score-evidence-calibration}
\widehat L^X\ge p_0+{(v(p_0)\lambda+\lambda^{-1}a_{T_1}(\alpha)^2)/2}
\ \Longrightarrow\ \tau(\lambda,{p_0})\le T_1.
\end{equation}
\item[{\textnormal{(2)}}] {With $r_t(\beta):=\min\{(1+T_0/t)a_{T_0}(\beta/2)/2,a_t(\beta/(2|\gW|))\}/2$, for every $\gF_0$-measurable $\beta\in(0,1)$ and either fixed sign,
$
\sP_y\!\left(
\exists t\in\gW:\pm(\widehat\mu_t^X-\mu_{t,y}^X)>r_t(\beta)
\,\middle|\,\gF_0\right)
\le\beta.
$
\par
}
\squishenumend
\end{proposition}
\begin{proof}
See Appendix~\ref{app:proof-pathwise-calibration}.
\end{proof}
{For $p_0<1$, \eqref{eq:score-evidence-calibration} is minimized by $\lambda=a_{T_1}(\alpha)/\sqrt{v(p_0)}$. Section~\ref{sec:learned-intervention-guarantees} uses this stake. Including the fitting sample in $\gF_0$ gives conditional-mean bounds for trajectories excluded from fitting.}

% ==========================================================================
% \section{Validity and Detection}
% Input from main.tex; see the \input list in the body.
% ==========================================================================
\section{Validity and Detection}
\label{sec:validity-and-detection}

We extend guarantees from testing by betting and anytime-valid inference~\citep{ramdas_hypothesis_2025,ramdas_game-theoretic_2023,waudby-smith_estimating_2024,howard_time-uniform_2021} to allow imperfect calibration of the betting reference. Bounds on the monitor's average conditional means yield false-intervention and detection guarantees.
{Put $\psi_\lambda(u):=\log(1-u+ue^\lambda)$. For $\lambda>0$, $p,q\in[0,1]$, and $\alpha\in(0,1)$, write $e_0=e_0(p;\lambda,q,\alpha)$ and $e_{1,t}=e_{1,t}(p;\lambda,q,\alpha)$, where
\begin{equation}
\label{eq:mean-false-intervention}
\begin{aligned}
e_0&:=\alpha\exp\!\left(T_1[\psi_\lambda(p)-\psi_\lambda(q)]_+-T_0[\psi_\lambda(q)-\psi_\lambda(p)]_+\right),\\
e_{1,t}&:=\exp\!\left(-t\,\KL\!\left(\Ber\!\left(\min\{p,\lambda^{-1}\{\psi_\lambda(q)+a_t(\alpha)^2/2\}\}\right)\,\middle\|\,\Ber(p)\right)\right).
\end{aligned}
\end{equation}\par
\begingroup
\makeatletter
\def\@currentlabel{\theequation}
\label{eq:detection-condition}
\makeatother
\endgroup
}

\begin{proposition}
\label{prop:statistical-guarantees}
\label{assump:avg-validity}% Merged into part (1).
\label{assump:avg-separation}% Merged into part (2).
{For $\lambda>0$, $p_0,p_1,q_0\in[0,1]$, $t\in\gW$, and $\epsilon_0,\epsilon_1\in[0,1)$, the following hold:}
\squishenum
\item[\textnormal{(1)}] {If
$\sP_0(\max_{t\in\gW}\mu_{t,0}^X\le p_0)\ge1-\epsilon_0$,
then $\sP_0(\tau(\lambda,q_0)<\infty)\le\epsilon_0+e_0(p_0;\lambda,q_0,\alpha)$.
}
\item[\textnormal{(2)}] {If
$\sP_1(\min_{t\in\gW}\mu_{t,1}^X\ge p_1)\ge1-\epsilon_1$,
then $\sP_1(\tau(\lambda,q_0)>t)\le\epsilon_1+e_{1,t}(p_1;\lambda,q_0,\alpha)$.
}
\squishenumend
\end{proposition}
\begin{proof}
See Appendix~\ref{app:proof-statistical-guarantees}.
\end{proof}

Part~(1) of Proposition~\ref{prop:statistical-guarantees} accounts for mismatch between $p_0$ and the betting reference $q_0$, with $\alpha$ controlling the betting contribution. {When $q_0=p_0$, it recovers the bound $\alpha+\epsilon_0$. For any $\gamma\in(0,1)$, part~(2) gives detection probability at least $1-\epsilon_1-\gamma$ whenever $e_{1,t}(p_1;\lambda,q_0,\alpha)\le\gamma$. For fixed $\lambda$, increasing $q_0$ raises the intervention threshold, weakening the detection bound in part~(2). For $0<q_0<p_1<1$, Appendix~\ref{app:betting-interpretation}, with reference $q_0$, gives $\max_{\lambda>0}\{\lambda p_1-\psi_\lambda(q_0)\}=\KL(\Ber(p_1)\|\Ber(q_0))$, attained at $\lambda=\log\{p_1(1-q_0)/[q_0(1-p_1)]\}$.}

\section{Learned Intervention Guarantees}
\label{sec:learned-intervention-guarantees}

We now learn a rule set from $\mD_{\text{in}}$ and establish intervention guarantees on the next trajectory $\omega_\star$ conditional on this sample. To distinguish the observed trajectories, we restore the trajectory index $i$ as a subscript and otherwise keep the preceding notation. For example,
$
X_{t,i}=X_t(Z_{t,i}),
\widehat\mu_{t,i}^X=\widehat\mu_t^X(Z_{t,i}),
\widehat U_i^X=\widehat U^X(Z_{\infty,i}),
\widehat L_i^X=\widehat L^X(Z_{\infty,i}).
$
The same convention applies to the predicates. In this section, write $\widehat U_{(k)}^X:=\widehat U_{(k:\gI_{\mathrm{in},0})}^X$ and $\widehat L_{(k)}^X:=\widehat L_{(k:\gI_{\mathrm{in},1})}^X$, suppressing the training index sets. For a fixed monitor $X$, $\widehat U_i^X$ and $\widehat L_i^X$ depend only on trajectory $i$, while their order statistics depend on the corresponding class samples. Put $k_y:=n_y-\lfloor n_y\epsilon_y\rfloor$. Replacing conditional means by observed averages and {class-conditional probabilities} by empirical frequencies in \eqref{eq:population-separation} gives the empirical separation
\begin{equation}
\label{eq:computable-empirical-separation}
\widehat\Delta^{\epsilon_0,\epsilon_1}_{\mD_{\text{in}}}(\gS)
:=
\left[
\widehat L_{(n_1-k_1+1)}^{X(\gS)}
-
\widehat U_{(k_0)}^{X(\gS)}
\right]_+.
\end{equation}
When this value is positive, the two order statistics attain the largest gap for which at most $\lfloor n_0\epsilon_0\rfloor$ values of $\widehat U_i^{X(\gS)}$ in class~0 exceed $p_0$ and at most $\lfloor n_1\epsilon_1\rfloor$ values of $\widehat L_i^{X(\gS)}$ in class~1 fall below $p_1$. The positive part assigns zero when no positive gap is feasible. Let $\widehat{\gS}(\mD)\in\gC$ denote the rule set obtained by optimizing $\max_{\gS\in\gC}\widehat\Delta^{\epsilon_0,\epsilon_1}_{\mD}(\gS)$ with a fixed algorithm. Note that we do not require $\widehat{\gS}(\mD)$ to attain the maximum.
We compare training samples with the fixed class sizes $n_0,n_1$:
\[
\mathfrak D_{\text{in/out}}
:=
\left\{
(\omega_{i_1},\ldots,\omega_{i_n}):
1\le i_1<\cdots<i_n\le n+m,\
\sum_{r=1}^n\1\{Y_{i_r}=y\}=n_y,\ y\in\{0,1\}
\right\}.
\]
Write $\widehat X(\mD):=X(\widehat{\gS}(\mD))$.
{The key technical observation is as follows. Proposition~\ref{prop:pathwise-calibration} bounds average conditional means using observed averages.
However, the full training sample already contains the complete trajectory being evaluated.
Thus, for $X=\widehat X(\mD_{\mathrm{in}})$ and $i\in\gI_{\mathrm{in},y}$,
$t^{-1}\sum_{s=1}^t\E_y[X_{s,i}\mid\mD_{\mathrm{in}},Z_{s-1,i}]=\widehat\mu_{t,i}^X$.
We therefore use leave-one-out fits to keep the evaluated trajectory out of the conditioning sample.
Let $\mD^{-j}$ delete occurrence $j$ from $\mD$ if present, and equal $\mD$ otherwise.
Use the same learning procedure with the remaining class sizes, returning $\emptyset$ if either class is absent.
Write $\widehat\mu_{t,i}(\mD):=\widehat\mu_{t,i}^{\widehat X(\mD)}$ and $\gF_i(\mD):=(\sigma(\mD,Z_{u,i}))_{u\ge0}$.
For nonnegative $r=(r_t)_{t\in\gW}$, define
\begin{equation}
\label{eq:leave-one-out-conditional-means}
\begin{aligned}
s_{y,i,j}(\mD;\mI,r)&:=\max_{t\in\gW}[(-1)^y
\mu_{t,y}^{\widehat X(\mD_{\mI}^{-j})}(Z_{t,i};\gF_i(\mD_{\mI}^{-j}))+r_t],
\\
\widehat s_{y,i,j}(\mD;\mI,r)&:=\max_{t\in\gW}[(-1)^y\widehat\mu_{t,i}(\mD_{\mI}^{-j})+r_t].
\end{aligned}
\end{equation}
}
Next, we impose leave-one-out stability on the conditional-mean extrema in \eqref{eq:leave-one-out-conditional-means}, following the stability arguments of \citet[Sections~4.2--4.3]{liang_algorithmic_2025}. 
Stability transfers the conditional-mean bounds to the next trajectory, whose observed averages may still cross the reference; betting controls the resulting risk of false intervention.
See Appendix~\ref{app:exchangeability-stability-discussion} for further discussion.
{For integers $k\ge1$ and $h\ge0$, let
\begin{equation}
\label{def:sampling-laws}
\sQ_{k,h}:=\operatorname{Uniform}\bigl(\{K\subseteq[k+h]:|K|=k\}\bigr).
\end{equation}
Let $\sQ_\pi$ be the uniform law on the group generated by
$\Pi_{\gI_{\mathrm{in/out},0}}\cup\Pi_{\gI_{\mathrm{in/out},1}}$.
For $\mD\sim\sP_y$ and an independent $\pi\sim\sQ_\pi$, write
$\sP_{y,\pi}:=\sP_y\otimes\sQ_\pi$ for their joint law.
}

\begin{assumption}[Leave-one-out stability]
\label{assump:score-m-stability}
Write {$S_{y,i,j}(\mD;\mI):=s_{y,i,j}(\mD;\mI,0)$}. {Let $\pi\sim\sQ_\pi$ be independent of $\mD$.
There exist $\theta_0,\theta_1,\kappa_m\in[0,1)$ such that
$\E^{\sQ_\pi}\sP((\gE_{\pi(\mI_{\mathrm{in}}),\mI_{\mathrm{in}}}^m)^c)\le\kappa_m$, where, for $\mD_{\mI},\mD_{\mI'}\in\mathfrak D_{\mathrm{in/out}}$,}
{\begin{equation}
\label{eq:score-swap-stability}
{\gE_{\mI,\mI'}^m}:=
\bigcap_{y\in\{0,1\}}
\left\{
\frac{1}{n_y+m_y}\sum_{i\in{\gI_{\mathrm{in/out},y}}}
\1\!\left\{\left|
{S_{y,i,i}(\mD;\mI)-S_{y,i,i}(\mD;\mI')}
\right|>{B_m}\right\}\le\theta_y
\right\}.
\end{equation}
}
\end{assumption}

{Write $\widehat\Delta(\mD):=\widehat\Delta^{\epsilon_0,\epsilon_1}_{\mD}(\widehat{\gS}(\mD))$. For each $\mD\in\mathfrak D_{\mathrm{in/out}}$, suppress the argument $\mD$ in the betting parameters.
If $\widehat\Delta(\mD)>0$, set
\begin{equation}
\label{eq:learned-betting-parameters}
\widehat p_0:=\widehat U_{(k_0:\mD_0)}^{\widehat X(\mD)},
\quad
\widehat p_1:=\widehat L_{(n_1-k_1+1:\mD_1)}^{\widehat X(\mD)},
\quad
\widehat p:=(\widehat p_0+\widehat p_1)/2,
\quad
\widehat\lambda:=a_{T_1}(\alpha)/\sqrt{v(\widehat p)}.
\end{equation}
\begingroup
\makeatletter
\def\@currentlabel{\theequation}
\label{eq:learned-empirical-levels}
\label{eq:learned-mean-levels}
\makeatother
\endgroup
Otherwise, set $(\widehat p,\widehat\lambda):=(1,1)$.
}
The following theorem gives the intervention guarantees for LeBet.
\begin{theorem}[Learned intervention guarantees]
\label{thm:learned-intervention-guarantees}
\label{cor:learned-conditional-means}% Merged into this theorem.
\label{def:rank-error-bound}% Definition incorporated into this theorem.
Suppose Assumptions~\ref{assump:finite-exchangeable-extension} and~\ref{assump:score-m-stability} hold. Assume (1) $\sP(Y_\star=y)>0$ for each $y$, (2) the selector $\widehat{\gS}$ is measurable and invariant under permutations of the training sample.
{Put $b_y:=\lfloor(n_y+m_y)\theta_y\rfloor$ and let $\gK_y\sim\sQ_{n_y,m_y}$ as in \eqref{def:sampling-laws}.
With $J_{\gK}^a$ from Lemma~\ref{lem:uniform-subset-rank}, set $W_y:=\min\{1,(J_{\gK_y}^{n_y-k_y+b_y}+b_y)/m_y\}$ if $b_y<k_y$, and $W_y:=1$ otherwise.
For $\alpha,\beta\in(0,1)$, define
\begin{equation}
\label{eq:total-statistical-errors}
\delta_y(\beta):=\min_{0\le u\le1}
\left\{u+\frac{\E^{\sQ_{n_y,m_y}}[(W_y-u)_+]
+(1-u)\kappa_m/\sP(Y_\star=y)}{\beta}\right\}.
\end{equation}
}
For {$\mI=(i_1,\ldots,i_n)$ with $\mD_{\mI}\in\mathfrak D_{\text{in/out}}$}, and
$i\in{\gI_{\mathrm{in/out},y}}\cup\{\star\}$, define
{\begin{equation}
\label{eq:learned-detection-condition}
\begin{aligned}
\gV_{y,i}(\mD;\mI)&:=
\left\{S_{y,i,i}(\mD;\mI)>
\left(S_{y,i,i}(\mD;\mI)\right)_{(k_y:(\mD_{\mI})_y)}+2{B_m}\right\},
\\
C_y(\mD;\mI)&:=
\sum_{i\in\gI_{\mathrm{in/out},y}\setminus\gI}
\1\{\gV_{y,i}(\mD;\mI)\}.
\end{aligned}
\end{equation}
\begingroup
\makeatletter
\def\@currentlabel{\theequation}
\label{eq:symmetrized-future-error}
\makeatother
\endgroup
Here $C_y(\mD;\mI)$ counts the class-$y$ errors outside $\mD_{\mI}$.
}
These definitions are unchanged by reordering {$\mI$}.
{Write $r(\beta):=(r_t(\beta))_{t\in\gW}$ using Proposition~\ref{prop:pathwise-calibration}(2), and put
\begin{equation}
\label{eq:learned-predictable-levels}
\widehat q_y:=\min\left\{1-y,
\left(\widehat s_{y,i,i}\left(\mD;\mI_{\mathrm{in}},r\left(\frac{\beta}{2n_y}\right)\right)\right)_{(k_y:\mD_{\mathrm{in},y})}
+2{B_m}\right\}.
\end{equation}
\begingroup
\makeatletter
\def\@currentlabel{\theequation}
\label{eq:leave-one-out-radius}% Radius substituted into the confidence bounds.
\makeatother
\endgroup
}
Let {$\widehat\tau:=\tau(\widehat\lambda,\widehat p)$} in \eqref{eq:intervention-time} for $\widehat X(\mD_{\mathrm{in}})$. Then, for each $y$ and $s>0$:
\squishenum
\item[\textnormal{(1)}]
{$\sP_y(\gV_{y,\star}(\mD;\mI_{\mathrm{in}})\mid\mD_{\mathrm{in}})
=\E_y[C_y(\mD;\mI_{\mathrm{in}})/m_y
\mid\mD_{\mathrm{in}}]$} $\sP_y$-almost surely.
\item[\textnormal{(2)}]
{$\sP_y(C_y(\mD;\mI_{\mathrm{in}})/m_y
>\epsilon_y+\theta_y\frac{(n_y+m_y)^2}{n_ym_y}+s)
\le e^{-2n_ym_ys^2/(n_y+m_y+2)}+\kappa_m/\sP(Y_\star=y)$.}
\item[\textnormal{(3)}]
{$
\sP_y\!\left(
\sP_y(\gV_{y,\star}(\mD;\mI_{\mathrm{in}})\mid\mD_{\mathrm{in}})>{\delta_y(\beta)}
\right)\le\beta.
$}
\item[{\textnormal{(4)}}] {$\sP_0\big(
\sP_0(\widehat\tau<\infty\mid\mD_{\mathrm{in}})
>\delta_0(\beta/2)+e_0(\widehat q_0;\widehat\lambda,\widehat p,\alpha)
\big)\le\beta.$
}
\item[{\textnormal{(5)}}] {$\sP_1\big(
\exists t\in\gW:\sP_1(\widehat\tau>t\mid\mD_{\mathrm{in}})
>\delta_1(\beta/2)+e_{1,t}(-\widehat q_1;\widehat\lambda,\widehat p,\alpha)
\big)\le\beta.$
}
\squishenumend
\end{theorem}

\begin{proof}
See Appendix~\ref{app:proof-learned-intervention-guarantees}.
\end{proof}
{Parts (4) and (5) give finite-sample PAC bounds on next-trajectory errors conditional on the training sample.
With the training quantities fixed and the default stake, the probability bounds, truncated at one, are nondecreasing in $\alpha$ for false intervention and nonincreasing for missed detection at each fixed $t$.
As $\alpha\downarrow0$, $e_0\to0$ and $e_{1,t}\to1$, so the false-intervention bound approaches $\delta_0(\beta/2)$ while the detection guarantee becomes vacuous.
For example, if $\widehat q_0\le\widehat p$ and $\delta_0(\beta/2)=0.02$, choosing $\alpha=0.01$ or $0.05$ gives false-intervention bounds at most $0.03$ or $0.07$, respectively; Appendix~\ref{app:statistical-bound-illustrations} gives the proof and illustrations.
}
{When $m_y=1$ and $\theta_y=\kappa_m=0$, \eqref{eq:total-statistical-errors} reduces to
$\delta_y(\beta)=\min\{1,(\lfloor n_y\epsilon_y\rfloor+1)/(\beta(n_y+1))\}$.
In general, \eqref{eq:general-coverage-threshold} gives
\[
\delta_y(\beta)\le
\epsilon_y+\theta_y\frac{(n_y+m_y)^2}{n_ym_y}
+\frac{\kappa_m}{\beta\sP(Y_\star=y)}
+\sqrt{\frac{n_y+m_y+2}{2n_ym_y}\left(1+\log\frac1\beta\right)}.
\]}
Theorem~\ref{thm:learned-intervention-guarantees} allows changes in the selected rule set under training-sample exchanges, provided that the stated stability bounds for {$S_{y,i,i}(\mD;\mI)$} hold. For a measurable, permutation-invariant objective over a finite rule class, a fixed, data-independent tie-breaking rule makes the optimizer measurable and permutation-invariant~\citep{liang_assumption-free_2025}.
{The guarantee has no explicit dependence on $|\gC|$; selection enters through ${B_m}$, $\theta_y$, and $\kappa_m$, which may depend on the learning class and algorithm. A larger empirical separation need not improve the guarantee if instability also increases. Uniform bounds over a finite rule class may provide an alternative proof route.}

Our proof follows the techniques and stability arguments of \citet{liang_algorithmic_2025} but removes the i.i.d. assumption. 
The key observation for removing the i.i.d. assumption is that, within each class and conditional on the training data, every extension error indicator has the same mean as the next trajectory's error indicator; their average therefore has this mean without requiring conditional independence (Theorem~\ref{thm:learned-intervention-guarantees}(1)).
{Part (2) bounds the extension error rate under finite within-class exchangeability, including the stability failure probability, and may be of independent interest for training-conditional guarantees in conformal prediction.
This extension requires a finite within-class exchangeable extension and leave-one-out stability averaged over training-sample exchanges.} 
At a high level, the proof combines three ingredients with distinct roles.
Leave-one-out fitting keeps the evaluated trajectory out of the conditioning sample, so its conditional means need not equal its observed averages.
Stability controls changes in the conditional-mean extrema under training-sample exchanges.
The rank argument uses ordering and exchangeability to convert training thresholds into error bounds for the next trajectory.

\section{Rule Learning}
\label{sec:rule-learning}

The Boolean representation gives a mixed-integer linear formulation, as in integer programming approaches to rule learning \citep{su_interpretable_2015,dash_boolean_2018}. As the predicate pool and rule-set capacity grow, this combinatorial problem can become costly to solve directly. We instead use a stochastic local search adapted from FastSRS \citep{molero-rio_fast_2026}.

FastSRS shares an intuition with stochastic gradient descent for nonconvex optimization: sampled examples guide promising local updates. Unlike submodular formulations that support greedy approximation guarantees \citep{yang_learning_2021,harshaw_submodular_2019,wolsey_analysis_1982,balcan_submodular_2018, chen_bicriteria_2023}, our empirical separation is not generally submodular. However, the Boolean structure gives necessary conditions for an edit to change the monitor at a sampled step.
\begin{lemma}[Structure of rule edits]
\label{lem:rule-edit-structure}
Fix $\gS, \gS'\in\gC$. For every $i\in[n]$ and $t\in[T_1]$:
\squishenum
\item
$
X_{t,i}(\gS)=1
\Longleftrightarrow
\{\gR\in\gS:\xi_{t,i}^{\gR}=1\}\ne\emptyset.
$
\item For $\gR\in\gS$ and $\gR'\subseteq[d]$ with $1\le|\gR'|\le K_{\gR}$, let
$\gS':=(\gS\setminus\{\gR\})\cup\{\gR'\}$. Then
\[
X_{t,i}(\gS')-X_{t,i}(\gS)
=\bigl(1-X_{t,i}(\gS\setminus\{\gR\})\bigr)
\bigl(\xi_{t,i}^{\gR'}-\xi_{t,i}^{\gR}\bigr).
\]
\item For $\gR\in\gS$ and $\ell\in[d]$,
$\xi_{t,i}^{\gR\cup\{\ell\}}
=\xi_{t,i}^{\gR}\xi_{t,i}^{\ell}$ and
$
X_{t,i}(\gS\cup\{\{\ell\}\})-X_{t,i}(\gS)
=\bigl(1-X_{t,i}(\gS)\bigr)\xi_{t,i}^{\ell}.
$
\item If $X_{s,i}(\gS')\le X_{s,i}(\gS)$ for every $s\in[T_1]$, then $\widehat U_i^{X(\gS')}\le\widehat U_i^{X(\gS)}$ and $\widehat L_i^{X(\gS')}\le\widehat L_i^{X(\gS)}$. If this holds for every $i\in[n]$, then, for each valid rank $k$, $\widehat U_{(k:\gI_{\mathrm{in},0})}^{X(\gS')}\le\widehat U_{(k:\gI_{\mathrm{in},0})}^{X(\gS)}$ and $\widehat L_{(k:\gI_{\mathrm{in},1})}^{X(\gS')}\le\widehat L_{(k:\gI_{\mathrm{in},1})}^{X(\gS)}$.
These inequalities hold when adding an AND predicate or deleting a rule.
They reverse when adding a rule or removing a predicate from a conjunction of length at least two.
\item If $\gR',\gR\in\gS$ and $\gR'\subsetneq\gR$, then
$
X_{t,i}(\gS\setminus\{\gR\})=X_{t,i}(\gS).
$
\squishenumend
\end{lemma}
\begin{proof}
See Appendix~\ref{app:proof-rule-edit-structure}.
\end{proof}

Sampled trajectories guide local edits, while the Boolean conditions exclude edits that cannot produce the desired change. Appendices~\ref{app:sls-rule-trimming} and~\ref{app:sls-candidate-generation} describe their use in rule trimming and candidate generation. An edit can move both endpoints and need not increase the signed training gap $\widehat G(\gS)$ in \eqref{eq:optimization-signed-gap}, so each candidate is evaluated on the full training sample. We allow equal-gap moves and temporary decreases while keeping the best candidate for final selection.
With exploration probability $\chi\in[0,1]$, the algorithm proceeds as follows.
\squishenum
\item Initialize a random feasible rule set and include $\emptyset$ as a baseline.
\item Sample trajectories with large $\widehat U_i^{X(\gS)}$ in class~0 and small $\widehat L_i^{X(\gS)}$ in class~1, together with relevant prefixes and steps.
\item Generate candidates by single additions or deletions of rules or predicates, with occasional proposals independent of the sampled trajectories for exploration. Trim redundant rules.
\item Evaluate candidates on the full training sample and {keep the best candidate encountered} under \eqref{eq:sls-candidate-order}. Move to the best candidate in the current batch with probability $1-\chi$ and to a uniformly sampled candidate with probability $\chi$.
\item Repeat for $M$ rounds and return the best candidate encountered.
\squishenumend
We repeat the search with multiple seeds and select the returned candidate with the largest signed training gap, breaking ties by \eqref{eq:sls-candidate-order}. Appendix~\ref{app:stochastic-local-search} gives the complete algorithm.

\section{Empirical Illustration}
\label{sec:empirical-illustration}

\paragraph{Setup.}
Here $Y=0$ denotes baseline success and $Y=1$ baseline failure.
We use a Ministral~3 8B coding agent on all 300 SWE-bench Lite tasks
with 180 trajectories for learning and
120 for evaluation using random splitting. We implement the monitor using the hooks interface of {an open-source agent SDK}.
We set $K_{\gR}=K_{\gS}=8$, $(\epsilon_0,\epsilon_1)=(0.1,0.6)$, $\gW=\{50,\ldots,500\}$, and {$\alpha=0.01$}.
We run the single-edit search with twenty seeds
and select the candidate with the largest training gap.
Appendix~\ref{app:stochastic-local-search}
gives the search settings.

\paragraph{Predicate pool.}
We use an LLM to propose candidate predicates and construct $\Xi$ from observable events, including tool errors, repeated actions, and edit calls. We threshold event counts over the observed prefix, initial segments, and recent windows, including lagged windows. We also include step thresholds and negations, yielding $d=5{,}356$ predicates. Appendix~\ref{app:predicate-proposal} describes the proposal and expansion.

\paragraph{Result.}
Across twenty seeds and $200{,}000$ rounds, the largest training gap
is attained by seed~8. The selected monitor is an OR of seven rules
with ten predicates. It combines conditions on repeated failed actions,
tool errors, and edit and test activity across initial segments and recent windows.
See Appendix~\ref{app:selected-monitor} for the full rule definitions.

\begin{table}[b]
\centering
\caption{Results for the candidate with the largest training gap across twenty seeds.
TP/FP: stopped failures/successes; TN/FN: unstopped successes/failures.
TPR  $=\mathrm{TP}/(\mathrm{TP}+\mathrm{FN})$;
Stop prec.\ $=\mathrm{TP}/(\mathrm{TP}+\mathrm{FP})$; Output prec.\ $=\mathrm{TN}/(\mathrm{TN}+\mathrm{FN})$. All rates are percentages.}
\label{tab:optimization-results}
\small
\setlength{\tabcolsep}{5pt}
\begin{tabular}{lrrrrrrrrr}
\toprule
Split & TP & FP & TN & FN & Stop prec. & FPR & TPR & Output prec. & Tokens saved \\
\midrule
Train & {46} & {1} & {33} & {100} & {97.87} & {2.94} & {31.51} & {24.81} & {38.80} \\
Eval & {29} & {3} & {19} & {69} & {90.62} & {13.64} & {29.59} & {21.59} & {34.46} \\
Eval (no early stop) & 0 & 0 & 22 & 98 & -- & 0.00 & 0.00 & 18.33 & 0.00 \\
\bottomrule
\end{tabular}
\end{table}

The selected candidate has $\widehat U_{(31:\gI_{\mathrm{in},0})}^{X(\widehat{\gS}(\mD_{\mathrm{in}}))}=0.68$ and
$\widehat L_{(88:\gI_{\mathrm{in},1})}^{X(\widehat{\gS}(\mD_{\mathrm{in}}))}=0.88$, giving $\widehat\Delta=0.20$.
We evaluate it using
reference $0.78$ and {stake $\widehat\lambda=a_{T_1}(\alpha)/\sqrt{v(\widehat p)}\approx0.32764$}.
Table~\ref{tab:optimization-results} reports intervention decisions and
the fraction of baseline input-plus-output tokens saved.
Appendix~\ref{app:experimental-details} gives more details and evidence trajectories.
The policy stops {$29$} failed and {$3$} successful evaluation
trajectories, with detection rate {$29/98=29.59\%$}, false-positive rate
{$3/22=13.64\%$}, and token savings of {$34.46\%$}.
Among trajectories that continue, output precision is
{$19/88=21.59\%$}, compared with $22/120=18.33\%$ without early stopping.

% ==========================================================================
% \section{Related Work}
% Input from main.tex; see the \input list in the body.
% ==========================================================================
\section{Related Work}
\label{sec:related-work}

Monitoring LLM agents is important for detecting failures and enforcing safety requirements during execution~\citep{sadhuka_e-valuator_2026,wang_agentspec_2026}, especially when autonomous actions carry financial risks~\citep{hua_quantifying_2026}.
LeBet studies how statistical requirements for intervention can guide monitor learning.
Existing monitors enforce explicit specifications \citep{wang_agentspec_2026}
or predict violations from learned models of agent behavior
\citep{wang_probguard_2026}.
Other work uses interventions for debugging and causal repair of agent failures \citep{ma_dover_2026,bonagiri_causalflow_2026}.
Calibrated threshold rules and e-valuator turn verifier outputs into
online decisions with error control
\citep{schirmer_online_2026,sadhuka_e-valuator_2026}.
Related sequential methods include learning betting strategies for a fixed testing horizon \citep{taga_learning_2026} and detecting distribution shifts with conditional conformal test martingales \citep{shaer_testing_2026}.
Betting methods have also been applied to financial risk backtesting \citep{wang_e-backtesting_2026} and robot performance evaluation using simulations \citep{mahboob_betting_2026}.
{\citet{pmlr-v306-gauthier26a} construct betting processes from
payoff differences or policy likelihood ratios to detect strategic deviations
in multi-agent systems, with anytime-valid error control and detection-time
guarantees.}
We connect a learned, composable binary signal to false-intervention
control and detection through conditions on its average conditional mean.
Interpretable rule learning provides the OR-of-AND class and methods for
balancing classification accuracy against sparsity
\citep{wang_bayesian_2017,dash_boolean_2018, rudin_position_2024}.
Association rules have also been analyzed for sequential event prediction using algorithmic stability \citep{rudin_learning_2013}.
FastSRS searches this class by sampling misclassified examples and
proposing local rule edits \citep{molero-rio_fast_2026}.
We adapt this idea to trajectories and prefixes that limit separation
between classwise order statistics.
FunSearch and subsequent work on mathematical discovery use LLMs to propose programs and evaluation feedback to guide search~\citep{romera-paredes_mathematical_2024,ellenberg_generative_2025}.
We similarly use an LLM to propose candidate predicates for the fixed pool $\Xi$, then evaluate their Boolean combinations using empirical separation.
Learn then Test frames risk calibration as multiple hypothesis testing
over a family of predictors \citep{angelopoulos_learn_2025}.
% ==========================================================================
% \section{Discussion}
% Input from main.tex; see the \input list in the body.
% ==========================================================================
\section{Discussion}
\label{sec:discussion}

Although motivated by agent intervention, LeBet addresses a broader problem in online decision-making: learning from complete trajectories while acting only on the observed history. Related work spans early time-series classification~\citep{hartvigsen_adaptive-halting_2019}, anomaly prediction~\citep{chao_early_2025}, optimal stopping~\citep{becker_deep_2019}, and portfolio management~\citep{sutiene_enhancing_2024}. Its central idea is to use statistical conditions for reliable decisions to guide what the monitor learns. {The resulting intervention rule uses the betting error budget $\alpha$ to control the trade-off between false intervention and detection.} This approach may extend to software execution and automated workflows whenever retrospective outcomes provide suitable trajectory labels and the required statistical conditions hold.

\section*{AI Disclosure}
The authors developed the main research topic, ideas, and insights, drafted the mathematical proofs, and selected and checked the literature references. We used AI agents extensively to help complete the manuscript and code, with careful review by the authors. We reviewed the Lean 4 Challenges against the paper's definitions and theorem statements to ensure that the claims were not weakened and no assumptions were added by using Comparator~\citep{lean_fro_comparator}. The Solutions were generated by AI to follow the proofs in the paper and verified by Lean 4.

\bibliography{ref}

% ===========================================================================
%  Appendices.  Comments match the printed order of active inputs.
% ===========================================================================
\clearpage
\appendix

% \crefalias{section}{appendix} % uncomment if you use cleveref
% List main-text sections and appendices, with page numbers and links.
\setcounter{tocdepth}{1}
\etocsettagdepth{paper}{section}
\etocsettagdepth{supplement}{none}
\tableofcontents
\clearpage
                               % Contents
\section{Notation}
\label{app:notation}

We collect the main notation in two tables: quantities and parameters,
and sets and spaces. The third column gives key distinctions and definition locations.
The indices $t$, $i$, and $y$ refer to time, evaluated trajectories,
and intervention classes. In ${s_{y,i,j}}$, $j$ denotes the deleted occurrence;
in $\xi_t^j$, it denotes a predicate. The trajectory index $\star=n+m+1$
denotes the next trajectory and is suppressed when clear from context.
Sample and index vectors are written in bold. Auxiliary
variables introduced within a proof keep their stated local meanings.

\begingroup
\small
\renewcommand{\arraystretch}{1.12}
\setlength{\tabcolsep}{5pt}
\setlength{\defaultaddspace}{3pt}
\setlength{\LTcapwidth}{\textwidth}

\begin{longtable}{@{}>{\raggedright\arraybackslash}p{0.22\textwidth}>{\raggedright\arraybackslash}p{0.24\textwidth}>{\raggedright\arraybackslash}p{\dimexpr0.54\textwidth-4\tabcolsep\relax}@{}}
\caption{Quantities and parameters.}
\label{tab:notation-quantities}\\
\toprule
Symbol & Brief meaning & Additional details \\
\midrule
\endfirsthead
\multicolumn{3}{l}{\tablename~\thetable\ (continued)}\\
\toprule
Symbol & Brief meaning & Additional details \\
\midrule
\endhead
\bottomrule
\endfoot
$n_y,m_y$; $n,m$ & Sample sizes & Fixed training and extension counts in class $y$; $n=n_0+n_1$, $m=m_0+m_1$. The next trajectory is separate. See Section~\ref{sec:problem-statement}.\\
\addlinespace
$T_0,T_1$ & Monitoring endpoints & Earliest intervention time and maximum execution horizon, with $1<T_0<T_1$. See Section~\ref{sec:motivation}.\\
\addlinespace
$d,K_{\gR},K_{\gS}$ & Monitor capacities & Predicate count, maximum predicates per rule, and maximum rules per monitor. See Section~\ref{sec:motivation}.\\
\addlinespace
$p_0,p_1$ & Comparison levels & Population levels in Section~\ref{sec:motivation}; $p_0$ is also the fixed betting reference in Section~\ref{sec:intervening-by-betting}.\\
\addlinespace
{$q_0$} & {Chosen betting reference} & {Reference used by $E_t(\lambda,q_0)$ and $\tau(\lambda,q_0)$ in Proposition~\ref{prop:statistical-guarantees}; $p_0,p_1$ describe the conditional-mean bounds.}\\
\addlinespace
$\lambda,\alpha$ & Betting parameters & Positive stake and {betting error budget}; intervention requires evidence at least $1/\alpha$. See \eqref{eq:monitor-evidence}.\\
\addlinespace
${a_t}(\alpha)$ & Betting scale & $\sqrt{2\log(1/\alpha)/t}$; see Section~\ref{sec:intervening-by-betting}.\\
\addlinespace
{$v(p)$} & {One-sided curvature bound} & {$1/4-[p-1/2]_+^2$; the stake is $a_{T_1}(\alpha)/\sqrt{v(p)}$ for $p<1$. See \eqref{eq:bernoulli-one-sided-hoeffding}.}\\
\addlinespace
$\epsilon_0,\epsilon_1$ & Classwise error budgets & Used for population separation and empirical calibration; $\epsilon_y\in[0,1)$. See Section~\ref{sec:motivation}.\\
\addlinespace
$\beta$ & Failure probability & Confidence failure probability in Proposition~\ref{prop:pathwise-calibration}(2) and Theorem~\ref{thm:learned-intervention-guarantees}.\\
\addlinespace
$k_y$ & Calibration rank & $k_y=n_y-\lfloor n_y\epsilon_y\rfloor$; $n_y-k_y$ is the permitted number of training violations. See Section~\ref{sec:learned-intervention-guarantees}.\\
\addlinespace
{$b_y,W_y$} & {Rank error bound} & {$b_y=\lfloor(n_y+m_y)\theta_y\rfloor$; $W_y$ bounds the extension error rate under a uniform training subset. See Theorem~\ref{thm:learned-intervention-guarantees}.}\\
\addlinespace
${B_m}$ & Stability margin & {Bounds changes in $S_{y,i,i}(\mD;\mI)$ under training-sample exchanges, including changes in conditioning data, apart from exceptional trajectories.} See Assumption~\ref{assump:score-m-stability}.\\
\addlinespace
$\theta_y,\kappa_m$ & Stability allowances & {Exceptional fraction in class $y$ and stability failure probability averaged over training-sample exchanges.} See Assumption~\ref{assump:score-m-stability}.\\
\addlinespace
{$\delta_y(\beta)$} & Error bound & {Rank-based bound on the conditional extension error at confidence $1-\beta$, defined in \eqref{eq:total-statistical-errors}. The intervention guarantees use $\delta_y(\beta/2)$.}\\
\addlinespace
{$r_t(\beta)$, $r(\beta)$} & {Mean-estimation radius} & {Combined uniform and fixed-time Hoeffding radius in Proposition~\ref{prop:pathwise-calibration}(2); $r(\beta):=(r_t(\beta))_{t\in\gW}$. Leave-one-out calibration uses $\beta/(2n_y)$.}\\
\addlinespace
{$\widehat q_y$} & {Conditional-mean bound} & {Takes the $k_y$th training order statistic of $\widehat s_{y,i,i}(\mD;\mI_{\mathrm{in}},r(\beta/(2n_y)))$, adds $2{B_m}$, and clips at $1-y$. $\widehat q_0$ is an upper mean bound and $-\widehat q_1$ a lower mean bound. See \eqref{eq:learned-predictable-levels}.}\\
\addlinespace
{$\widehat\epsilon_y,\delta_y^{\mathrm{bal}}$} & {Selected-quantile errors} & {Latent training exceedance fraction and its conditional extension bound for $m_y=n_y$; see \eqref{eq:selected-training-error} and \eqref{eq:balanced-selected-error}.}\\
\addlinespace
$\sP,\sP_y$ & Probability laws & Joint law $\sP$ and conditional law $\sP_y=\sP(\cdot\mid Y_\star=y)$ when $\sP(Y_\star=y)>0$. Write $\E_y$ for expectation under $\sP_y$. See Section~\ref{sec:problem-statement}.\\
\addlinespace
{$\sQ_{k,h},\sQ_\pi$} & {Sampling laws} & {Uniform subset and permutation laws; $\sP_{y,\pi}=\sP_y\otimes\sQ_\pi$. See \eqref{def:sampling-laws}.}\\
\addlinespace
{$\psi_\lambda(u)$} & {Log normalizer} & {$\psi_\lambda(u)=\log(1-u+ue^\lambda)$, used in \eqref{eq:mean-false-intervention}.}\\
\addlinespace
{$e_0,e_{1,t}$} & {Betting error bounds} & {Betting contributions to false intervention and missed detection, defined in \eqref{eq:mean-false-intervention}.}\\
\addlinespace
$b(\lambda)$ & Linear betting fraction & {$(\exp(\lambda)-1)/e^{\psi_\lambda(p_0)}$}; see \eqref{eq:linear-betting-fraction}.\\
\addlinespace
$q(\lambda),\lambda^\star$ & Tilted probability and optimal stake & {$q(\lambda)=p_0\exp(\lambda)/e^{\psi_\lambda(p_0)}$}; $\lambda^\star=\log\frac{p_1(1-p_0)}{p_0(1-p_1)}$ for $0<p_0<p_1<1$. See Appendix~\ref{app:betting-interpretation}.\\
\addlinespace
$\chi$ & Exploration probability & Probability of unguided proposals and random candidate moves, in $[0,1]$. See Appendices~\ref{app:sls-target-selection} and~\ref{app:sls-evaluation}.\\
\addlinespace
{$M,B,H$} & Search budgets & Rounds, batch size per direction, and maximum attempts per batch, respectively, in Algorithm~\ref{alg:stochastic-local-search}.\\
\addlinespace
$\omega_i,\omega_\star$ & Sampled interactions & Coordinate random elements under $\sP$; $\omega_\star$ is the next agent--environment interaction. See Section~\ref{sec:problem-statement}.\\
\addlinespace
$O_0,R_t,A_t,O_t$ & Interaction variables & Initial request, reasoning trace, action, and resulting observation. The fixed policy satisfies $(R_t,A_t)=f(Z_{t-1})$. See Section~\ref{sec:problem-statement}.\\
\addlinespace
$Z_t,Z_\infty,Z_{t,i}$ & Trajectory histories & History through step $t$, complete trajectory without intervention, and history of trajectory $i$. See Section~\ref{sec:problem-statement}.\\
\addlinespace
$Y_i,Y_\star$ & Intervention labels & $1$ means intervention was needed; $0$ means it was unnecessary. Training and extension labels are fixed by stratification; $Y_\star$ is random. See Section~\ref{sec:problem-statement}.\\
\addlinespace
$\xi_t^j,\xi_t^{\gR}$ & Predicate and rule values & Binary evaluations on the current history. The predicate maps are fixed before training; $\xi_t^{\gR}=\bigwedge_{j\in\gR}\xi_t^j$. See Section~\ref{sec:motivation}.\\
\addlinespace
$X_t(\gS),X_{t,i}(\gS)$ & Monitor values & Disjunction of rules in $\gS$, evaluated on the current history or on trajectory $i$. See Sections~\ref{sec:motivation} and~\ref{sec:learned-intervention-guarantees}.\\
\addlinespace
{$\mu_{t,y}^X$} & Average conditional mean & {$t^{-1}\sum_{u=1}^t\E_y[X_u\mid\gF_{u-1}]$. Arguments $(Z_t;\gF)$ are suppressed when clear from context.} See Section~\ref{sec:motivation}.\\
\addlinespace
{$\widehat\mu_{t,i}(\mD)$} & {Fitted observed average} & {Observed prefix mean on trajectory $i$ using $\widehat X(\mD)$; See Section~\ref{sec:learned-intervention-guarantees}.}\\
\addlinespace
{$\mu_{t,y}^{\widehat X}$} & {Learned average conditional mean} & {Average conditional mean for $\widehat X(\mD_{\mathrm{in}})$ with $\gF_\star(\mD_{\mathrm{in}})$; see \eqref{eq:learned-conditional-means}.}\\
\addlinespace
$\widehat\mu_t^X$ & Observed prefix mean & $t^{-1}\sum_{s=1}^tX_s$; the average includes steps before $T_0$. See \eqref{eq:trajectory-confidence-bounds}.\\
\addlinespace
$\widehat U_i^X,\widehat L_i^X$ & Trajectory extrema & Maximum and minimum of $\widehat\mu_{t,i}^X$ over $t\in\gW$. See \eqref{eq:trajectory-confidence-bounds}.\\
\addlinespace
$v_{(k:\gI)}$, {$v_{(k:\mD)}$} & Order statistic & $k$th smallest value over $\gI$ or the occurrences in $\mD$, counting repetitions; see Sections~\ref{sec:preliminaries} and~\ref{sec:problem-statement}. Section~\ref{sec:learned-intervention-guarantees} suppresses the training sets $\gI_{\mathrm{in},0}$ and $\gI_{\mathrm{in},1}$ for $\widehat U_{(k)}^X$ and $\widehat L_{(k)}^X$, respectively.\\
\addlinespace
{$s_{y,i,j}(\mD;\mI,r)$, $\widehat s_{y,i,j}(\mD;\mI,r)$, $S_{y,i,j}(\mD;\mI)$} & {Signed temporal maxima} & {Add $r_t$ to the signed conditional or observed mean before maximizing over $\gW$, using $\widehat X(\mD_{\mI}^{-j})$ on trajectory $i$ in $\mD$. Conditional means use $\gF_i(\mD_{\mI}^{-j})$; $S_{y,i,j}(\mD;\mI):=s_{y,i,j}(\mD;\mI,0)$, and $j=i$ gives leave-one-out fits. See \eqref{eq:leave-one-out-conditional-means}.}\\
\addlinespace
$\Delta$ & Population separation & Optimal feasible population gap over $\gC$; see \eqref{eq:population-separation}.\\
\addlinespace
$\widehat\Delta_{\mD_{\text{in}}}^{\epsilon_0,\epsilon_1}(\gS)$, $\widehat\Delta$ & Empirical separation & Positive part of the training gap in \eqref{eq:computable-empirical-separation}; $\widehat\Delta$ is its value at $\widehat{\gS}(\mD_{\mathrm{in}})$.\\
\addlinespace
$\widehat G(\gS)$ & Signed training gap & Defined in \eqref{eq:optimization-signed-gap}; $[\widehat G(\gS)]_+$ is the empirical separation.\\
\addlinespace
$\widehat X(\mD)$ & Learned monitor & $X(\widehat{\gS}(\mD))$, with the training sample written explicitly. See Section~\ref{sec:learned-intervention-guarantees}.\\
\addlinespace
{$\widehat p_0,\widehat p_1$} & {Selected empirical levels} & {Classwise training order statistics in \eqref{eq:learned-empirical-levels}, before stability and concentration adjustments.}\\
\addlinespace
$\widehat p,\widehat\lambda$ & Deployed parameters & {On $\gB_0$, midpoint of the selected empirical levels and stake $\widehat\lambda=a_{T_1}(\alpha)/\sqrt{v(\widehat p)}$. The fallback is $(1,1)$ on $\gB_0^c$.}\\
\addlinespace
$E_t(\lambda,{p_0})$ & Betting evidence & Product of one-step betting factors in \eqref{eq:monitor-evidence}.\\
\addlinespace
$\tau(\lambda,{p_0}),\widehat\tau$ & Intervention times & Threshold-crossing times within $\gW$ for fixed and learned monitors and parameters; $\infty$ means no intervention. See \eqref{eq:intervention-time} and Theorem~\ref{thm:learned-intervention-guarantees}.\\
\addlinespace
{$C_y(\mD;\mI)$} & Extension error count & Number of class-$y$ errors outside the training sample {$\mD_{\mI}$}, excluding $\omega_\star$; lies in $\{0,\ldots,m_y\}$. See \eqref{eq:symmetrized-future-error}.\\
\end{longtable}

\begin{longtable}{@{}>{\raggedright\arraybackslash}p{0.22\textwidth}>{\raggedright\arraybackslash}p{0.24\textwidth}>{\raggedright\arraybackslash}p{\dimexpr0.54\textwidth-4\tabcolsep\relax}@{}}
\caption{Sets and spaces.}
\label{tab:notation-sets}\\
\toprule
Symbol & Brief meaning & Additional details \\
\midrule
\endfirsthead
\multicolumn{3}{l}{\tablename~\thetable\ (continued)}\\
\toprule
Symbol & Brief meaning & Additional details \\
\midrule
\endhead
\bottomrule
\endfoot
$\sN,[n]$ & Integer index sets & $\{0,1,\ldots\}$ and $\{1,\ldots,n\}$, respectively. See Section~\ref{sec:preliminaries}.\\
$\Omega,\Omega_y$ & Interaction spaces & Space of single interactions and class-$y$ subset $\{\omega:Y(\omega)=y\}$. See Section~\ref{sec:problem-statement}.\\
\addlinespace
$\mOmega$ & Joint sample space & $\Omega_0^{n_0+m_0}\times\Omega_1^{n_1+m_1}\times\Omega$, including the next interaction. See Section~\ref{sec:problem-statement}.\\
\addlinespace
${\gF,}\gF_\infty,\gF_t,\sigma(Z_t)$ & Information structures & Ambient and time-$t$ $\sigma$-algebras, and the $\sigma$-algebra generated by $Z_t$. See Section~\ref{sec:preliminaries}.\\
\addlinespace
$\gZ_t,\gZ$ & History spaces & Histories of length $t$ and all finite histories, with $\gZ=\bigcup_{t\ge0}\gZ_t$. See Section~\ref{sec:problem-statement}.\\
\addlinespace
$\gR_{\mathrm{trace}},\gA_{\mathrm{tool\_call}},\gO$ & Interaction component spaces & Spaces of reasoning traces, tool-call actions, and observations, respectively, in Section~\ref{sec:problem-statement}.\\
\addlinespace
$\gW$ & Monitoring window & $\{T_0,\ldots,T_1\}$; evidence is updated from step~1, but intervention is restricted to this window. See Section~\ref{sec:motivation}.\\
\addlinespace
$\Xi$ & Fixed predicate pool & Collection of $d$ adapted binary predicate maps, fixed before observing $\mD_{\text{in}}$. See Section~\ref{sec:motivation}.\\
\addlinespace
$\gR,\gS$ & Rule and rule set & Predicate-index set defining a conjunction, and collection of rules defining a disjunction. A learned rule set is denoted $\widehat{\gS}$. See Section~\ref{sec:motivation}.\\
\addlinespace
$\gC$ & Feasible rule-set class & $\{\gS\subseteq 2^{[d]}:|\gS|\le K_{\gS},\ 1\le|\gR|\le K_{\gR}\ \forall\,\gR\in\gS\}$; fixed independently of training data and containing $\emptyset$. See \eqref{eq:population-separation}.\\
\addlinespace
$\widehat{\gS}(\mD)$ & Learned rule set & Output of the fixed algorithm on $\mD$, taking values in $\gC$; need not attain the maximum empirical separation. See Section~\ref{sec:learned-intervention-guarantees}.\\
\addlinespace
$\gP(\gS)$ & Population-feasible levels & Pairs $(p_0,p_1)$ satisfying classwise average-mean bounds for $X(\gS)$ with error budgets $\epsilon_0,\epsilon_1$. See Section~\ref{sec:motivation}.\\
\addlinespace
$\gI_\star$ & All trajectory indices & $[n+m+1]$, with $\star=n+m+1$ for the next trajectory. See Section~\ref{sec:problem-statement}.\\
\addlinespace
$\mI,\gI$ & Index vector and set & $\mI=(i_1,\ldots,i_r)$ has distinct entries in $\gI_\star$; $\gI=\{i_1,\ldots,i_r\}$ omits their order. See Sections~\ref{sec:preliminaries} and~\ref{sec:problem-statement}.\\
\addlinespace
{$\mI_{\mathrm{in}}$} & {Training index vector} & {$(1,\ldots,n)$ in the original indexing; $\mD_{\mathrm{in}}=\mD_{\mI_{\mathrm{in}}}$. It is carried along under the proof's relabeling.}\\
\addlinespace
$\gI_{\text{in},y},\gI_{\text{out},y}$ & Classwise sample indices & Observed and extension indices in class $y$, with fixed sizes $n_y$ and $m_y$. See Assumption~\ref{assump:finite-exchangeable-extension}.\\
\addlinespace
{$\gI_{\mathrm{in/out},y}$} & {Combined classwise indices} & {$\gI_{\text{in},y}\cup\gI_{\text{out},y}$, of size $n_y+m_y$, excluding $\star$. See Section~\ref{sec:problem-statement}.}\\
\addlinespace
$\Pi_{\gI}$ & Permutation group & Permutations of $\gI_\star$ fixing every index outside $\gI$. See Section~\ref{sec:problem-statement}.\\
\addlinespace
$\pi(\mD)$ & Permuted sample vector & $[\pi(\mD)]_i=\omega_{\pi(i)}$. See Section~\ref{sec:problem-statement}.\\
\addlinespace
$\mD$ & Sample vector & $(\omega_1,\ldots,\omega_{n+m+1})$; written as $(1,\ldots,n+m+1)$ using occurrence labels in Appendix~\ref{app:proof-learned-intervention-guarantees}. Also denotes a training vector when used as input to $\widehat{\gS}$.\\
\addlinespace
$\mD_{\mI}$ & Indexed sample vector & $(\omega_{i_1},\ldots,\omega_{i_r})$, preserving the order in $\mI$ and repeated sample values. See Section~\ref{sec:problem-statement}.\\
\addlinespace
{$\mD_y,\mD_{\mathrm{in},y}$} & {Class samples} & {Class-$y$ subvectors of $\mD$ and $\mD_{\mathrm{in}}$, preserving order and occurrences. See Section~\ref{sec:problem-statement}.}\\
\addlinespace
$\mD_{\text{in}}$ & Training data & {$\mD_{\mI_{\mathrm{in}}}$}; its membership is carried along under the ordering used in the proof. Excludes $\omega_\star$. See Section~\ref{sec:motivation}.\\
\addlinespace
{$\mD^{-j}$} & {Leave-one-out training data} & {Deletes occurrence $j$ when it is in $\mD$ and equals $\mD$ otherwise. Equal-valued occurrences remain distinct. See \eqref{eq:leave-one-out-conditional-means}.}\\
\addlinespace
$\mathfrak D_{\text{in/out}}$ & Admissible training vectors & Vectors obtained by within-class swaps with the extension sample, listed in increasing index order and keeping class sizes $n_0,n_1$. See Section~\ref{sec:learned-intervention-guarantees}.\\
\addlinespace
$\gK_\pi,J_{\gK_\pi}^a$ & Training ranks & In the fixed-sample ordering, $\gK_\pi=\pi(\gI_{\mathrm{in},y})=\{i_1,\ldots,i_{n_y}\}$ with $i_1<\cdots<i_{n_y}$; $J_{\gK_\pi}^a=i_{a+1}-a-1$ counts extension occurrences before the $(a+1)$st training occurrence. See \eqref{eq:permuted-training-ranks}.\\
\addlinespace
$\gA,\gQ_y,\gQ$ & Search candidates & {Best rule set encountered}, class-$y$ proposal batch, and batch to evaluate. See Appendix~\ref{app:stochastic-local-search}.\\
\addlinespace
$\gJ_y(\gS),\gT_y(i,\gS)$ & Search target sets & Eligible trajectories and prefix endpoints in \eqref{eq:sls-bottleneck-trajectories} and \eqref{eq:sls-bottleneck-prefixes}.\\
\addlinespace
{$\gE_{\mI,\mI'}^m$} & Stability event & {Compares $S_{y,i,i}(\mD;\mI)$ and $S_{y,i,i}(\mD;\mI')$ in both classes, keeping $\mD$ fixed.} See Assumption~\ref{assump:score-m-stability}.\\
\addlinespace
{$\gB_0$} & {Empirical-separation event} & {Learned empirical separation is positive; see Appendix~\ref{app:auxiliary-intervention}.}\\
\addlinespace
{$\gV_{y,i}(\mD;\mI)$} & Exceedance event & {$S_{y,i,i}(\mD;\mI)$ exceeds its class-$y$ training order statistic by more than $2{B_m}$.} The guarantees in Theorem~\ref{thm:learned-intervention-guarantees} use {$\mI=\mI_{\mathrm{in}}$}. See \eqref{eq:learned-detection-condition}.\\
\addlinespace
\end{longtable}

\endgroup
                                % Appendix A
% ==========================================================================
% \section{Additional Examples and Statistical Illustrations}
% Input from main.tex; see the \input list in the body.
% ==========================================================================
\section{Additional Examples and Statistical Illustrations}

\subsection{Monitor and Intervention Examples}
\label{app:monitor-examples}

Note that our guarantees control false intervention and detection.
The effect of the chosen action on the subsequent run is application dependent.
The examples below pair binary monitors with possible interventions.
Each monitor feeds the betting rule in \eqref{eq:monitor-evidence};
the associated action is taken at its stopping time $\tau(\lambda,{p_0})$.
Predicates involving step $t-1$ are set to zero at $t=1$.

\paragraph{Example 1: resource inefficiency.}
Resource inefficiency may appear as a long trajectory, an expensive step,
or repeated actions. Consider the monitor
\begin{equation*}
X_t^{(1)}
=
\underbrace{\1\{t\ge100\}}_{\text{long trajectory}}
\;\vee\;
\underbrace{
\left(
\1\{c_t>512\}
\wedge
\1\{A_t=A_{t-1}\}
\right)
}_{\text{costly repeated action}},
\end{equation*}
where $c_t$ denotes the number of tokens consumed at step $t$.
At intervention, the system can terminate the run and return the best
available result, or assign the remaining work to a lower-cost model.

\paragraph{Example 2: repeated execution failure.}
Execution failure may appear as self-reported difficulty, consecutive
tool errors, or a failed repeated action. A monitor could be
\begin{align*}
X_t^{(2)}
=
&
\underbrace{
\1\{
R_t
\text{ expresses difficulty}
\}
}_{\text{self-reported difficulty}}
\\
&\vee
\underbrace{
\left(
\1\{O_t\text{ reports an error}\}
\wedge
\1\{O_{t-1}\text{ reports an error}\}
\right)
}_{\text{consecutive tool errors}}
\nonumber\\
&\vee
\underbrace{
\left(
\1\{\exists s<t:A_s=A_t\}
\wedge
\1\{O_t\text{ reports an error}\}
\right)
}_{\text{repeated failed action}}.
\nonumber
\end{align*}
At intervention, the system can transfer the task, current state, and
error history to a more capable model.

\paragraph{Example 3: failed retries and replanning.}
To detect repeated failures of the same action, define
\begin{equation*}
X_t^{(3)}
=
\1\!\left\{
\exists s\in[t-1]:
A_s=A_t,\quad
O_s\text{ and }O_t\text{ report errors}
\right\}.
\end{equation*}
At intervention, the system can pause the current plan and require
a different next action.

\subsection{Illustration of the Statistical Bounds}
\label{app:statistical-bound-illustrations}

\subsubsection{Betting Error Budget}
\label{app:alpha-error-bounds}

\begin{figure}[htbp]
\centering
\includegraphics[width=\textwidth]{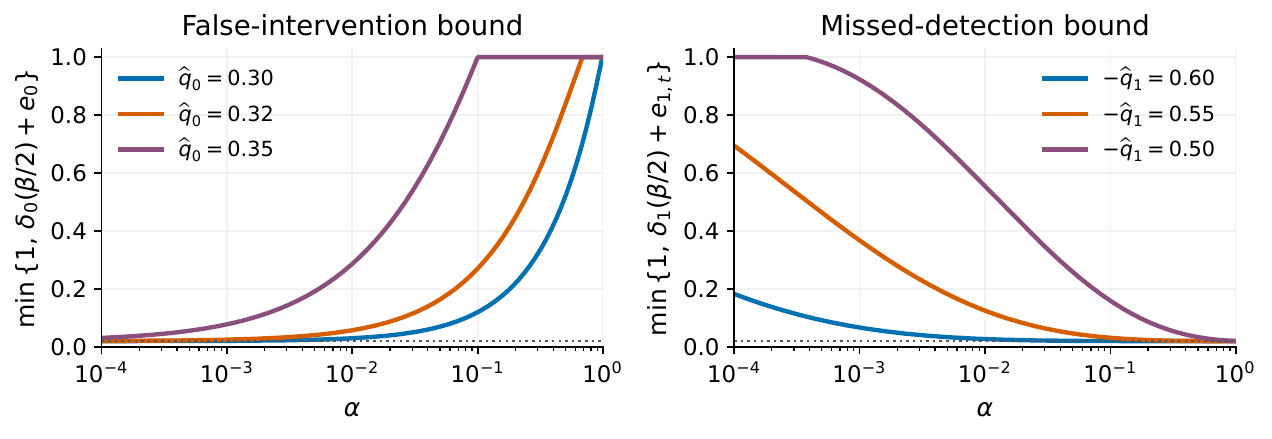}
\caption{{Effect of $\alpha$ on the false-intervention and missed-detection bounds.
The illustrative values are $\widehat p=0.30$, $T_0=50$, $T_1=t=100$, and $\delta_0(\beta/2)=\delta_1(\beta/2)=0.02$.
The stake is recomputed as $\widehat\lambda=a_{T_1}(\alpha)/\sqrt{v(\widehat p)}$ at each $\alpha$.
The left panel varies $\widehat q_0$; the right varies $-\widehat q_1$.
Dotted lines mark the fixed learning contribution.}}
\label{fig:alpha-error-bounds}
\end{figure}

{\begin{proposition}
\label{prop:alpha-error-bounds}
Fix the training quantities and $\beta$ in Theorem~\ref{thm:learned-intervention-guarantees}, and use the default stake for each $\alpha\in(0,1)$.
Here $e_0$ and $e_{1,t}$ have the arguments in parts~(4) and~(5).
For each fixed $t\in\gW$:
\squishenum
\item[\textnormal{(1)}]
The bounds $\min\{1,\delta_0(\beta/2)+e_0\}$ and $\min\{1,\delta_1(\beta/2)+e_{1,t}\}$ are respectively nondecreasing and nonincreasing in $\alpha$.
\item[\textnormal{(2)}]
As $\alpha\downarrow0$, $e_0\to0$ and $e_{1,t}\to1$. On deployment, as $\alpha\uparrow1$,
\[
e_0\to1,\qquad e_{1,t}\to\exp\{-t\KL(\Ber(\min\{-\widehat q_1,\widehat p\})\|\Ber(-\widehat q_1))\}.
\]
\squishenumend
\end{proposition}

\begin{proof}
For part~(1), on deployment, $0<\widehat p<1$ and $\log(1/\alpha)=T_1v(\widehat p)\widehat\lambda^2/2$, so $\widehat\lambda$ decreases with $\alpha$.

For false intervention, if $\widehat q_0\le\widehat p$, then $e_0=\alpha\exp\{T_0[\psi_{\widehat\lambda}(\widehat q_0)-\psi_{\widehat\lambda}(\widehat p)]\}$.
The difference in brackets is nonincreasing in $\widehat\lambda$, so $e_0$ is nondecreasing in $\alpha$.
Otherwise, put
\[
g(\lambda):=\psi_\lambda(\widehat q_0)-\psi_\lambda(\widehat p)-v(\widehat p)\lambda^2/2.
\]
By \eqref{eq:bernoulli-cgf-curvature}, for $\lambda\ge0$,
\[
\partial_\lambda^2\psi_\lambda(\widehat q_0)\le v(\widehat q_0)\le v(\widehat p),
\qquad
\partial_\lambda^2\psi_\lambda(\widehat p)\ge0.
\]
Thus $g(0)=0$, $g''(\lambda)\le0$, and $g'(\lambda)\le g(\lambda)/\lambda<0$ whenever $g(\lambda)<0$.
Since $e_0=\exp(T_1g(\widehat\lambda))$, once the truncated bound falls below one, it decreases as $\widehat\lambda$ increases.

For detection, the threshold is
\[
\widehat\lambda^{-1}\{\psi_{\widehat\lambda}(\widehat p)+a_t(\alpha)^2/2\}
=\widehat\lambda^{-1}\psi_{\widehat\lambda}(\widehat p)
+T_1v(\widehat p)\widehat\lambda/(2t).
\]
Convexity of $\lambda\mapsto\psi_\lambda(\widehat p)$ and $\psi_0(\widehat p)=0$ imply that $\psi_\lambda(\widehat p)/\lambda$ is nondecreasing.
Thus the threshold is nondecreasing in $\widehat\lambda$. Since $\KL(\Ber(h)\|\Ber(p))$ is nonincreasing in $h\le p$, $e_{1,t}$ is also nondecreasing in $\widehat\lambda$.

For part~(2), $0\le\psi_\lambda(u)\le\lambda$ for $u\in[0,1]$, so
$\log e_0\le-\log(1/\alpha)+T_1\widehat\lambda\to-\infty$ as $\alpha\downarrow0$.
The threshold diverges as $\alpha\downarrow0$ and tends to $\widehat p$ as $\alpha\uparrow1$, since $\psi_\lambda(\widehat p)/\lambda\to\widehat p$ as $\lambda\downarrow0$.
Also, $\psi_{\widehat\lambda}(u)\to0$ as $\alpha\uparrow1$, giving $e_0\to1$.

Under the fallback $(\widehat p,\widehat\lambda)=(1,1)$, $e_0=\alpha\exp\{T_0[\psi_1(\widehat q_0)-1]\}$ and $e_{1,t}=1$, so the monotonicity and the limits as $\alpha\downarrow0$ still hold.
\end{proof}

The untruncated $e_0$ need not be monotone where it exceeds one.
Figure~\ref{fig:alpha-error-bounds} shows the truncated bounds for several conditional-mean bounds.
Smaller $\widehat q_0$ and larger $-\widehat q_1$ improve the corresponding guarantees.
}

\subsubsection{Exact Rank Bound}
\label{app:exact-rank-bound-illustration}

\begin{figure}[htbp]
\centering
\includegraphics[width=\textwidth]{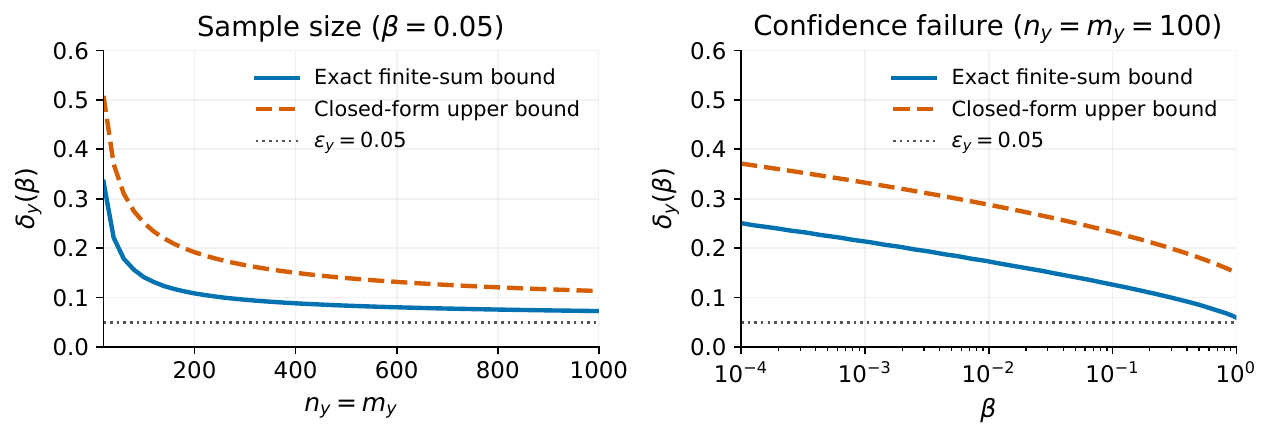}
\caption{{Exact evaluation of \eqref{eq:total-statistical-errors} and its closed-form upper bound \eqref{eq:general-coverage-threshold}.
The illustrative parameters are $m_y=n_y$, $\epsilon_y=0.05$, and $\theta_y=\kappa_m=0$.
The left panel varies $n_y$ from $20$ to $1000$ in increments of $20$, with $\beta=0.05$.
The right panel varies $\beta$, with $n_y=m_y=100$.
The dashed curve is the closed-form bound, truncated at one; the dotted line marks $\epsilon_y$.}}
\label{fig:rank-bound-comparison}
\end{figure}

{Figure~\ref{fig:rank-bound-comparison} compares the exact finite-sum bound $\delta_y(\beta)$ in \eqref{eq:total-statistical-errors} with the closed-form upper bound in \eqref{eq:general-coverage-threshold}, truncated at one.
We evaluate the law in \eqref{eq:uniform-rank-law} and form $W_y$ as in Theorem~\ref{thm:learned-intervention-guarantees}.
The objective in \eqref{eq:total-statistical-errors} is piecewise linear in $u$, so evaluating $0$, $1$, and the support values of $W_y$ gives the minimum.
The difference between the curves reflects the concentration and moment bounds used to obtain \eqref{eq:general-coverage-threshold}.
}

  % Appendix B
\section{Additional Experimental Details}
\label{app:experimental-details}

We evaluate the selected monitor from Section~\ref{sec:empirical-illustration}
on all 180 training and 120 evaluation trajectories.
Each step corresponds to one tool call and its result.
Both splits use the same frozen monitor, reference $\widehat p=0.78$,
stake {$\widehat\lambda\approx0.32764$}, and threshold parameter {$\alpha=0.01$}.
Evidence follows \eqref{eq:monitor-evidence} from the first step,
with threshold checks restricted to $\gW=\{50,\ldots,500\}$.

The monitor stops before the next model call after the triggering tool batch
has completed. We count an intervention only when a later model call
is avoided, and charge all tokens used through the triggering batch.
Thus, crossings after the final model call do not count as interventions;
there are {five} such cases in training and {none} in evaluation.
Trajectories ending before step 50 cannot trigger.
Token savings are the fraction of total baseline input-plus-output
tokens avoided, as reported in Table~\ref{tab:optimization-results}.

{\subsection{LLM-Guided Predicate Proposal}
\label{app:predicate-proposal}
}

The pool in Section~\ref{sec:empirical-illustration} contains 2,678 predicates
and their negations, giving $d=5{,}356$.
Tables~\ref{tab:predicate-proposal-patterns} and~\ref{tab:predicate-expansions}
show only the unnegated forms. Each predicate is a Boolean function of
the observed trajectory prefix.

\paragraph{Proposal.}
Given failure-mode descriptions and implemented predicate types, an LLM
proposes parameterized predicates and candidate conjunctions.
We validate the specification and return errors for correction.

% Source: CyoulinOuroboros/runtime_inputs/early_stop/pool_steps_20_500.yaml.
% Proposal: archive/20260922-pre-refactor/ouroboros/app/lebet_learn/step2_propose_rules.py.
% The examples below correspond to existing entries in the pool's rules list.
\begingroup
\small
\renewcommand{\arraystretch}{1.1}
\setlength{\tabcolsep}{5pt}
\setlength{\defaultaddspace}{3pt}
\setlength{\LTcapwidth}{\textwidth}
\begin{longtable}{@{}>{\raggedright\arraybackslash}p{0.18\textwidth}>{\raggedright\arraybackslash}p{0.32\textwidth}>{\raggedright\arraybackslash}p{\dimexpr0.50\textwidth-4\tabcolsep\relax}@{}}
\caption{Observable patterns and example candidate rules. Each example is a conjunction of one or more predicates.}
\label{tab:predicate-proposal-patterns}\\
\toprule
Pattern & Observable signals & Example candidate rule \\
\midrule
\endfirsthead
\multicolumn{3}{l}{\tablename~\thetable\ (continued)}\\
\toprule
Pattern & Observable signals & Example candidate rule \\
\midrule
\endhead
\bottomrule
\endfoot
Tool failures
& Error status, nonzero exit code, consecutive errors, or recent error fraction.
& Three consecutive tool calls fail.\\
\addlinespace
Repeated actions
& Recurrence of the same tool name and input, with or without a failed result.
& The current action fails and has occurred at least three times.\\
\addlinespace
Edit--test cycles
& Counts of edit calls, repeated edits to one file, and failed test or reproduction commands.
& At least three matching commands have failed, and the current call edits a file for at least the fourth time.\\
\addlinespace
Lack of edits
& No edit by a given step, or many calls since the last edit.
& At step 60 or later, no edit has occurred.\\
\addlinespace
Resource use
& Tool-call count, output tokens per model turn, and cumulative output tokens.
& At least 120 tool calls and more than 20,000 output tokens have been used.\\
\addlinespace
Reported difficulty
& Fixed text patterns indicating difficulty, such as ``stuck'' or ``confused.''
& A difficulty pattern matches and the current test or reproduction command fails.\\
\end{longtable}
\endgroup

Text and command predicates use fixed regular expressions.
Repeated actions use the same tool name and input.

\paragraph{Expansion.}
We vary count, step, token, and error-fraction thresholds over fixed grids,
add temporal conditions, and remove duplicate specifications.

% Semantics: ouroboros/lebet/predicates.py and ouroboros/lebet/recovery.py.
\begingroup
\small
\renewcommand{\arraystretch}{1}
\setlength{\tabcolsep}{5pt}
\setlength{\defaultaddspace}{3pt}
\setlength{\LTcapwidth}{\textwidth}
\begin{longtable}{@{}>{\raggedright\arraybackslash}p{0.20\textwidth}>{\raggedright\arraybackslash}p{0.38\textwidth}>{\raggedright\arraybackslash}p{\dimexpr0.42\textwidth-4\tabcolsep\relax}@{}}
\caption{Temporal constructions before negation.}
\label{tab:predicate-expansions}\\
\toprule
Construction & Meaning & Example \\
\midrule
\endfirsthead
\multicolumn{3}{l}{\tablename~\thetable\ (continued)}\\
\toprule
Construction & Meaning & Example \\
\midrule
\endhead
\bottomrule
\endfoot
Lag
& Evaluate a predicate at an earlier observed step.
& A difficulty pattern matched eight steps ago.\\
\addlinespace
Cumulative count
& Threshold the event count through the current or a lagged step.
& At least three edits occurred through step $t-8$.\\
\addlinespace
Initial-segment count
& Count events in the first fixed number of steps, using the available prefix.
& At least two errors occur in the first $\min\{t,20\}$ steps.\\
\addlinespace
Recent-window count
& Require at least or at most a given count in a window, possibly ending at a lagged step.
& At least eight repeated failed actions occur in the last 50 steps.\\
\addlinespace
Available-history count
& Allow a recent window to use fewer steps before its full length is available.
& At least two repeated failed actions occur in the latest five observed steps.\\
\addlinespace
Unresolved failures
& Count failed attempts for one action or distinct failed actions until a matching success; optionally expire old failures.
& At least three failed attempts of the same action remain unresolved.\\
\end{longtable}
\endgroup

Recent-window predicates are false until the full window is available,
unless available-history evaluation is specified. Negation is applied
after this check.

\subsection{Visualizations}
\label{app:experimental-visualizations}

\begin{figure}[p]
\centering
\includegraphics[width=\textwidth]{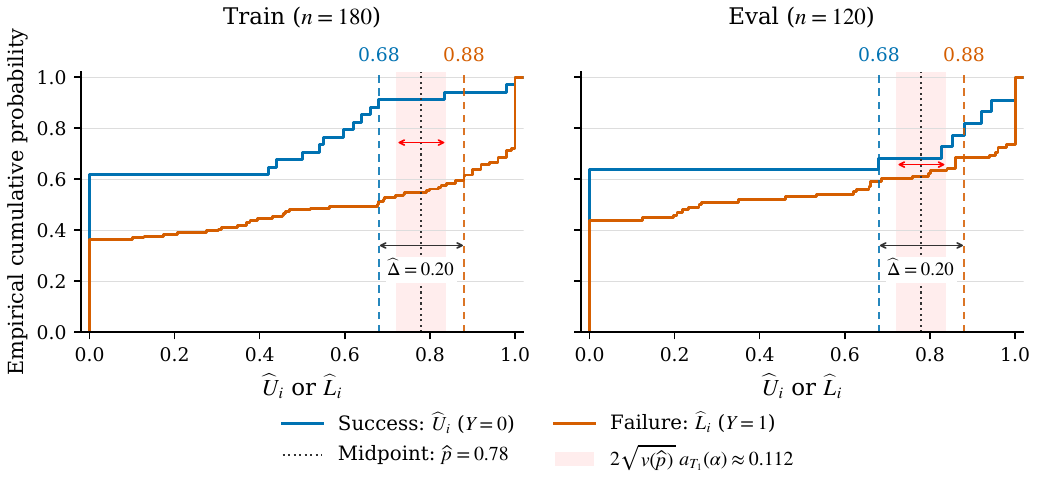}
\caption{Empirical cumulative distributions of $\widehat U_i^{\widehat X(\mD_{\mathrm{in}})}$
for baseline successes and $\widehat L_i^{\widehat X(\mD_{\mathrm{in}})}$ for baseline failures.
Each curve is normalized within its class.
Dashed lines mark the training endpoints $0.68$ and $0.88$;
the dotted line marks their midpoint $\widehat p=0.78$.
The lower arrows span $\widehat\Delta=0.20$.
The centered band and upper arrows span
{$2\sqrt{v(\widehat p)}\,a_{T_1}(\alpha)\approx0.112$}.}
\label{fig:calibration-ul-ecdf}

\medskip
\includegraphics[width=\textwidth]{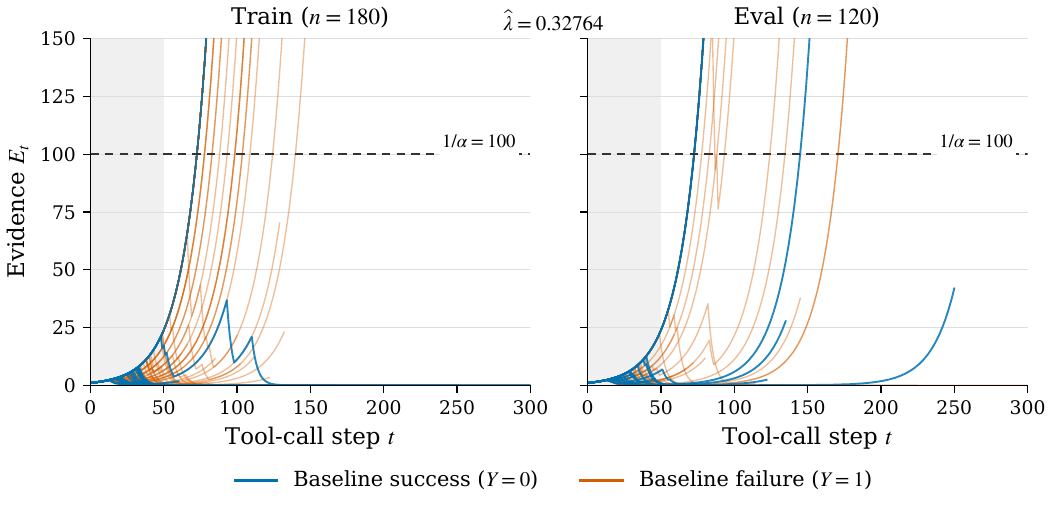}
\caption{Betting evidence on training (left) and evaluation (right) trajectories.
Each curve represents one baseline trajectory, colored by its final outcome.
The horizontal dashed line marks {$1/\alpha=100$}; the shaded region precedes
the monitoring window. Both panels use a linear vertical scale restricted
to {$[0,150]$}; values above {150} are not shown.
Curves continue along the recorded baseline after a threshold crossing
and are shown through the last observed step or $t=300$, whichever comes first.}
\label{fig:experiment-evidence}
\end{figure}

Figure~\ref{fig:calibration-ul-ecdf} compares $\widehat U_i^{\widehat X(\mD_{\mathrm{in}})}$
and $\widehat L_i^{\widehat X(\mD_{\mathrm{in}})}$.
Both are computed on observed prefixes in $\gW$ and set to zero when a
trajectory ends before step 50.
The training endpoints are $0.68$ and $0.88$, giving
$\widehat\Delta=0.20$, and are held fixed in the evaluation panel.

\FloatBarrier
\subsection{Case Studies}
\label{app:experimental-case-studies}

\begin{figure}[!ht]
\centering
\includegraphics[width=0.96\textwidth]{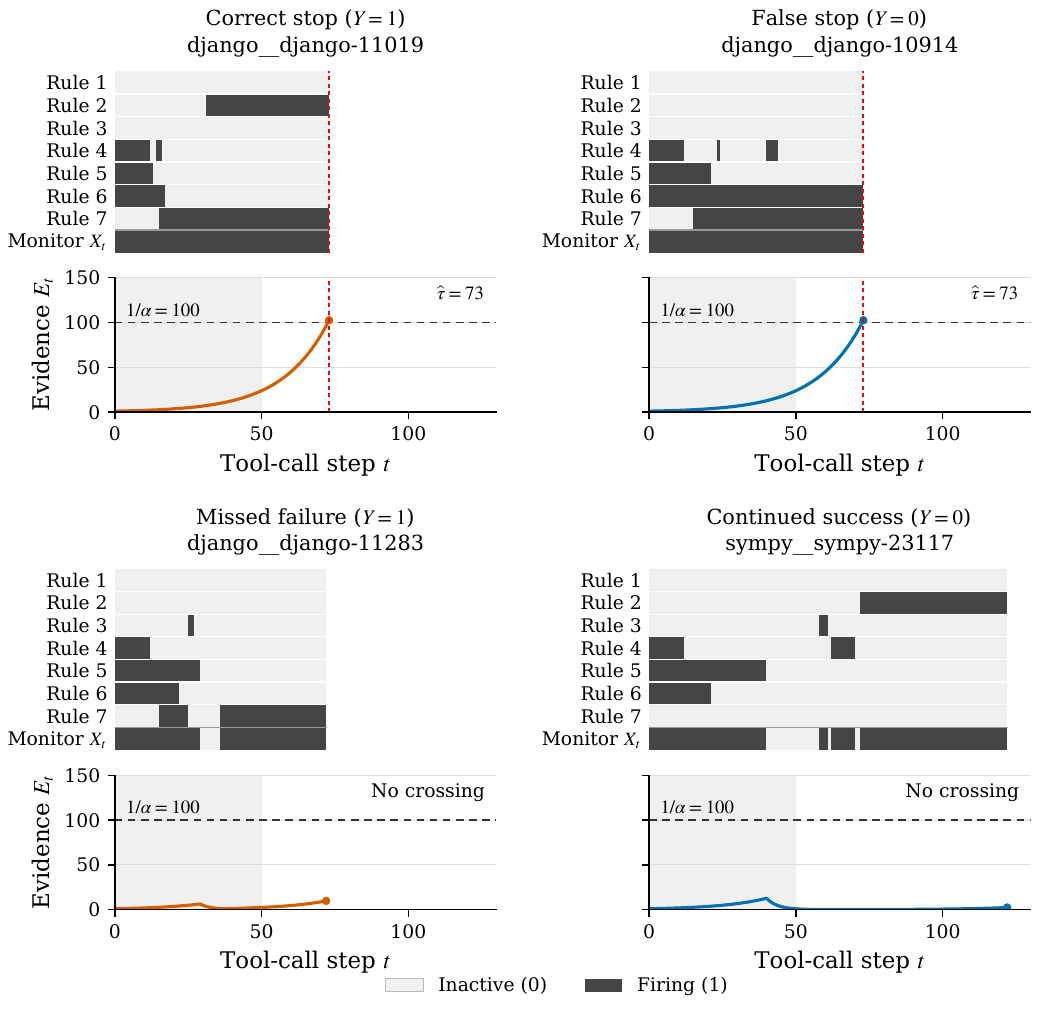}
\caption{Rule values and evidence for four evaluation cases:
TP (top left), FP (top right), FN (bottom left), and TN (bottom right).
Within each panel, the monitor is the OR of the seven rules.
Curves end at the threshold crossing for stopped cases and at the last
observed step for unstopped cases.
Vertical dashed lines mark threshold crossings; horizontal dashed lines
mark {$1/\alpha=100$}. Shading in the evidence plots precedes the monitoring window.}
\label{fig:rule-firing-examples}
\end{figure}

We select one evaluation case for each outcome: TP, FP, FN, and TN.
Within each baseline class, stopped cases are closest to the median trigger
time. Unstopped cases are closest to the median observed length among
trajectories reaching step 50 without a threshold crossing.
Ties are resolved by task identifier.
Rule numbers in Figure~\ref{fig:rule-firing-examples} and
Tables~\ref{tab:case-study-failure}--\ref{tab:case-study-continued-success}
follow Appendix~\ref{app:selected-monitor}.
The tables summarize recorded reasoning, actions, and observations.
A dash denotes no accompanying reasoning text or no firing rules.
Values of $E_t$ are rounded to two decimal places.

\FloatBarrier
\begin{table}[!ht]
\centering
\caption{Selected steps from \texttt{django\_\_django-11019}:
a baseline failure ($Y=1$), correctly stopped (TP).}
\label{tab:case-study-failure}
\small
\setlength{\tabcolsep}{3pt}
\renewcommand{\arraystretch}{1.1}
\begin{tabular}{@{}>{\raggedright\arraybackslash}p{0.035\textwidth}
>{\raggedright\arraybackslash}p{\dimexpr0.23\textwidth-4\tabcolsep\relax}
>{\raggedright\arraybackslash}p{\dimexpr0.25\textwidth-4\tabcolsep\relax}
>{\raggedright\arraybackslash}p{\dimexpr0.27\textwidth-4\tabcolsep\relax}
>{\centering\arraybackslash}p{0.05\textwidth}
>{\raggedleft\arraybackslash}p{0.075\textwidth}
@{\hspace{12pt}}
>{\raggedright\arraybackslash}p{\dimexpr0.09\textwidth-6pt\relax}@{}}
\toprule
$t$ & $R_t$ & $A_t$ & $O_t$ & $X_t$ & $E_t$ & Rules \\
\midrule
13 & Track ordering constraints across media lists.
& \texttt{Edit}: replace the merge method.
& Rejected: \texttt{file\_path} is missing.
& 1 & {2.28} & 5, 6 \\
\addlinespace
14 & Track all ordering constraints.
& \texttt{Edit}: update the merge method.
& File updated.
& 1 & {2.43} & 6 \\
\addlinespace
17 & Try a different approach.
& Retry \texttt{Edit} with the file path.
& Source string not found.
& 1 & {2.94} & 6, 7 \\
\addlinespace
21 & Try a different approach.
& \texttt{Read}: inspect the merge method.
& Code now tracks ordering constraints.
& 1 & {3.79} & 7 \\
\addlinespace
32 & Try a different approach.
& Repeat the same \texttt{Edit}.
& Source string not found.
& 1 & {7.61} & 2, 7 \\
\addlinespace
{73} & Try a different approach.
& Repeat the same \texttt{Edit}.
& Source string not found.
& 1 & {102.57} & 2, 7 \\
\bottomrule
\end{tabular}
\end{table}

\begin{table}[!ht]
\centering
\caption{Selected steps from \texttt{django\_\_django-10914}:
a baseline success ($Y=0$), falsely stopped (FP).}
\label{tab:case-study-success}
\small
\setlength{\tabcolsep}{3pt}
\renewcommand{\arraystretch}{1.1}
\begin{tabular}{@{}>{\raggedright\arraybackslash}p{0.035\textwidth}
>{\raggedright\arraybackslash}p{\dimexpr0.23\textwidth-4\tabcolsep\relax}
>{\raggedright\arraybackslash}p{\dimexpr0.25\textwidth-4\tabcolsep\relax}
>{\raggedright\arraybackslash}p{\dimexpr0.27\textwidth-4\tabcolsep\relax}
>{\centering\arraybackslash}p{0.05\textwidth}
>{\raggedleft\arraybackslash}p{0.075\textwidth}
@{\hspace{12pt}}
>{\raggedright\arraybackslash}p{\dimexpr0.09\textwidth-6pt\relax}@{}}
\toprule
$t$ & $R_t$ & $A_t$ & $O_t$ & $X_t$ & $E_t$ & Rules \\
\midrule
12 & Check the current behavior.
& \texttt{Bash}: run the reproduction script.
& Django settings are not configured.
& 1 & {2.14} & 4, 5, 6 \\
\addlinespace
18 & ---
& \texttt{Bash}: rerun the reproduction script.
& Temporary uploads use \texttt{0o600}; memory uploads use \texttt{0o644}.
& 1 & {3.13} & 5, 6, 7 \\
\addlinespace
19 & Use a consistent default permission.
& \texttt{Edit}: set upload permissions to \texttt{0o644}.
& Settings file updated.
& 1 & {3.34} & 5, 6, 7 \\
\addlinespace
28 & ---
& \texttt{Bash}: run the settings test.
& Both upload types use \texttt{0o644}; cleanup raises an exception.
& 1 & {5.91} & 6, 7 \\
\addlinespace
32 & Run tests with \texttt{unittest}.
& \texttt{Bash}: discover storage tests.
& Tests fail: settings are not configured.
& 1 & {7.61} & 6, 7 \\
\addlinespace
{73} & {Check \texttt{collectstatic} documentation.}
& {\texttt{Grep}: search command documentation.}
& No files found.
& 1 & {102.57} & 6, 7 \\
\bottomrule
\end{tabular}
\end{table}

In the stopped failure, repeated edit attempts cannot find the source string.
In the stopped success, the agent changes the default permissions, yet
rules 6 and 7 remain active.
Both stopped cases have $X_t=1$ for every {$t\le73$}, so their evidence curves
are identical despite their different baseline outcomes.

\begin{table}[!ht]
\centering
\caption{Selected steps from {\texttt{django\_\_django-11283}}:
a baseline failure ($Y=1$), not stopped (FN).}
\label{tab:case-study-missed-failure}
\small
\setlength{\tabcolsep}{3pt}
\renewcommand{\arraystretch}{1.1}
\begin{tabular}{@{}>{\raggedright\arraybackslash}p{0.035\textwidth}
>{\raggedright\arraybackslash}p{\dimexpr0.23\textwidth-4\tabcolsep\relax}
>{\raggedright\arraybackslash}p{\dimexpr0.25\textwidth-4\tabcolsep\relax}
>{\raggedright\arraybackslash}p{\dimexpr0.27\textwidth-4\tabcolsep\relax}
>{\centering\arraybackslash}p{0.05\textwidth}
>{\raggedleft\arraybackslash}p{0.075\textwidth}
>{\raggedright\arraybackslash}p{0.09\textwidth}@{}}
\toprule
$t$ & $R_t$ & $A_t$ & $O_t$ & $X_t$ & $E_t$ & Rules \\
\midrule
{4} & {Examine the migration.}
& {\texttt{Read}: inspect the migration file.}
& {Code updates proxy model permissions.}
& {1} & {1.29} & {4, 5, 6} \\
\addlinespace
{19} & {Reproduce the migration error.}
& {\texttt{Write}: create a reproduction script.}
& {File created.}
& {1} & {3.34} & {5, 6, 7} \\
\addlinespace
{26} & {Run the test script.}
& {\texttt{Bash}: run the reproduction script.}
& {Django settings are not configured.}
& {1} & {5.20} & {3, 5} \\
\addlinespace
{30} & {Simplify the migration test.}
& {\texttt{Write}: replace the reproduction script.}
& {File created.}
& {0} & {4.83} & {---} \\
\addlinespace
{37} & {Simplify the migration test.}
& {\texttt{Write}: replace the reproduction script.}
& {File created.}
& {1} & {1.06} & {7} \\
\addlinespace
{72} & {Simplify the migration test.}
& {\texttt{Write}: replace the reproduction script.}
& {File created.}
& {1} & {9.72} & {7} \\
\bottomrule
\end{tabular}
\end{table}

\begin{table}[!ht]
\centering
\caption{Selected steps from {\texttt{sympy\_\_sympy-23117}}:
a baseline success ($Y=0$), not stopped (TN).}
\label{tab:case-study-continued-success}
\small
\setlength{\tabcolsep}{3pt}
\renewcommand{\arraystretch}{1.1}
\begin{tabular}{@{}>{\raggedright\arraybackslash}p{0.035\textwidth}
>{\raggedright\arraybackslash}p{\dimexpr0.23\textwidth-4\tabcolsep\relax}
>{\raggedright\arraybackslash}p{\dimexpr0.25\textwidth-4\tabcolsep\relax}
>{\raggedright\arraybackslash}p{\dimexpr0.27\textwidth-4\tabcolsep\relax}
>{\centering\arraybackslash}p{0.05\textwidth}
>{\raggedleft\arraybackslash}p{0.075\textwidth}
>{\raggedright\arraybackslash}p{0.09\textwidth}@{}}
\toprule
$t$ & $R_t$ & $A_t$ & $O_t$ & $X_t$ & $E_t$ & Rules \\
\midrule
{30} & {---}
& {\texttt{Bash}: run the reproduction script.}
& {Empty matrix succeeds; empty array raises an unpacking error.}
& {1} & {6.71} & {5} \\
\addlinespace
{34} & {Test again.}
& {\texttt{Bash}: run the reproduction script.}
& {Empty matrix and array tests succeed.}
& {1} & {8.64} & {5} \\
\addlinespace
{40} & {Test again.}
& {\texttt{Bash}: test edge cases.}
& {Zero-sized multidimensional arrays raise division errors.}
& {1} & {12.65} & {5} \\
\addlinespace
{41} & {Handle zero-sized arrays.}
& {\texttt{Edit}: update array bounds checks.}
& {File updated.}
& {0} & {9.71} & {---} \\
\addlinespace
{59} & {Handle zero-sized arrays.}
& {\texttt{Edit}: modify array creation.}
& {Rejected: \texttt{new\_string} is missing.}
& {1} & {0.12} & {3} \\
\addlinespace
{122} & {Try a different approach.}
& {\texttt{Edit}: modify array creation.}
& {Source string not found.}
& {1} & {2.36} & {2} \\
\bottomrule
\end{tabular}
\end{table}

\FloatBarrier
{The unstopped failure repeatedly rewrites a reproduction script, while
the unstopped success fixes the initial empty-array error.
Their evidence remains below 100 and ends at $9.72$ and $2.36$, respectively.}

\subsection{Selected Monitor}
\label{app:selected-monitor}

The selected monitor is an OR of seven rules with ten predicates:
\squishenum
\item At least 8 repeated failed actions occur in the last 50 steps.
\item All 8 steps in the window ending at step $t-8$ match a fixed pattern indicating difficulty.
\item At least 2 repeated failed actions occur in the latest 5 observed steps, using available history.
\item The predicate for at most 2 tool errors in the window of 5 steps ending at step $t-8$ is false.
\item Fewer than 3 edit events occur by step $t-8$.
\item Fewer than 32 repeated failed actions occur by step $t-2$, and fewer than 2 repeated actions occur in the first $\min\{t,50\}$ steps.
\item At least 2 tool errors occur in the first $\min\{t,20\}$ steps, at most 2 matching test or reproduction commands fail in the last 16 steps, and fewer than 8 such failures occur in the first $\min\{t,50\}$ steps.
\squishenumend
          % Appendix C
\section{Stochastic Local Search}
\label{app:stochastic-local-search}

\paragraph{Notation and objective.}
Fix $\Xi=\{\xi_t^\ell:\ell\in[d]\}$, $\mD_{\text{in}}$, $\gW$,
and $\epsilon_0,\epsilon_1$. We describe the search directly on
$\gS\in\gC$, using the rules $\gR\in\gS$ and monitor $X(\gS)$
from the main text.
Define the signed training gap
\begin{equation}
\label{eq:optimization-signed-gap}
\widehat G(\gS)
:=\widehat L_{(n_1-k_1+1:\gI_{\mathrm{in},1})}^{X(\gS)}-\widehat U_{(k_0:\gI_{\mathrm{in},0})}^{X(\gS)}.
\end{equation}
Its positive part $[\widehat G(\gS)]_+$ is the empirical separation in
\eqref{eq:computable-empirical-separation}. The signed gap distinguishes
candidates even when their separation is zero.
All candidates are evaluated on the same training sample.
The preferred candidate maximizes, in lexicographic order,
\begin{equation}
\label{eq:sls-candidate-order}
\left(
\widehat G(\gS'),\quad
-\sum_{\gR\in\gS'}|\gR|,\quad
-|\gS'|
\right).
\end{equation}
Thus ties favor fewer predicates, then fewer rules.
For remaining ties, keep the first candidate encountered.
Within each search run, {$\gA$ keeps the preferred candidate}
under this ordering. Each run uses $M$ rounds.

To describe the effects of rule edits, define coverage at fixed thresholds
$p_0,p_1\in[0,1]$ by
\begin{equation}
\label{eq:sls-class-coverage}
\begin{aligned}
\operatorname{coverage}_0(\gS;p_0)
&:=\bigl|\{i\in\gI_{\text{in},0}:
\widehat U_i^{X(\gS)}\le p_0\}\bigr|,\\
\operatorname{coverage}_1(\gS;p_1)
&:=\bigl|\{i\in\gI_{\text{in},1}:
\widehat L_i^{X(\gS)}\ge p_1\}\bigr|.
\end{aligned}
\end{equation}
The first counts class~0 trajectories whose prefix average is at most $p_0$
throughout $\gW$; the second counts class~1 trajectories whose prefix
average is at least $p_1$ throughout $\gW$.
The empirical error budgets require
$\operatorname{coverage}_y(\gS;p_y)\ge k_y$ for each class $y$.

The procedure uses five modules: initialization, target selection,
rule trimming, candidate generation, and evaluation and update.

\subsection{Initialization}
\label{app:sls-initialization}

Sample $q$ uniformly from $[\min\{K_{\gS},d\}]$, then choose $q$
distinct predicates uniformly to form singleton rules.
Evaluate $\emptyset$ first, followed by the initial rule set, and keep
the preferred candidate under
\eqref{eq:sls-candidate-order}.

\begin{algorithm2e}[H]
\small
\DontPrintSemicolon
\caption{\textsc{Initialize Search}{$(d,K_{\gS})$}}
\label{alg:sls-initialize-search}
\KwIn{Predicate pool size $d\ge1$, and rule-set capacity $K_{\gS}\ge1$.}
\KwOut{Initial $(\gS,\gA)$.}
$q\sim\operatorname{Uniform}([\min\{K_{\gS},d\}])$\;
$J\sim\operatorname{Uniform}(\{J\subseteq[d]:|J|=q\})$\;
$\gS\gets\{\{\ell\}:\ell\in J\}$\;
Compute and store $\widehat G(\emptyset)=0$ and $\widehat G(\gS)$\;
Set $\gA$ to the {preferred candidate in $\{\emptyset,\gS\}$}
under \eqref{eq:sls-candidate-order}\;
\Return $(\gS,\gA)$\;
\end{algorithm2e}

\subsection{Target Selection}
\label{app:sls-target-selection}

Uniform sampling over all trajectories can waste proposals.
Lowering the upper order statistic requires decreasing some
$\widehat U_i^{X(\gS)}$ at or above its current value; raising the lower one
requires increasing some $\widehat L_i^{X(\gS)}$ at or below its current value.
We therefore target more promising
trajectories by sampling from
\begin{equation}
\label{eq:sls-bottleneck-trajectories}
\begin{aligned}
\gJ_0(\gS)
&:=\{i\in\gI_{\text{in},0}:
\widehat U_i^{X(\gS)}\ge\widehat U_{(k_0:\gI_{\mathrm{in},0})}^{X(\gS)}\},\\
\gJ_1(\gS)
&:=\{i\in\gI_{\text{in},1}:
\widehat L_i^{X(\gS)}\le\widehat L_{(n_1-k_1+1:\gI_{\mathrm{in},1})}^{X(\gS)}\}.
\end{aligned}
\end{equation}

For each selected trajectory $i$, to lower $\widehat U_i^{X(\gS)}$ or raise
$\widehat L_i^{X(\gS)}$, we target times $t$ at which the prefix average $\widehat\mu_{t,i}^{X(\gS)}$
attains the corresponding maximum or minimum. The extremum may be
attained at multiple times. The eligible prefix endpoints form the sets
\begin{equation}
\label{eq:sls-bottleneck-prefixes}
\begin{aligned}
\gT_0(i,\gS)
&:=\left\{t\in\gW:
\widehat\mu_{t,i}^{X(\gS)}=\widehat U_i^{X(\gS)}\right\}
\cap\left\{t\in\gW:
\widehat\mu_{t,i}^{X(\gS)}>0\right\},\\
\gT_1(i,\gS)
&:=\left\{t\in\gW:
\widehat\mu_{t,i}^{X(\gS)}=\widehat L_i^{X(\gS)}\right\}
\cap\left\{t\in\gW:
\widehat\mu_{t,i}^{X(\gS)}<1\right\}.
\end{aligned}
\end{equation}
Since $X_{s,i}(\gS)\in\{0,1\}$,
\[
\begin{aligned}
\widehat\mu_{t,i}^{X(\gS)}>0
&\Longleftrightarrow \exists\,s\in[t]:X_{s,i}(\gS)=1,\\
\widehat\mu_{t,i}^{X(\gS)}<1
&\Longleftrightarrow \exists\,s\in[t]:X_{s,i}(\gS)=0.
\end{aligned}
\]
Thus the second set in each intersection ensures
$\{s\in[t]:X_{s,i}(\gS)=1-y\}\ne\emptyset$
for $t\in\gT_y(i,\gS)$.
For each class $y$, sample successively
\begin{equation}
\label{eq:sls-target-sampling}
\begin{aligned}
i_y&\sim\operatorname{Uniform}
\bigl(\{i\in\gJ_y(\gS):\gT_y(i,\gS)\ne\emptyset\}\bigr),\\
t_y&\sim\operatorname{Uniform}\bigl(\gT_y(i_y,\gS)\bigr),\\
s_y&\sim\operatorname{Uniform}
\bigl(\{s\in[t_y]:X_{s,i_y}(\gS)=1-y\}\bigr).
\end{aligned}
\end{equation}
This sampling targets trajectories and prefixes where edits may
improve the corresponding order statistic.

\paragraph{Exploration.}
Let $\chi\in[0,1]$ be the exploration probability. For each
batch, sample $e_y\sim\Ber(\chi)$. When $e_y=0$, use the sampled
$(i_y,s_y)$ to guide edits; when $e_y=1$, sample edits without
trajectory conditions. Algorithm~\ref{alg:sls-sample-target}
combines these choices and returns $(e_y,i_y,s_y)$.

\begin{algorithm2e}[H]
\small
\DontPrintSemicolon
\caption{\textsc{Sample Target}$(\gS,y,\chi)$}
\label{alg:sls-sample-target}
\KwIn{Rule set $\gS$, class $y$, and exploration
probability $\chi\in[0,1]$.}
\KwOut{$(e_y,i_y,s_y)$: $e_y=1$ for exploration, $e_y=0$ for a sampled target.}
$e_y\sim\Ber(\chi)$\;
\lIf{$e_y=1$}{\Return $(1,0,0)$}
$\gJ\gets\{i\in\gJ_y(\gS):\gT_y(i,\gS)\ne\emptyset\}$\;
\lIf{$\gJ=\emptyset$}{\Return $(1,0,0)$}
$i_y\sim\operatorname{Uniform}(\gJ)$\;
$t_y\sim\operatorname{Uniform}(\gT_y(i_y,\gS))$\;
$s_y\sim\operatorname{Uniform}
\bigl(\{s\in[t_y]:X_{s,i_y}(\gS)=1-y\}\bigr)$\;
\Return $(0,i_y,s_y)$\;
\end{algorithm2e}

\subsection{Rule Trimming}
\label{app:sls-rule-trimming}

We remove rules that strictly contain another selected rule to free
capacity and avoid repeated evaluations. Define
\begin{equation}
\label{eq:sls-canonicalization}
\operatorname{Trim}(\gS')
:=\{\gR\in\gS':\nexists\,\gR'\in\gS'\text{ with }
\gR'\subsetneq\gR\}.
\end{equation}
Repeated application of Lemma~\ref{lem:rule-edit-structure}(5)
shows that this preserves the monitor.
For a canonical ordering, sort rules by size and then by predicate
indices. Algorithm~\ref{alg:sls-trim} keeps a rule only if it
contains no previously kept rule.

\begin{algorithm2e}[H]
\small
\DontPrintSemicolon
\caption{\textsc{Trim}$(\gS')$}
\label{alg:sls-trim}
\KwIn{Proposed rule set $\gS'$.}
\KwOut{A rule set with the same monitor and no rule strictly containing another.}
$\gS''\gets\emptyset$\;
\ForEach{$\gR\in\gS'$ \textnormal{in canonical order}}{
  \lIf{$\nexists\,\gR'\in\gS'':\gR'\subsetneq\gR$}{
    $\gS''\gets\gS''\cup\{\gR\}$}
}
\Return $\gS''$\;
\end{algorithm2e}

\subsection{Candidate Generation}
\label{app:sls-candidate-generation}

Given a sampled trajectory and prefix, the Boolean structure constrains
the direction of change under each edit. For example, adding an AND
predicate cannot increase the monitor. These monotonicity properties
and the necessary conditions in Lemma~\ref{lem:rule-edit-structure}
let us exclude edits that cannot produce the desired change at the
sampled step.

\paragraph{Four rule edits.}
Table~\ref{tab:sls-edits} gives the candidate sets obtained by
adding or deleting a rule or predicate. All edits satisfy the
capacities in \eqref{eq:population-separation}. Lemma~\ref{lem:rule-edit-structure}
gives their effects on the monitor, $\widehat U_i^{X(\gS)}$,
$\widehat L_i^{X(\gS)}$, and their order statistics; its proof
appears in Appendix~\ref{app:proof-rule-edit-structure}.
The coverage inequalities follow from \eqref{eq:sls-class-coverage}
with $p_0,p_1$ fixed. In the Effect column, we suppress coverage arguments
and monitor superscripts. The effects on $\widehat U_i$ and
$\widehat L_i$ hold for every trajectory $i$.
The arrows $\downarrow$ and $\uparrow$ denote
nonincrease and nondecrease, respectively.

\begin{table}[ht]
\centering
\caption{Effects of feasible rule edits on coverage, trajectory extrema, and order statistics.}
\label{tab:sls-edits}
\small
\setlength{\tabcolsep}{4pt}
\begin{tabular}{@{}llll@{}}
\toprule
Edit & Conditions & Proposed rule set $\gS'$ & Effect \\
\midrule
\shortstack[l]{Add AND\\predicate}
& \shortstack[l]{$\gR\in\gS,\ \ell\notin\gR$\\$|\gR|<K_{\gR}$}
& $(\gS\setminus\{\gR\})\cup\{\gR\cup\{\ell\}\}$
& \shortstack[l]{$\operatorname{coverage}_0\uparrow,\ \operatorname{coverage}_1\downarrow$\\
$\widehat U_i\downarrow,\ \widehat L_i\downarrow$\\
$\widehat U_{(k_0:\gI_{\mathrm{in},0})}\downarrow,\ \widehat L_{(n_1-k_1+1:\gI_{\mathrm{in},1})}\downarrow$} \\
\addlinespace
Delete rule
& $\gR\in\gS$
& $\gS\setminus\{\gR\}$
& \shortstack[l]{$\operatorname{coverage}_0\uparrow,\ \operatorname{coverage}_1\downarrow$\\
$\widehat U_i\downarrow,\ \widehat L_i\downarrow$\\
$\widehat U_{(k_0:\gI_{\mathrm{in},0})}\downarrow,\ \widehat L_{(n_1-k_1+1:\gI_{\mathrm{in},1})}\downarrow$} \\
\addlinespace
\shortstack[l]{Add singleton\\OR}
& \shortstack[l]{$\ell\in[d]$\\$|\gS|<K_{\gS}$}
& $\gS\cup\{\{\ell\}\}$
& \shortstack[l]{$\operatorname{coverage}_0\downarrow,\ \operatorname{coverage}_1\uparrow$\\
$\widehat U_i\uparrow,\ \widehat L_i\uparrow$\\
$\widehat U_{(k_0:\gI_{\mathrm{in},0})}\uparrow,\ \widehat L_{(n_1-k_1+1:\gI_{\mathrm{in},1})}\uparrow$} \\
\addlinespace
\shortstack[l]{Remove AND\\predicate}
& \shortstack[l]{$\gR\in\gS,\ \ell\in\gR$\\$|\gR|\ge2$}
& $(\gS\setminus\{\gR\})\cup\{\gR\setminus\{\ell\}\}$
& \shortstack[l]{$\operatorname{coverage}_0\downarrow,\ \operatorname{coverage}_1\uparrow$\\
$\widehat U_i\uparrow,\ \widehat L_i\uparrow$\\
$\widehat U_{(k_0:\gI_{\mathrm{in},0})}\uparrow,\ \widehat L_{(n_1-k_1+1:\gI_{\mathrm{in},1})}\uparrow$} \\
\bottomrule
\end{tabular}
\end{table}

Here $\ell\in[d]$ throughout. Removing the last predicate is treated
as deleting its rule. The expansion edit therefore requires $|\gR|\ge2$.
An edit may leave the monitor unchanged
because another rule fires at the same steps.
Contractions can increase class~0 coverage while reducing class~1 coverage;
expansions have the reverse effects. We therefore evaluate the signed gap
after each edit.

\paragraph{Edits that decrease $\widehat U_i^{X(\gS)}$.}
For $e_0=0$, let $(i,s)=(i_0,s_0)$ and draw
\begin{equation}
\label{eq:sls-firing-rules}
\gR\sim\operatorname{Uniform}
\bigl(\{\gR\in\gS:\xi_{s,i}^{\gR}=1\}\bigr).
\end{equation}
Lemma~\ref{lem:rule-edit-structure}(1) ensures that this set is
nonempty. A fair coin chooses between adding an AND predicate and
deleting $\gR$. For addition, draw
\[
\ell\sim\operatorname{Uniform}
\bigl(\{\ell\in[d]:\xi_{s,i}^{\ell}=0\}\bigr),
\qquad
\gS'=(\gS\setminus\{\gR\})\cup\{\gR\cup\{\ell\}\}.
\]
The sampled predicate is absent from $\gR$ because $\gR$ fires at
$(s,i)$. If $|\gR|=K_{\gR}$ or the predicate set is empty, delete
$\gR$ instead. Deletion gives $\gS'=\gS\setminus\{\gR\}$ and is
available whenever $\gR\in\gS$.

For $e_0=1$, sample $\gR$ uniformly from $\gS$ and $\ell$ uniformly
from $[d]$. The same capacity
fallback applies; an AND addition with $\ell\in\gR$ is discarded.
If $\gS=\emptyset$, no contraction is available.

\paragraph{Edits that increase $\widehat L_i^{X(\gS)}$.}
A fair coin chooses between adding a singleton OR rule and removing
an AND predicate. For $e_1=0$, let $(i,s)=(i_1,s_1)$; draw
$\ell\sim\operatorname{Uniform}(\{\ell\in[d]:\xi_{s,i}^{\ell}=1\})$
and propose $\gS'=\gS\cup\{\{\ell\}\}$.
By Lemma~\ref{lem:rule-edit-structure}(3), this makes the selected
step covered. For $e_1=1$, sample $\ell$ from $[d]$.
If $|\gS|=K_{\gS}$ or no eligible predicate exists, attempt removal
instead. Removal samples
\begin{equation}
\label{eq:sls-removable-entries}
\gR\sim\operatorname{Uniform}
\bigl(\{\gR\in\gS:|\gR|\ge2\}\bigr),
\qquad \ell\sim\operatorname{Uniform}(\gR),
\end{equation}
and proposes
$\gS'=(\gS\setminus\{\gR\})\cup\{\gR\setminus\{\ell\}\}$.
This removal is sampled independently of $(i,s)$. Discard the
attempt if no rule has at least two predicates.

\paragraph{Batches.}
Given $(e_y,i_y,s_y)$, a target batch size $B\ge1$, and a maximum
number of attempts $H\ge1$,
Algorithm~\ref{alg:sls-generate-candidates} repeats the class~$y$
edit sampler above. Each proposal is trimmed before checking whether
it is unchanged or duplicates another proposal. Return $B$ distinct
candidates, or $\emptyset$ if fewer than $B$ are collected in $H$ attempts.
Candidates are kept in the order first generated.

\begin{algorithm2e}[H]
\small
\DontPrintSemicolon
\caption{\textsc{Generate Candidates}$(\gS,y,B,H;e_y,i_y,s_y)$}
\label{alg:sls-generate-candidates}
\KwIn{Current rule set $\gS$, class $y$,
target batch size $B\ge1$, maximum attempts $H\ge1$, and $(e_y,i_y,s_y)$.}
\KwOut{$B$ distinct candidates, or $\emptyset$.}
$\gQ_y\gets\emptyset$, $h\gets0$\;
\While{$|\gQ_y|<B$}{
  \lIf{$h=H$}{\Return $\emptyset$}
  $h\gets h+1$\;
  Attempt a class~$y$ edit of $\gS$ using $(e_y,i_y,s_y)$\;
  \If{a proposal $\gS'$ is produced}{
    $\gS'\gets\operatorname{Trim}(\gS')$\;
    \lIf{$\gS'\ne\gS$}{$\gQ_y\gets\gQ_y\cup\{\gS'\}$}
  }
}
\Return $\gQ_y$\;
\end{algorithm2e}

\subsection{Evaluation and Update}
\label{app:sls-evaluation}

For each candidate $\gS'\in\gQ$, compute and store
$\widehat G(\gS')$ on $\mD_{\text{in}}$.
Keep the {preferred candidate in $\{\gA\}\cup\gQ$} under
\eqref{eq:sls-candidate-order}.
Then move to the preferred candidate in $\gQ$ with probability
$1-\chi$, or to a uniformly sampled candidate with probability $\chi$.
The selected move is taken even if the gap decreases.
If $\gQ=\emptyset$, return the current state and sample new targets
in the next round.

Prefix means include all steps from step~1.
Comparisons for target trajectories and tied prefixes use absolute
tolerance $10^{-12}$.

\begin{algorithm2e}[H]
\small
\DontPrintSemicolon
\caption{\textsc{Evaluate and Update}{$(\gS,\gA,\gQ;\chi)$}}
\label{alg:sls-eval}
\KwIn{Current rule set $\gS$, best rule set encountered $\gA$,
proposals $\gQ$, and exploration probability $\chi\in[0,1]$.}
\KwOut{Updated $(\gS,\gA)$.}
\lIf{$\gQ=\emptyset$}{\Return $(\gS,\gA)$}
\ForEach{$\gS'\in\gQ$ \textnormal{in generation order}}{
  Compute and store $\widehat G(\gS')$\;
}
Keep the {preferred candidate in $\{\gA\}\cup\gQ$}
under \eqref{eq:sls-candidate-order} as $\gA$\;
$e\sim\Ber(\chi)$\;
\eIf{$e=1$}{
  $\gS\sim\operatorname{Uniform}(\gQ)$\;
}{
  Set $\gS$ to the preferred candidate in $\gQ$
    under \eqref{eq:sls-candidate-order}\;
}
\Return $(\gS,\gA)$\;
\end{algorithm2e}

\subsection{Stochastic Local Search}
\label{app:sls-complete-search}

Algorithm~\ref{alg:stochastic-local-search} combines the five modules
for one search seed. The target batch size $B$ and maximum attempts $H$
apply separately to each direction. Each candidate is obtained by one edit
followed by trimming. Pass both batches to Algorithm~\ref{alg:sls-eval},
with contraction candidates first.
After $M$ rounds, return {$\gA$}.
Training trajectories are ordered by identifier before random draws.

\begin{algorithm2e}[H]
\small
\DontPrintSemicolon
\caption{\textsc{Stochastic Local Search} (SLS)}
\label{alg:stochastic-local-search}
\KwIn{$\Xi$, $\mD_{\text{in}}$, $\gW$, $(\epsilon_0,\epsilon_1)$,
$(K_{\gR},K_{\gS})$; $M\ge1$ rounds,
target batch size $B\ge1$ per direction, maximum attempts $H\ge1$,
exploration probability $\chi\in[0,1]$, and a search seed.}
\KwOut{{Best rule set encountered $\gA$.}}
Initialize the random generator with the search seed\;
$(\gS,\gA)\gets{\operatorname{InitializeSearch}(d,K_{\gS})}$\;
\For{$h=1,\ldots,M$}{
  \For{$y\in\{0,1\}$}{
    $(e_y,i_y,s_y)\gets\operatorname{SampleTarget}(\gS,y,\chi)$\;
    $\gQ_y\gets
      \operatorname{GenerateCandidates}(\gS,y,B,H;e_y,i_y,s_y)$\;
  }
  $(\gS,\gA)\gets
    {\operatorname{EvaluateAndUpdate}(\gS,\gA,\gQ_0\cup\gQ_1;\chi)}$\;
}
\Return $\gA$\;
\end{algorithm2e}

We run SLS with multiple seeds and select the candidate with the largest
signed training gap across all runs, breaking ties by
\eqref{eq:sls-candidate-order}.

\paragraph{Reported configuration.}
We use $M=10{,}000$ rounds per search seed, seeds $1,\ldots,20$,
target batch size
$B=16$ per direction, maximum attempts $H=256$,
$\chi=0.25$.
Reported runs stored ten candidates per seed; the additional candidates do not affect the search or final selection.
All runs use $(\epsilon_0,\epsilon_1)=(0.1,0.6)$ and the capacities
and window in Section~\ref{sec:empirical-illustration}.
                 % Appendix D
% ==========================================================================
% \section{Betting Interpretation of the Monitor Evidence}
% Input from main.tex; see the \input list in the body.
% ==========================================================================
\section{Betting Interpretation of the Monitor Evidence}
\label{app:betting-interpretation}

In this appendix, we give two equivalent interpretations of the monitor evidence process in \eqref{eq:monitor-evidence}. The first is a linear betting interpretation, while the second is a Bernoulli likelihood-ratio interpretation.

For any $\lambda>0$,
\begin{equation}
\label{eq:betting-normalizer}
{e^{\psi_\lambda(p_0)}}
=
1-p_0+p_0\exp(\lambda)
=
1+p_0\big(\exp(\lambda)-1\big).
\end{equation}
Let
\begin{equation}
\label{eq:linear-betting-fraction}
b(\lambda)
:=
\frac{\exp(\lambda)-1}{{e^{\psi_\lambda(p_0)}}}.
\end{equation}
Since $\lambda>0$, we have $\exp(\lambda)-1>0$ and, therefore,
$b(\lambda)>0$. Furthermore,
\begin{equation*}
p_0b(\lambda)
=
\frac{p_0(\exp(\lambda)-1)}
{1+p_0(\exp(\lambda)-1)}
<1.
\end{equation*}
Thus, both possible one-step multipliers
$1-p_0b(\lambda)$ and $1+(1-p_0)b(\lambda)$ are strictly positive.

Since $X_s\in\{0,1\}$,
\begin{align*}
\frac{\exp(\lambda X_s)}{{e^{\psi_\lambda(p_0)}}}
&=\frac{1+(\exp(\lambda)-1)X_s}{{e^{\psi_\lambda(p_0)}}}\\
&\eqby{eq:betting-normalizer}
1+\frac{\exp(\lambda)-1}{{e^{\psi_\lambda(p_0)}}}(X_s-p_0)
\eqby{eq:linear-betting-fraction}
1+b(\lambda)(X_s-p_0).
\end{align*}
It follows that
\begin{equation*}
E_t(\lambda,{p_0})
=
\prod_{s=1}^t
\left[
1+b(\lambda)(X_s-p_0)
\right].
\end{equation*}
Therefore, $E_t(\lambda,{p_0})$ can be viewed as the wealth process of
a bettor~\citep{waudby-smith_estimating_2024, taga_learning_2026} who repeatedly bets against the reference level $p_0$
using the constant betting fraction $b(\lambda)$.

For $0<p_0<1$, we next give the likelihood-ratio interpretation. Define
\begin{equation}
\label{eq:exponential-tilted-probability}
q(\lambda)
:=
\frac{p_0\exp(\lambda)}
{1-p_0+p_0\exp(\lambda)}
=
\frac{p_0\exp(\lambda)}{{e^{\psi_\lambda(p_0)}}}.
\end{equation}
Then
\begin{equation*}
\frac{q(\lambda)}{p_0}
=
\frac{\exp(\lambda)}{{e^{\psi_\lambda(p_0)}}},
\qquad
\frac{1-q(\lambda)}{1-p_0}
=
\frac{1}{{e^{\psi_\lambda(p_0)}}}.
\end{equation*}
Using $X_s\in\{0,1\}$, we thus obtain
\begin{align}
\frac{\exp(\lambda X_s)}
{1-p_0+p_0\exp(\lambda)}
&=
\left(
\frac{q(\lambda)}{p_0}
\right)^{X_s}
\left(
\frac{1-q(\lambda)}{1-p_0}
\right)^{1-X_s}.
\label{eq:bernoulli-likelihood-ratio-factor}
\end{align}
Multiplying \eqref{eq:bernoulli-likelihood-ratio-factor}
from $s=1$ through $s=t$ gives
\begin{equation*}
E_t(\lambda,{p_0})
=
\prod_{s=1}^t
\left(
\frac{q(\lambda)}{p_0}
\right)^{X_s}
\left(
\frac{1-q(\lambda)}{1-p_0}
\right)^{1-X_s}.
\end{equation*}
Hence, $E_t(\lambda,{p_0})$ is algebraically identical to the likelihood
ratio for $t$ independent observations under $\Ber(q(\lambda))$
versus $\Ber(p_0)$.

For $p_1\in(p_0,1)$, the log-evidence growth satisfies
\begin{align}
\lambda p_1-{\psi_\lambda(p_0)}
&=
p_1\log\frac{\exp(\lambda)}{{e^{\psi_\lambda(p_0)}}}
+(1-p_1)\log\frac{1}{{e^{\psi_\lambda(p_0)}}}
\nonumber\\
&\eqby{eq:exponential-tilted-probability}
p_1\log\frac{q(\lambda)}{p_0}
+(1-p_1)\log\frac{1-q(\lambda)}{1-p_0}
\nonumber\\
&=
\KL(\Ber(p_1)\|\Ber(p_0))
-\KL(\Ber(p_1)\|\Ber(q(\lambda)))
\nonumber\\
&\le\KL(\Ber(p_1)\|\Ber(p_0)).
\label{eq:bernoulli-log-growth-kl}
\end{align}
The last inequality uses nonnegativity of KL divergence.
{Lemma~\ref{lem:bernoulli-kl-duality} gives the maximizing stake
$\lambda^\star=\log\{p_1(1-p_0)/(p_0(1-p_1))\}>0$.
Substituting \eqref{eq:lambda-star-exponential} and \eqref{eq:lambda-star-normalizer}
into \eqref{eq:exponential-tilted-probability} gives $q(\lambda^\star)=p_1$.}
Therefore,
\begin{equation*}
E_t(\lambda^\star,{p_0})
=
\prod_{s=1}^t
\left(
\frac{p_1}{p_0}
\right)^{X_s}
\left(
\frac{1-p_1}{1-p_0}
\right)^{1-X_s}.
\end{equation*}
Since $q(\lambda^\star)=p_1$, equality holds in
\eqref{eq:bernoulli-log-growth-kl}.

%%%%%%%%%%%%%%%%%%%%%%%%%%%%%%%%%%%%%%%%%%%%%%%%%%%%%%%%%%%%%%%%%%%%%%
                   % Appendix E
% ==========================================================================
% \section{Exchangeability, Stability, and Generalization: A Discussion}
% Input from main.tex; see the \input list in the body.
% ==========================================================================
\section{Exchangeability, Stability, and Generalization: A Discussion}
\label{app:exchangeability-stability-discussion}

We relate Assumptions~\ref{assump:finite-exchangeable-extension} and~\ref{assump:score-m-stability} to exchangeability and stability conditions used in conformal prediction and learning theory.

\paragraph{Finite exchangeable extensions within classes.}
\citet[Section~6.2.1]{angelopoulos_theoretical_2027} use a finite exchangeable extension beyond the observed sample and one test point to sharpen marginal coverage guarantees for cross-conformal prediction; the additional points need not be observed. Label-conditional conformal prediction uses calibration observations from each class separately to control coverage conditional on the test label \citep[Section~4]{vovk_conditional_2013}. \citet{xie_conformal_2025} assume joint exchangeability and use test--calibration swaps to establish validity under selective sample splitting. Their assumption also implies within-class symmetry after conditioning on the label vector. Assumption~\ref{assump:finite-exchangeable-extension} combines a finite extension with within-class symmetry of complete trajectories. The extension does not follow from exchangeability of the observed trajectories and $\omega_\star$ alone. Exchangeability alone does not generally ensure training-conditional coverage. Under i.i.d. sampling, \citet[Section~3]{vovk_conditional_2013} establishes such guarantees for split conformal prediction, whereas \citet{bian_training-conditional_2023} show that they need not hold for full conformal and jackknife+ without further assumptions.

\paragraph{Stability and training-conditional generalization.}
\citet{bousquet_stability_2002} derive generalization bounds by controlling changes in a learner's loss under changes in the training data. In conformal prediction, \citet{liang_algorithmic_2025} establish training-conditional coverage for inflated versions of jackknife+ and full conformal under i.i.d. sampling and appropriate stability conditions. {\citet[Sections~4.1--4.3]{liang_algorithmic_2025} control prediction changes in expectation or probability, including swap comparisons that exclude the evaluated point from both fits. Assumption~\ref{assump:score-m-stability} uses the same exclusion but compares the conditional-mean extrema $S_{y,i,i}(\mD;\mI)$, so changes in the conditioning data also enter the comparison. Their guarantees assume i.i.d. sampling; ours use within-class exchanges under Assumption~\ref{assump:finite-exchangeable-extension} and average the stability failure probability over these exchanges.}
 % Appendix F
% \input{appendix/B-calibration-scores-and-finite-class-generalization} % Temporarily excluded
\section{Auxiliary Lemmas}
\label{app:auxiliary-lemmas}

\begin{lemma}[Evidence bounds]
\label{lem:monitor-evidence-bounds}
For $p_0\in[0,1]$, $\lambda>0$, and $t\in\gW$,
\begin{equation}
\label{eq:confidence-evidence-bounds}
t\left\{\lambda(\widehat\mu_t^X-p_0)-{(v(p_0)\lambda^2)/2}\right\}
\le\log E_t(\lambda,{p_0})
\le t\lambda(\widehat\mu_t^X-p_0).
\end{equation}
\end{lemma}

\begin{proof}
The definitions give
\begin{equation}
\label{eq:monitor-log-evidence}
\log E_t(\lambda,{p_0})
\eqby{eq:monitor-evidence}
\lambda\sum_{s=1}^t X_s-t\log(1-p_0+p_0e^\lambda)
\eqby{eq:trajectory-confidence-bounds}
t\lambda\widehat\mu_t^X-t\log(1-p_0+p_0e^\lambda).
\end{equation}
Jensen's inequality and Lemma~\ref{lem:bernoulli-hoeffding} give
\begin{equation}
\label{eq:pathwise-log-normalizer-bounds}
\lambda p_0
\stackrel{\mathrm{Jensen}}{\le}\log(1-p_0+p_0e^\lambda)
\leby{eq:bernoulli-one-sided-hoeffding}
\lambda p_0+{(v(p_0)\lambda^2)/2}.
\end{equation}
\begin{align*}
\log E_t(\lambda,p_0)
&\eqby{eq:monitor-log-evidence}
t\lambda\widehat\mu_t^X-t\log(1-p_0+p_0e^\lambda)\\
&\geby{eq:pathwise-log-normalizer-bounds}
t\{\lambda(\widehat\mu_t^X-p_0)-{v(p_0)\lambda^2/2}\},\\
\log E_t(\lambda,p_0)
&\leby{eq:pathwise-log-normalizer-bounds}t\lambda(\widehat\mu_t^X-p_0).
\end{align*}
\end{proof}

{\begin{lemma}[Bernoulli KL duality]
\label{lem:bernoulli-kl-duality}
For $0<p_0,p_1<1$, the maximum is attained at
$z^\star:=\log\{p_1(1-p_0)/(p_0(1-p_1))\}$, and
\begin{equation}
\label{eq:bernoulli-kl-duality}
\sup_{z\in\R}\{zp_1-\psi_z(p_0)\}
=\KL(\Ber(p_1)\|\Ber(p_0)).
\end{equation}
\end{lemma}

\begin{proof}
The objective is strictly concave, and
\begin{equation}
\label{eq:lambda-star-exponential}
\partial_z\{zp_1-\psi_z(p_0)\}
=p_1-\frac{p_0e^z}{1-p_0+p_0e^z}=0
\quad\Longleftrightarrow\quad
e^z=\frac{p_1(1-p_0)}{p_0(1-p_1)}.
\end{equation}
At $z=z^\star$,
\begin{equation}
\label{eq:lambda-star-normalizer}
e^{\psi_{z^\star}(p_0)}
\eqby{eq:lambda-star-exponential}
1-p_0+\frac{p_1(1-p_0)}{1-p_1}
=\frac{1-p_0}{1-p_1}.
\end{equation}
Hence
\[
z^\star p_1-\psi_{z^\star}(p_0)
\eqby{eq:lambda-star-normalizer}
p_1\log\frac{p_1}{p_0}+(1-p_1)\log\frac{1-p_1}{1-p_0}
=\KL(\Ber(p_1)\|\Ber(p_0)).
\]
\end{proof}
}

\subsection{Hoeffding Bounds}
\label{app:auxiliary-hoeffding}

% Keep long cited headings in the text so they can wrap.
\begin{lemma}
\label{lem:bernoulli-hoeffding}
{\normalfont\bfseries (Bernoulli case of Hoeffding's lemma
\citep[Lemma~1.3]{bardenet_concentration_2015})}
For $m\in[0,1]$ and $z\in\R$,
\begin{equation}
\label{eq:bernoulli-hoeffding-scalar}
(1-m+me^z)e^{-zm}\le e^{z^2/8}.
\end{equation}
{For $z\ge0$,
\begin{equation}
\label{eq:bernoulli-cgf-curvature}
0\le\partial_z^2\psi_z(m)\le v(m),
\end{equation}
and the bound sharpens to
\begin{equation}
\label{eq:bernoulli-one-sided-hoeffding}
(1-m+me^z)e^{-zm}\le e^{v(m)z^2/2}.
\end{equation}}
\end{lemma}
Note that we sharpen the standard bound for $z\ge0$ in \eqref{eq:bernoulli-one-sided-hoeffding}.

\begin{proof}[Proof of \eqref{eq:bernoulli-cgf-curvature} and \eqref{eq:bernoulli-one-sided-hoeffding}]
Put $f_m(z):=\log(1-m+me^z)-mz$ and
$q_m(z):=me^z/(1-m+me^z)$. Then $f_m(0)=f_m'(0)=0$.
For $z\ge0$, $q_m(z)\in[m,1]$, so
{\[
0\le\partial_z^2\psi_z(m)=f_m''(z)=q_m(z)(1-q_m(z))\le v(m).
\]}
{\[
(1-m+me^z)e^{-zm}
=\exp\!\left\{\int_0^z(z-u)f_m''(u)\,du\right\}
\leby{eq:bernoulli-cgf-curvature}\exp\{v(m)z^2/2\}.
\]}
\end{proof}

\begin{lemma}
\label{lem:hoeffding-without-replacement}
{\normalfont\bfseries ({Hoeffding--Serfling inequality}
\citep[{Theorem~2.4}]{bardenet_concentration_2015})}
{Let $x_1,\ldots,x_N\in[a,b]$, with $N\ge2$ and $a<b$, and let $K\sim\sQ_{k,N-k}$, where $1\le k\le N$. Put $\mu:=N^{-1}\sum_{i=1}^N x_i$ and $\rho_k:=1-(k-1)/N$.}
For every $s>0$ and either fixed sign,
\begin{equation}
\label{eq:hoeffding-without-replacement}
\sQ_{k,N-k}\!\left(\pm\left(\frac1k\sum_{i\in K}x_i-\mu\right)\ge s\right)
\le{\exp\!\left\{-\frac{2ks^2}{\rho_k(b-a)^2}\right\}}.
\end{equation}
\end{lemma}

{\begin{lemma}[Weighted tail bound]
\label{lem:weighted-tail-bound}
Suppose $R\ge0$ satisfies $\sP(R>s)\le e^{-cs^2}$ for some $c>0$ and every $s>0$.
For any random variable $h\in[0,1]$ with $q:=\E h>0$,
\begin{equation}
\label{eq:weighted-tail-bound}
\frac{\E[hR]}q\le\sqrt{\frac{\log(e/q)}c}.
\end{equation}
\end{lemma}

\begin{proof}
Cauchy--Schwarz and tail integration \citep[Eq.~(2.2.10)]{vershynin2017lecturesprobabilisticmethodsdata} give
\begin{align*}
\left(\frac{\E[hR]}q\right)^2
&\le\frac{\E[hR^2]}q
=\frac1q\int_0^\infty\E[h\1\{R>\sqrt u\}]\,du\\
&\le\frac1q\int_0^\infty\min\{q,e^{-cu}\}\,du
=\frac{\log(e/q)}c.
\end{align*}
\end{proof}
}

\subsection{Order Statistics and Finite Averaging}
\label{app:auxiliary-order-statistics}

\begin{lemma}[Order-statistic budgets]
\label{lem:order-statistic-budgets}
Let $a_1,\ldots,a_N\in\mathbb R$, with $N\ge1$. For $p\in\mathbb R$ and an integer $0\le r<N$,
\begin{equation}
\label{eq:order-statistic-budget-equivalence}
|\{i:a_i>p\}|\le r\ \Longleftrightarrow\ a_{(N-r:[N])}\le p,
\qquad
|\{i:a_i<p\}|\le r\ \Longleftrightarrow\ p\le a_{(r+1:[N])}.
\end{equation}
{For $b_1,\ldots,b_N\in\R$ and $k\in[N]$,
\begin{equation}
\label{eq:order-statistic-monotonicity}
a_i\le b_i\ \text{for every }i\in[N]
\quad\Longrightarrow\quad
a_{(k:[N])}\le b_{(k:[N])}.
\end{equation}}
\end{lemma}
\begin{proof}
If $a_{(N-r:[N])}\le p$, only the largest $r$ values can exceed $p$.
If $a_{(N-r:[N])}>p$, the largest $r+1$ values exceed $p$.
This proves the first equivalence.
For the second, if $p\le a_{(r+1:[N])}$, only the smallest $r$ values can
be below $p$; otherwise, the smallest $r+1$ values are below $p$.
These statements include ties because violations are strict.
{For \eqref{eq:order-statistic-monotonicity}, put $p:=b_{(k:[N])}$. Then
\[
|\{i:a_i>p\}|\le|\{i:b_i>p\}|
\leby{eq:order-statistic-budget-equivalence}N-k.
\]
Applying \eqref{eq:order-statistic-budget-equivalence} again gives
$a_{(k:[N])}\le p=b_{(k:[N])}$.}
\end{proof}

\begin{lemma}[Finite averaging]
\label{lem:finite-permutation-average}
Let $\sigma_1,\ldots,\sigma_M$ be measurable transformations
preserving a probability law $\sP$, where $M\ge1$.
For every measurable event $A$,
\begin{equation}
\label{eq:finite-permutation-average}
\sP(A)=\E\left[\frac1M\sum_{j=1}^M
\1\{\sigma_j(\omega)\in A\}\right].
\end{equation}
If the average inside the expectation is at most $c$ for every
$\omega$, then $\sP(A)\le c$.
\end{lemma}

\begin{proof}
For every $j$, preservation of the law gives
$\E[\1\{\sigma_j(\omega)\in A\}]=\sP(A)$.
Average these equalities to prove \eqref{eq:finite-permutation-average}.
Taking expectations preserves the bound by $c$.
\end{proof}

\begin{lemma}[A rank bound]
\label{lem:uniform-subset-rank}
Let $\gK\sim\sQ_{k,h}$, where $k,h\ge1$.
Write $\gK=\{i_1,\ldots,i_k\}$ with $i_1<\cdots<i_k$.
For an integer $0\le a<k$, define
\begin{equation*}
J_{\gK}^a:=\bigl|\{i\in[k+h]\setminus \gK:i<i_{a+1}\}\bigr|
=i_{a+1}-a-1.
\end{equation*}
{Its distribution is
\begin{equation}
\label{eq:uniform-rank-law}
\sQ_{k,h}(J_{\gK}^a=j)
=\frac{\binom{a+j}{a}\binom{k+h-a-j-1}{k-a-1}}{\binom{k+h}{k}},
\qquad 0\le j\le h.
\end{equation}}
For $s>0$,
\begin{equation}
\label{eq:uniform-rank-tail}
\sQ_{k,h}\left(J_{\gK}^a>h\left(\frac ak+s\right)\right)
\le {\exp\left\{-\frac{2khs^2}{k+h+2}\right\}}.
\end{equation}
If $h=1$, then
\begin{equation}
\label{eq:one-future-rank}
\sQ_{k,1}(J_{\gK}^a=1)=\frac{a+1}{k+1}.
\end{equation}
\end{lemma}

\begin{proof}
{The event $J_{\gK}^a=j$ places the $(a+1)$st training index at $a+j+1$.
Choose its $a$ predecessors from $[a+j]$ and the remaining $k-a-1$ training indices from the $k+h-a-j-1$ subsequent positions. Dividing this count by $\binom{k+h}{k}$ proves \eqref{eq:uniform-rank-law}.}
Put $j:=\lfloor h(a/k+s)\rfloor$ and define
\begin{equation*}
\gJ:=\{J_{\gK}^a>j\}.
\end{equation*}
If $j\ge h$, then $\gJ=\emptyset$, so \eqref{eq:uniform-rank-tail} holds.
Otherwise, write $\gK^c:=[k+h]\setminus \gK$. The definition of $J_{\gK}^a$ gives
\begin{equation*}
\gJ=
\left\{J_{\gK}^a>h\left(\frac ak+s\right)\right\}
=\{|\gK\cap[a+j+1]|\le a\}.
\end{equation*}
On the event $\gJ$,
\begin{equation*}
|\gK\cap[a+j+1]|\le a,
\qquad
|\gK^c\cap[a+j+1]|=a+j+1-|\gK\cap[a+j+1]|\ge j+1.
\end{equation*}
Hence,
\begin{equation}
\label{eq:rank-to-fixed-prefix}
\frac{|\gK^c\cap[a+j+1]|}{h}-\frac{|\gK\cap[a+j+1]|}{k}
\ge\frac{j+1}{h}-\frac ak>s.
\end{equation}
Since $|\gK\cap[a+j+1]|+|\gK^c\cap[a+j+1]|=a+j+1$,
\begin{align}
\frac{|\gK^c\cap[a+j+1]|}{h}-\frac{|\gK\cap[a+j+1]|}{k}
&=\frac{k+h}{h}
\left(\frac{a+j+1}{k+h}-\frac{|\gK\cap[a+j+1]|}{k}\right)
\nonumber\\
&=\frac{k+h}{k}
\left(\frac{|\gK^c\cap[a+j+1]|}{h}-\frac{a+j+1}{k+h}\right).
\label{eq:finite-population-mean-difference}
\end{align}
The population fraction of indices in $[a+j+1]$ is $(a+j+1)/(k+h)$.
By \eqref{eq:rank-to-fixed-prefix} and
\eqref{eq:finite-population-mean-difference}, on $\gJ$,
\begin{equation}
\label{eq:rank-prefix-tail-events}
\begin{aligned}
\frac{|\gK\cap[a+j+1]|}{k}
&<\frac{a+j+1}{k+h}-\frac{h}{k+h}s,\\
\frac{|\gK^c\cap[a+j+1]|}{h}
&>\frac{a+j+1}{k+h}+\frac{k}{k+h}s.
\end{aligned}
\end{equation}
Thus the fraction in $\gK$ is below the population fraction, while the fraction in $\gK^c$ is above it.
We apply Lemma~\ref{lem:hoeffding-without-replacement} to the indicators
$\1\{i\in[a+j+1]\}$ over the smaller sample:
the lower tail for $\gK$ when $k\le h$, and the upper tail for $\gK^c$ when $h<k$.
{In either case, the finite-population correction gives
\begin{align}
\sQ_{k,h}(\gJ)
&\leby{eq:hoeffding-without-replacement}
\exp\left\{-\frac{2kh}{k+h}
\frac{\max\{k,h\}}{\max\{k,h\}+1}s^2\right\}
\le\exp\left\{-\frac{2khs^2}{k+h+2}\right\}.
\label{eq:fixed-prefix-hoeffding}
\end{align}
The last inequality uses $2\max\{k,h\}\ge k+h$.}
This proves \eqref{eq:uniform-rank-tail}.
For \eqref{eq:one-future-rank}, the single element of $\gK^c$
is uniform on $[k+1]$; it precedes the $(a+1)$st smallest element of $\gK$
exactly when its rank is at most $a+1$.
\end{proof}

\begin{remark}
\label{rem:rank-error-bound}
{\squishenum
\item[\textnormal{(1)}]
The expectation in \eqref{eq:total-statistical-errors} is a finite sum using \eqref{eq:uniform-rank-law}.
\item[\textnormal{(2)}]
For $\kappa_m=0$, $\delta_y(\beta)$ is the mean of the upper $\beta$ fraction of the distribution of $W_y$, including fractional mass at its boundary \citep[Theorem~10]{rockafellar_conditional_2002}.
\squishenumend
}
\end{remark}

{\begin{lemma}[Balanced prefix differences]
\label{lem:balanced-prefix-difference}
Let $\gK\sim\sQ_{n,n}$ and put
$D_{\gK}:=n^{-1}\max_{0\le j\le2n}(|\gK^c\cap[j]|-|\gK\cap[j]|)$, with $[0]=\emptyset$.
For $s>0$,
\begin{equation}
\label{eq:balanced-prefix-tail}
\sQ_{n,n}(D_{\gK}>s)\le\exp\!\left\{-\frac{2n^2s^2}{2n+1}\right\}.
\end{equation}
\end{lemma}

\begin{proof}
We use the reflection formula and logarithmic bound in \citet[Section~3]{GREENE20163701}.
The prefix differences form a walk with $n$ steps of each sign.
For $1\le k\le n$, reflecting the steps after the first visit to $k$ gives
\begin{align}
\sQ_{n,n}(nD_{\gK}\ge k)
&=\frac{\binom{2n}{n-k}}{\binom{2n}{n}}
=\prod_{j=1}^k\frac{n-j+1}{n+j}
\le\exp\!\left\{-\frac{2k^2}{2n+1}\right\}.
\label{eq:balanced-reflection}
\end{align}
The inequality follows from $\log((1-x)/(1+x))\le-2x$ at $x=(2j-1)/(2n+1)$ and $\sum_{j=1}^k(2j-1)=k^2$.
Take $k=\lfloor ns\rfloor+1$; if $k>n$, the event is empty.
\end{proof}
}

\subsection{Conditional Means and Stability}
\label{app:auxiliary-conditional-means}

{\begin{lemma}[Conditional-mean equivariance]
\label{lem:conditional-mean-equivariance}
Suppose Assumption~\ref{assump:finite-exchangeable-extension} holds and $\widehat{\gS}$ is measurable and invariant under permutations of its training sample.
Fix $y$ with $\sP(Y_\star=y)>0$, and let $\Pi$ be the group generated by the permutations for $\sP_y$ in that assumption.
The one-step conditional means in \eqref{eq:leave-one-out-conditional-means} admit versions satisfying, simultaneously for $\pi\in\Pi$, index vectors $\mI$, and occurrence indices $i,j$,
\begin{equation}
\label{eq:conditional-mean-equivariance}
s_{y,i,j}(\pi(\mD);\mI,0)
=s_{y,\pi(i),\pi(j)}(\mD;\pi(\mI),0).
\end{equation}
Here $\pi$ acts componentwise on $\mI$.
The training values $s_{y,i,i}(\mD;\mI,0)$ and their order statistics are $\sigma(\mD_{\mI})$-measurable.
\end{lemma}

\begin{proof}
Work under $\sP_y$ on the joint sample space of $\mD$, where $\pi$ also denotes the induced map.
For a bounded random variable $H$ and a $\sigma$-algebra $\gH\subseteq\sigma(\mD)$, preservation of $\sP_y$ gives
\begin{equation}
\label{eq:permuted-conditional-expectation}
\E_y[H\mid\gH]\circ\pi
=\E_y[H\circ\pi\mid\pi^{-1}\gH].
\end{equation}
Choose $[0,1]$-valued versions of the one-step conditional means.
Since deletion commutes with permutation, replace each version by the average on the left:
\begin{align}
&\frac1{|\Pi|}\sum_{\rho\in\Pi}
\E_y\!\left[
X_{t,\rho(i)}\!\left(\widehat{\gS}(\mD_{\rho(\mI)}^{-\rho(j)})\right)
\,\middle|\,
\mD_{\rho(\mI)}^{-\rho(j)},Z_{t-1,\rho(i)}
\right]\circ\rho^{-1}
\nonumber\\
&\qquad\eqby{eq:permuted-conditional-expectation}
\E_y\!\left[
X_{t,i}\!\left(\widehat{\gS}(\mD_{\mI}^{-j})\right)
\,\middle|\,
\mD_{\mI}^{-j},Z_{t-1,i}
\right].
\label{eq:equivariant-conditional-mean-version}
\end{align}
Reindexing the sum by $\rho\mapsto\rho\circ\pi$ gives equivariance of these one-step versions.
Time averaging and the signed maximum in \eqref{eq:leave-one-out-conditional-means} at $r=0$ then give \eqref{eq:conditional-mean-equivariance}.
For a training occurrence $i$,
\begin{equation}
\label{eq:loo-training-measurability}
\sigma(\mD_{\mI}^{-i},Z_{t-1,i})\subseteq\sigma(\mD_{\mI}),
\end{equation}
so the training values and their order statistics are $\sigma(\mD_{\mI})$-measurable.
\end{proof}

\begin{lemma}[Class-conditional stability failure]
\label{lem:class-conditional-stability-failure}
Suppose Assumption~\ref{assump:score-m-stability} holds and $\sP(Y_\star=y)>0$.
\begin{equation}
\label{eq:class-conditional-stability-failure}
\sP_{y,\pi}((\gE_{\pi(\mI_{\mathrm{in}}),\mI_{\mathrm{in}}}^m)^c)
\le\frac{\E^{\sQ_\pi}\sP((\gE_{\pi(\mI_{\mathrm{in}}),\mI_{\mathrm{in}}}^m)^c)}{\sP(Y_\star=y)}
\le\frac{\kappa_m}{\sP(Y_\star=y)}.
\end{equation}
\end{lemma}

\begin{proof}
Conditioning on $Y_\star=y$ and applying Assumption~\ref{assump:score-m-stability} gives the bound.
\end{proof}
}

{\begin{lemma}[Training-sample averaging]
\label{lem:class-permutation-average}
Under the hypotheses and versions of Lemma~\ref{lem:conditional-mean-equivariance},
let $\pi$ have the independent uniform law in Assumption~\ref{assump:score-m-stability}.
For every bounded measurable $f$,
\begin{equation}
\E_y[f(\mD_{\mathrm{in}},C_y(\mD;\mI_{\mathrm{in}}))]
=\E^{\sP_{y,\pi}}[f(\mD_{\pi(\mI_{\mathrm{in}})},C_y(\mD;\pi(\mI_{\mathrm{in}})))].
\label{eq:class-permutation-average}
\end{equation}
\end{lemma}

\begin{proof}
Identify trajectories with their occurrence indices, keeping repeated values as distinct occurrences.
For each fixed $\pi$, the permuted training sample is
\begin{equation}
\label{eq:permuted-training-sample}
\pi(\mD)_{\mathrm{in}}=\mD_{\pi(\mI_{\mathrm{in}})}.
\end{equation}
Assumption~\ref{assump:finite-exchangeable-extension} gives, under $\sP_y$,
\begin{equation}
\label{eq:permuted-full-sample-law}
\pi(\mD)\overset{d}{=}\mD.
\end{equation}
By \eqref{eq:conditional-mean-equivariance} and \eqref{eq:symmetrized-future-error},
\begin{equation}
\label{eq:reindexed-extension-error}
C_y(\pi(\mD);\mI_{\mathrm{in}})
=\sum_{i\in\gI_{\mathrm{out},y}}
\1\{\gV_{y,\pi(i)}(\mD;\pi(\mI_{\mathrm{in}}))\}
=C_y(\mD;\pi(\mI_{\mathrm{in}})).
\end{equation}
Using \eqref{eq:permuted-full-sample-law},
\begin{align}
\label{eq:training-index-exchangeability}
(\mD_{\pi(\mI_{\mathrm{in}})},C_y(\mD;\pi(\mI_{\mathrm{in}})))
&\eqby{eq:reindexed-extension-error}
(\pi(\mD)_{\mathrm{in}},C_y(\pi(\mD);\mI_{\mathrm{in}}))
\nonumber\\
&\overset{d}{=}(\mD_{\mathrm{in}},C_y(\mD;\mI_{\mathrm{in}})).
\end{align}
Average \eqref{eq:training-index-exchangeability} over $\pi$ to obtain \eqref{eq:class-permutation-average}.
\end{proof}
}

\subsection{Martingales and Concentration}
\label{app:auxiliary-concentration}

The following lemmas state standard martingale and concentration results in our notation~\citep{chow_probability_1997,waudby-smith_estimating_2024}.
An $(\gF_t)$-adapted process $(M_t)_{t\in\sN}$ is a martingale under $\sP$ if $\E|M_t|<\infty$ for every $t$ and $\E[M_t\mid\gF_{t-1}]=M_{t-1}$ almost surely for every $t\ge1$.
Throughout this subsection, fix a class $y$ and work under $\sP_y=\sP(\cdot\mid Y=y)$.
{Let $X$ be adapted to $\gF$.} Use the notation
\begin{equation}
\label{eq:class-trajectory-conditional-mean}
S_t:=\sum_{s=1}^tX_s,
\qquad
A_{y,t}:={\sum_{s=1}^t\E_y[X_s\mid\gF_{s-1}]
=t\mu_{t,y}^X(Z_t;\gF)}.
\end{equation}
Choose conditional-expectation versions in $[0,1]$. All statements involving
conditional expectations hold almost surely under this same law.

\begin{lemma}[Bernoulli exponential martingale and Ville's inequality]
\label{lem:conditional-bernoulli-martingale}
\leavevmode\par
Let {$\lambda\in\R$ be $\gF_0$-measurable}.
\begin{equation}
\label{eq:conditional-bernoulli-martingale}
M_{y,t}(\lambda):=
\exp\left\{\lambda S_t-\sum_{s=1}^t\psi_\lambda({\E_y[X_s\mid\gF_{s-1}]})\right\},
\qquad M_{y,0}(\lambda)=1,
\end{equation}
is a positive martingale under $\sP_y$.
Any nonnegative supermartingale $M$ with $M_0=1$ satisfies, for a positive
{$\gF_0$-measurable} $a$ and finite $T$,
\begin{equation}
\label{eq:ville-conditional-martingale}
{\sP_y\left(\max_{0\le t\le T}M_t\ge a\,\middle|\,\gF_0\right)\le a^{-1}.}
\end{equation}
\end{lemma}
Note that we allow $\gF_0$-measurable $\lambda$ and $a$ in \eqref{eq:conditional-bernoulli-martingale} and \eqref{eq:ville-conditional-martingale}.

{\begin{proof}
Since $X_s\in\{0,1\}$ and $\lambda$ is $\gF_{s-1}$-measurable,
\[
\E_y[e^{\lambda X_s}\mid\gF_{s-1}]
=1+(e^\lambda-1)\E_y[X_s\mid\gF_{s-1}]
=e^{\psi_\lambda(\E_y[X_s\mid\gF_{s-1}])}.
\]
Thus \eqref{eq:conditional-bernoulli-martingale} gives
$\E_y[M_{y,s}(\lambda)\mid\gF_{s-1}]=M_{y,s-1}(\lambda)$ and $\E_y M_{y,s}(\lambda)=1$.
For \eqref{eq:ville-conditional-martingale}, let
$\sigma:=\inf\{t\ge0:M_t\ge a\}\wedge T$.
Conditional optional stopping \citep[Section~7.4, Fact~7.5]{ramdas_hypothesis_2025} gives
\[
a\,\sP_y\left(\max_{0\le t\le T}M_t\ge a\,\middle|\,\gF_0\right)
\le\E_y[M_\sigma\mid\gF_0]\le1.
\]
\end{proof}
}

{\begin{lemma}[Conditional Bernoulli lower-tail bound]
\label{lem:conditional-bernoulli-lower-tail}
For $t\ge1$ and $\gF_0$-measurable $0<h<p\le1$,
\begin{equation}
\label{eq:conditional-bernoulli-lower-tail}
\sP_y(S_t<th,\ A_{y,t}\ge tp\mid\gF_0)
\le \exp\{-t\KL(\Ber(h)\|\Ber(p))\}.
\end{equation}
For $p=1$, the bound is zero; for $h\le0$, the event is empty.
\end{lemma}

\begin{proof}
For a negative $\gF_0$-measurable $z$, $\psi_z$ is decreasing and concave.
On $\{S_t<th,\ A_{y,t}\ge tp\}$, Jensen's inequality gives
\begin{align}
\log M_{y,t}(z)
&\eqby{eq:conditional-bernoulli-martingale}
zS_t-\sum_{s=1}^t\psi_z(\E_y[X_s\mid\gF_{s-1}])
\nonumber\\
&\ge zS_t-t\psi_z(A_{y,t}/t)
>t\{zh-\psi_z(p)\}.
\label{eq:alternative-evidence-lower-bound}
\end{align}
Conditional Markov's inequality gives
\begin{align*}
\sP_y(S_t<th,\ A_{y,t}\ge tp\mid\gF_0)
&\leby{eq:alternative-evidence-lower-bound}
e^{-t(zh-\psi_z(p))}\E_y[M_{y,t}(z)\mid\gF_0]\\
&\eqby{eq:conditional-bernoulli-martingale}
e^{-t(zh-\psi_z(p))}.
\end{align*}
On $\{p<1\}$, take $z=\log\{h(1-p)/(p(1-h))\}<0$ and apply \eqref{eq:bernoulli-kl-duality}.
On $\{p=1\}$, take $z=-n$; the bound is $e^{-tn(1-h)}\to0$ as $n\to\infty$.
\end{proof}
}

\begin{lemma}[Martingale Hoeffding inequality]
\label{lem:monitor-martingale-concentration}
For a {real-valued $\gF_0$-measurable} $z$,
\begin{equation}
\label{eq:conditional-hoeffding}
\E_y[e^{z(X_s-{\E_y[X_s\mid\gF_{s-1}]})}\mid\gF_{s-1}]\le e^{z^2/8}.
\end{equation}
For $t\ge1$ and a positive {$\gF_0$-measurable} $u$,
\begin{equation}
\label{eq:monitor-lower-tail}
{\sP_y\left(\pm(S_t-A_{y,t})\ge u\,\middle|\,\gF_0\right)\le e^{-2u^2/t}.}
\end{equation}
\end{lemma}
Note that we allow $\gF_0$-measurable $z$ and $u$, with the tail bound \eqref{eq:monitor-lower-tail} conditional on $\gF_0$.

{\begin{proof}
Substitute $m=\E_y[X_s\mid\gF_{s-1}]$ in \eqref{eq:bernoulli-hoeffding-scalar} to obtain \eqref{eq:conditional-hoeffding}.
For $z=4u/t$, iterating \eqref{eq:conditional-hoeffding}
\citep[Section~5.1]{ramdas_game-theoretic_2023} gives
\[
\E_y[e^{\pm z(S_t-A_{y,t})-tz^2/8}\mid\gF_0]\le1.
\]
Conditional Markov's inequality yields
\[
\sP_y\left(\pm(S_t-A_{y,t})\ge u\,\middle|\,\gF_0\right)
\le e^{-zu+tz^2/8}=e^{-2u^2/t}.
\]
\end{proof}
}

\subsection{Intervention Events}
\label{app:auxiliary-intervention}

{\begin{lemma}[Conditional intervention bounds]
\label{lem:conditional-intervention-bounds}
Let $X$ be adapted to $\gF$, and let $\gH\subseteq\gF_0$ be a $\sigma$-algebra.
Let $\lambda>0$ and $p_0,p_1,q_0\in[0,1]$ be $\gH$-measurable.
For $\alpha\in(0,1)$ and $t\in\gW$, the following hold almost surely under the corresponding class law:
\begin{align}
\sP_0(\tau(\lambda,q_0)<\infty\mid\gH)
&\le \sP_0(\max_{s\in\gW}\mu_{s,0}^X>p_0\mid\gH)
+e_0(p_0;\lambda,q_0,\alpha),
\label{eq:conditional-false-intervention}\\
\sP_1(\tau(\lambda,q_0)>t\mid\gH)
&\le \sP_1(\mu_{t,1}^X<p_1\mid\gH)
+e_{1,t}(p_1;\lambda,q_0,\alpha).
\label{eq:conditional-missed-detection}
\end{align}
\end{lemma}

\begin{proof}
Use \eqref{eq:class-trajectory-conditional-mean}.
For \eqref{eq:conditional-false-intervention}, put
$c:=T_1[\psi_\lambda(p_0)-\psi_\lambda(q_0)]_+-T_0[\psi_\lambda(q_0)-\psi_\lambda(p_0)]_+$ and let
\begin{equation}
\label{eq:null-validity-event-probability}
G_0:=\{\max_{s\in\gW}\mu_{s,0}^X\le p_0\}.
\end{equation}
For $\lambda>0$, $\psi_\lambda$ is increasing and concave on $[0,1]$.
Since $s[\psi_\lambda(p_0)-\psi_\lambda(q_0)]\le c$ for $s\in\gW$, on $G_0$,
\begin{align}
\sum_{r=1}^s\psi_\lambda(\E_0[X_r\mid\gF_{r-1}])
&\stackrel{\mathrm{Jensen}}{\le} s\psi_\lambda(A_{0,s}/s)
\nonumber\\
&\leby{eq:null-validity-event-probability}s\psi_\lambda(p_0)
\le s\psi_\lambda(q_0)+c.
\label{eq:conditional-cgf-domination-null}
\end{align}
Consequently,
\begin{align}
\log E_s(\lambda,q_0)-\log M_{0,s}(\lambda)
&\eqby{eq:monitor-log-evidence}\lambda S_s-s\psi_\lambda(q_0)-\log M_{0,s}(\lambda)
\nonumber\\
&\eqby{eq:conditional-bernoulli-martingale}
\sum_{r=1}^s\psi_\lambda(\E_0[X_r\mid\gF_{r-1}])-s\psi_\lambda(q_0)
\nonumber\\
&\leby{eq:conditional-cgf-domination-null}c.
\label{eq:evidence-dominated-by-martingale}
\end{align}
The definition \eqref{eq:intervention-time} gives
$\{\tau(\lambda,q_0)<\infty\}\subseteq
G_0^c\cup\{\max_{0\le s\le T_1}M_{0,s}(\lambda)\ge e^{-c}/\alpha\}$.
The tower property and $\gH$-measurability of $c$ yield
\begin{align*}
\sP_0(\tau(\lambda,q_0)<\infty\mid\gH)
&\le \sP_0(G_0^c\mid\gH)
+\E_0\!\left[\sP_0\!\left(\max_{0\le s\le T_1}M_{0,s}(\lambda)\ge e^{-c}/\alpha
\,\middle|\,\gF_0\right)\,\middle|\,\gH\right]\\
&\leby{eq:ville-conditional-martingale}
\sP_0(G_0^c\mid\gH)+\E_0[\min\{1,\alpha e^c\}\mid\gH]\\
&\le \sP_0(G_0^c\mid\gH)+\alpha e^c\\
&\eqby{eq:mean-false-intervention}
\sP_0(G_0^c\mid\gH)+e_0(p_0;\lambda,q_0,\alpha).
\end{align*}

For \eqref{eq:conditional-missed-detection}, put
$h:=\lambda^{-1}\{\psi_\lambda(q_0)+a_t(\alpha)^2/2\}$.
On $\{h\ge p_1\}$, $e_{1,t}(p_1;\lambda,q_0,\alpha)=1$ and the bound is immediate.
On $\{h<p_1\}$, $0<h<p_1\le1$ and \eqref{eq:mean-false-intervention} gives
$e_{1,t}(p_1;\lambda,q_0,\alpha)=\exp\{-t\KL(\Ber(h)\|\Ber(p_1))\}$.
By \eqref{eq:monitor-log-evidence},
\[
\{\tau(\lambda,q_0)>t,\ A_{1,t}\ge tp_1\}
\subseteq\{S_t<th,\ A_{1,t}\ge tp_1\}.
\]
The tower property and $\gH$-measurability of $h,p_1$ give
\begin{align}
\sP_1(\tau(\lambda,q_0)>t,\ A_{1,t}\ge tp_1\mid\gH)
&\le \E_1[\sP_1(S_t<th,\ A_{1,t}\ge tp_1\mid\gF_0)\mid\gH]
\nonumber\\
&\leby{eq:conditional-bernoulli-lower-tail}
\E_1[e_{1,t}(p_1;\lambda,q_0,\alpha)\mid\gH]
\nonumber\\
&=e_{1,t}(p_1;\lambda,q_0,\alpha).
\label{eq:alternative-event-failures}
\end{align}
Thus
\begin{align*}
\sP_1(\tau(\lambda,q_0)>t\mid\gH)
&\leby{eq:alternative-event-failures}
\sP_1(A_{1,t}<tp_1\mid\gH)+e_{1,t}(p_1;\lambda,q_0,\alpha)\\
&\eqby{eq:class-trajectory-conditional-mean}
\sP_1(\mu_{t,1}^X<p_1\mid\gH)+e_{1,t}(p_1;\lambda,q_0,\alpha).
\end{align*}
\end{proof}
}

For every binary trajectory,
\begin{equation}
\label{eq:confidence-score-range}
0\le\widehat U^X(Z_\infty)\le1,
\qquad
0\le\widehat L^X(Z_\infty)\le1.
\end{equation}
Moreover, for every $t\in\gW$,
\begin{equation}
\label{eq:score-observed-average}
\widehat L^X(Z_\infty)\le\widehat\mu_t^X
\le\widehat U^X(Z_\infty).
\end{equation}
These bounds follow because each $\widehat\mu_t^X$ is an average of binary
values and $\widehat U^X$, $\widehat L^X$ are their temporal extrema.

Write $S_{t,\star}:=\sum_{s=1}^t X_{s,\star}(\widehat{\gS}(\mD_{\mathrm{in}}))$ and {$\gB_0:=\{\widehat\Delta(\mD_{\mathrm{in}})>0\}$}.

\begin{remark}[Fallback betting parameters]
\label{rem:learned-domain}
On ${\gB_0^c}$, the parameters $(\widehat p,\widehat\lambda)=(1,1)$ give
\begin{equation}
\label{eq:fallback-evidence}
E_t(\widehat\lambda,{\widehat p})
\eqby{eq:monitor-evidence}
\exp\{S_{t,\star}-t\}
\stackrel{S_{t,\star}\le t}{\le}1
\stackrel{0<\alpha<1}{<}1/\alpha.
\end{equation}
Thus $\widehat\tau=\infty$ on every such trajectory.
On ${\gB_0}$, the bounds in \eqref{eq:confidence-score-range} and
{the definition of $\gB_0$ place}
the midpoint $\widehat p$ in $(0,1)$, even when a learned level
equals zero or one. {The chosen stake in
\eqref{eq:learned-betting-parameters} is finite and positive.}
\end{remark}
                          % Appendix G
\section{Proof of Proposition~\ref{prop:pathwise-calibration}}
\label{app:proof-pathwise-calibration}

\begin{proof}
For part~\textnormal{(1)}, suppose $\widehat U^X\le p_0$.
For every $t\in\gW$,
\begin{align}
\log E_t(\lambda,{p_0})
&\leby{eq:confidence-evidence-bounds}
t\lambda(\widehat\mu_t^X-p_0)
\leby{eq:score-observed-average}
t\lambda(\widehat U^X-p_0)
\le0<\log\frac1\alpha.
\label{eq:pathwise-null-evidence}
\end{align}
Thus \eqref{eq:intervention-time} gives $\tau(\lambda,{p_0})=\infty$.

Now suppose $\widehat L^X\ge p_0+{(v(p_0)\lambda+\lambda^{-1}a_{T_1}(\alpha)^2)/2}$.
Then
\begin{align}
\log E_{T_1}(\lambda,{p_0})
&\geby{eq:confidence-evidence-bounds}
T_1\left\{\lambda(\widehat\mu_{T_1}^X-p_0)-{(v(p_0)\lambda^2)/2}\right\}
\nonumber\\
&\geby{eq:score-observed-average}
T_1\left\{\lambda(\widehat L^X-p_0)-{(v(p_0)\lambda^2)/2}\right\}
\nonumber\\
&\ge {T_1a_{T_1}(\alpha)^2/2}
=\log\frac1\alpha.
\label{eq:pathwise-alternative-evidence}
\end{align}
Since $T_1\in\gW$, \eqref{eq:intervention-time} gives $\tau(\lambda,{p_0})\le T_1$.

{For part~\textnormal{(2)}, use $S_t$ and $A_{y,t}$ from
\eqref{eq:class-trajectory-conditional-mean} with the stated filtration.
For $\gF_0$-measurable $z$, put
\begin{equation}
\label{eq:monitor-hoeffding-process}
N_t(z):=\exp\{z(S_t-A_{y,t})-tz^2/8\},\qquad N_0(z)=1,
\end{equation}
so $0<N_t(z)\le\exp\{t(|z|-z^2/8)\}\le e^{2t}$.
By \eqref{eq:class-trajectory-conditional-mean},
\begin{align}
\E_y[N_t(z)\mid\gF_{t-1}]
&=N_{t-1}(z)e^{-z^2/8}
\E_y[e^{z(X_t-\E_y[X_t\mid\gF_{t-1}])}\mid\gF_{t-1}]
\nonumber\\
&\leby{eq:conditional-hoeffding}N_{t-1}(z).
\label{eq:monitor-hoeffding-step}
\end{align}
For a positive, $\gF_0$-measurable $a$ and either fixed sign, if
$\pm(\widehat\mu_t^X-\mu_{t,y}^X)>(1+T_0/t)a/2$ at some $t\in\gW$, then
\[
N_t(\pm4a)>\exp\{2(t+T_0)a^2-2ta^2\}=e^{2T_0a^2}.
\]
Applying \eqref{eq:ville-conditional-martingale} gives
\begin{equation}
\label{eq:uniform-monitor-deviation}
\sP_y\!\left(\exists t\in\gW:\pm(\widehat\mu_t^X-\mu_{t,y}^X)>(1+T_0/t)a/2
\,\middle|\,\gF_0\right)\le e^{-2T_0a^2}.
\end{equation}
For each fixed $t$, \eqref{eq:monitor-lower-tail} gives
\begin{equation}
\label{eq:fixed-time-monitor-deviation}
\sP_y\!\left(\pm(\widehat\mu_t^X-\mu_{t,y}^X)>a
\,\middle|\,\gF_0\right)\le e^{-2ta^2}.
\end{equation}
Use $a=a_{T_0}(\beta/2)/2$ in \eqref{eq:uniform-monitor-deviation} and $a=a_t(\beta/(2|\gW|))/2$ in \eqref{eq:fixed-time-monitor-deviation}.
Exceeding their minimum requires exceeding at least one of them, so a conditional union bound gives
\begin{align*}
&\sP_y\!\left(\exists t\in\gW:
\pm(\widehat\mu_t^X-\mu_{t,y}^X)>r_t(\beta)
\,\middle|\,\gF_0\right)\\
&\leby{eq:uniform-monitor-deviation}
\frac{\beta}{2}+\sum_{t\in\gW}
\sP_y\!\left(\pm(\widehat\mu_t^X-\mu_{t,y}^X)>
\frac{a_t(\beta/(2|\gW|))}{2}\,\middle|\,\gF_0\right)\\
&\leby{eq:fixed-time-monitor-deviation}
\frac{\beta}{2}+\sum_{t\in\gW}\frac{\beta}{2|\gW|}=\beta.
\end{align*}
}
\end{proof}
              % Appendix H
\section{Proof of Proposition~\ref{prop:statistical-guarantees}}
\label{app:proof-statistical-guarantees}

{\begin{proof}
Take $\gH$ to be the trivial $\sigma$-algebra in Lemma~\ref{lem:conditional-intervention-bounds}.
For part~(1),
\begin{align}
\sP_0(\tau(\lambda,q_0)<\infty)
&\leby{eq:conditional-false-intervention}
\sP_0(\max_{s\in\gW}\mu_{s,0}^X>p_0)+e_0(p_0;\lambda,q_0,\alpha)
\nonumber\\
&\le\epsilon_0+e_0(p_0;\lambda,q_0,\alpha).
\label{eq:false-intervention-conclusion}
\end{align}
For part~(2), let
\begin{equation}
\label{eq:alternative-good-events}
G_1:=\{\min_{s\in\gW}\mu_{s,1}^X\ge p_1\},
\qquad \sP_1(G_1^c)\le\epsilon_1.
\end{equation}
\begin{align}
\sP_1(\tau(\lambda,q_0)>t)
&\leby{eq:conditional-missed-detection}
\sP_1(\mu_{t,1}^X<p_1)+e_{1,t}(p_1;\lambda,q_0,\alpha)
\nonumber\\
&\leby{eq:alternative-good-events}
\sP_1(G_1^c)+e_{1,t}(p_1;\lambda,q_0,\alpha)
\nonumber\\
&\le\epsilon_1+e_{1,t}(p_1;\lambda,q_0,\alpha).
\label{eq:detection-conclusion}
\end{align}
\end{proof}
}
             % Appendix I
% \input{appendix/F-proof-confidence-sequences} % Temporarily excluded
% \input{appendix/G-empirical-separation-milp} % Temporarily excluded
\section{Proof of Theorem~\ref{thm:learned-intervention-guarantees}}
\label{app:proof-learned-intervention-guarantees}
The proof has four steps:
\squishenum
\csname @beginparpenalty\endcsname=10000\relax
\item[\textnormal{(1)}]
Use exchangeability to express the conditional error on the next trajectory
as the conditional mean of the extension error.
Each extension indicator has the same conditional mean, so this step
does not require conditional independence.
\item[\textnormal{(2)}]
On the pairwise stability event, compare the values after a training exchange
with those at $\mD_{\mathrm{in}}$. Order the class-$y$ trajectories by decreasing
{${S_{y,i,i}(\mD;\mI_{\mathrm{in}})}$} and use Lemma~\ref{lem:uniform-subset-rank}
and finite averaging to control the tail of the extension error.
{For balanced extensions, count training and extension exceptions under their shared budget and control all prefixes.}
\item[\textnormal{(3)}]
{Apply the positive-part variational formula and conditional Jensen's inequality, accounting for stability failure, to prove item~\textnormal{(3)}. The exact rank law includes the single-extension case; tail integration gives a closed-form upper bound.}
\item[\textnormal{(4)}]
{Use Proposition~\ref{prop:pathwise-calibration}(2) to bound the training conditional means with leave-one-out observed averages. Transfer their order statistics to the next trajectory and apply Proposition~\ref{prop:statistical-guarantees}.}
\squishenumend

{For each $y$, work under $\sP_y$ on the joint sample space of $\mD$.
Use the conditional-mean versions in Lemma~\ref{lem:conditional-mean-equivariance}
and the joint law $\sP_{y,\pi}$.
}

\subsection{Conditional Error}
\begin{lemma}[Conditional error]
\label{lem:conditional-error-identity}
For each $y$, almost surely under $\sP_y$,
{\begin{equation}
\label{eq:conditional-future-error}
\sP_y(\gV_{y,\star}(\mD;\mI_{\mathrm{in}})\mid\mD_{\mathrm{in}})
=\E_y\left[
\frac{C_y(\mD;\mI_{\mathrm{in}})}{m_y}
\,\middle|\,\mD_{\mathrm{in}}\right].
\end{equation}
}
\end{lemma}

\begin{proof}
{For $i\in\gI_{\mathrm{out},y}$, let $\pi$ exchange $i$ and $\star$, so $\pi(\mI_{\mathrm{in}})=\mI_{\mathrm{in}}$.}
Assumption~\ref{assump:finite-exchangeable-extension} gives, under $\sP_y$,
\begin{equation}
\label{eq:extension-next-exchangeability}
(\mD_{\mathrm{in}},\omega_i)\overset{d}{=}
(\mD_{\mathrm{in}},\omega_\star).
\end{equation}
{Since $S_{y,i,i}(\mD;\mI_{\mathrm{in}})$ is $\sigma(\mD_{\mathrm{in}},\omega_i)$-measurable, \eqref{eq:extension-next-exchangeability} gives
\begin{align}
(\mD_{\mathrm{in}},S_{y,i,i}(\mD;\mI_{\mathrm{in}}))
&\overset{d}{=}(\mD_{\mathrm{in}},S_{y,i,i}(\pi(\mD);\mI_{\mathrm{in}}))
\nonumber\\
&\eqby{eq:conditional-mean-equivariance}
(\mD_{\mathrm{in}},S_{y,\star,\star}(\mD;\mI_{\mathrm{in}})).
\label{eq:extension-next-conditional-mean-law}
\end{align}}
The training order statistic is $\sigma(\mD_{\mathrm{in}})$-measurable, so
{\begin{align*}
\E_y\left[\frac{C_y(\mD;\mI_{\mathrm{in}})}{m_y}
\,\middle|\,\mD_{\mathrm{in}}\right]
&\eqby{eq:symmetrized-future-error}
\E_y\left[
\frac1{m_y}\sum_{i\in\gI_{\mathrm{out},y}}\1\{\gV_{y,i}(\mD;\mI_{\mathrm{in}})\}
\,\middle|\,\mD_{\mathrm{in}}\right]
\\
&=
\frac1{m_y}\sum_{i\in\gI_{\mathrm{out},y}}
\sP_y(\gV_{y,i}(\mD;\mI_{\mathrm{in}})\mid\mD_{\mathrm{in}})
\\
&\eqby{eq:extension-next-conditional-mean-law}
\frac1{m_y}\sum_{i\in\gI_{\mathrm{out},y}}
\sP_y(\gV_{y,\star}(\mD;\mI_{\mathrm{in}})\mid\mD_{\mathrm{in}})
\\
&=
\sP_y(\gV_{y,\star}(\mD;\mI_{\mathrm{in}})\mid\mD_{\mathrm{in}}).
\end{align*}
}
\end{proof}

\subsection{Extension Error under Stability}

\begin{lemma}[Extension error under stability]
\label{lem:stable-extension-error}
For each $y$ and $s>0$,
{\begin{equation}
\label{eq:hypergeometric-error-tail}
\sP_y\left(
\frac{C_y(\mD;\mI_{\mathrm{in}})}{m_y}
>\epsilon_y+\theta_y\frac{(n_y+m_y)^2}{n_ym_y}+s
\right)
\le {e^{-2n_ym_ys^2/(n_y+m_y+2)}+\frac{\kappa_m}{\sP(Y_\star=y)}}.
\end{equation}
}
\end{lemma}

The proof has three steps:
\squishenum
\item[\textnormal{(1)}]
Use finite exchangeability to average over training subsets in both classes
while fixing the complete sample.
\item[\textnormal{(2)}]
Order the class-$y$ trajectories by decreasing {${S_{y,i,i}(\mD;\mI_{\mathrm{in}})}$}
and compare each array with the values at $\mD_{\mathrm{in}}$.
Stability and the training rank budget then bound the future error count
by $J_{\gK}^a$ from Lemma~\ref{lem:uniform-subset-rank} plus the number of exceptional
trajectories, even though the fitted threshold depends on the training subset.
\item[\textnormal{(3)}]
Apply the tail bound for $J_{\gK}^a$ and average over the complete sample,
adding the average stability failure probability.
\squishenumend

We use \eqref{eq:class-permutation-average} to average over training samples.

\subsubsection{Stability and Rank Domination}

Without loss of generality, we can again write
$\mD=(1,\ldots,n+m+1)$ using occurrence labels.
Fix this complete sample. For this fixed-sample calculation, number the
class-$y$ training and extension occurrences as $1,\ldots,n_y+m_y$ in decreasing order of
their values at the fixed reference training sample $\mD_{\mathrm{in}}$:
{\begin{equation}
\label{eq:fixed-monitor-score-order}
{S_{y,1,1}(\mD;\mI_{\mathrm{in}})}\ge\cdots\ge
{S_{y,n_y+m_y,n_y+m_y}(\mD;\mI_{\mathrm{in}})}.
\end{equation}
}
Break ties by original indices. Carry $\mI_{\mathrm{in}}$ and the class index sets along
with this relabeling: $\mD_{\mI_{\mathrm{in}}}=\mD_{\mathrm{in}}$ still denotes the original
training trajectories, and $\mD_{\pi(\mI_{\mathrm{in}})}$ denotes their images
under $\pi$. Uniform averaging over $\pi$ is unchanged.
The class-$y$ training indices after permutation are
\begin{equation}
\label{eq:permuted-training-ranks}
\begin{gathered}
\gK_\pi:=\pi(\gI_{\mathrm{in},y})=\{i_1,\ldots,i_{n_y}\},
\quad 1\le i_1<\cdots<i_{n_y}\le n_y+m_y,
\\
J_{\gK_\pi}^a:=i_{a+1}-a-1,\quad 0\le a<n_y.
\end{gathered}
\end{equation}
The corresponding extension indices are $[n_y+m_y]\setminus\gK_\pi$,
so $J_{\gK_\pi}^a$ counts those preceding the $(a+1)$st training occurrence.
{For each permutation of the other class, each $n_y$-subset of ranks arises from $n_y!m_y!$ class-$y$ permutations.} Thus, for
$K\subseteq[n_y+m_y]$ with $|K|=n_y$,
\begin{equation}
\label{eq:uniform-training-ranks}
\sQ_\pi(\gK_\pi=K)
=\frac{n_y!m_y!}{(n_y+m_y)!}
=\binom{n_y+m_y}{n_y}^{-1}.
\end{equation}
Lemma~\ref{lem:uniform-subset-rank} therefore applies.
The fitted level {${\left(S_{y,i,i}(\mD;\pi(\mI_{\mathrm{in}}))\right)_{(k_y:\gK_\pi)}}$} may still depend on $\pi$.

\begin{lemma}[Rank domination under stability]
\label{lem:stable-rank-domination}
If $\lfloor(n_y+m_y)\theta_y\rfloor<k_y$, then for every $\pi$,
{\begin{equation}
\label{eq:future-error-rank-domination}
C_y(\mD;\pi(\mI_{\mathrm{in}}))\le
J_{\gK_\pi}^{n_y-k_y+\lfloor(n_y+m_y)\theta_y\rfloor}
+\lfloor(n_y+m_y)\theta_y\rfloor
\qquad\text{on }\gE_{\pi(\mI_{\mathrm{in}}),\mI_{\mathrm{in}}}^m.
\end{equation}
}
\end{lemma}

\begin{proof}
Fix $\pi$. The class-$y$ training indices are $\gK_\pi$, so
{\begin{equation}
\label{eq:symmetrized-training-budget}
\sum_{i\in\gK_\pi}
\1\{{S_{y,i,i}(\mD;\pi(\mI_{\mathrm{in}}))}>{\left(S_{y,i,i}(\mD;\pi(\mI_{\mathrm{in}}))\right)_{(k_y:\gK_\pi)}}\}
\leby{eq:order-statistic-budget-equivalence}n_y-k_y.
\end{equation}
}
On {$\gE_{\pi(\mI_{\mathrm{in}}),\mI_{\mathrm{in}}}^m$},
{\begin{equation}
\label{eq:comparison-monitor-stability}
\sum_{i\in[n_y+m_y]}
\1\!\left\{\left|
{S_{y,i,i}(\mD;\pi(\mI_{\mathrm{in}}))}-{S_{y,i,i}(\mD;\mI_{\mathrm{in}})}
\right|>{B_m}\right\}
\leby{eq:score-swap-stability}\lfloor(n_y+m_y)\theta_y\rfloor.
\end{equation}
}
For $i\in[n_y+m_y]$ with {$|{S_{y,i,i}(\mD;\pi(\mI_{\mathrm{in}}))}-{S_{y,i,i}(\mD;\mI_{\mathrm{in}})}|\le {B_m}$},
{\begin{equation}
\label{eq:swap-score-comparison}
\begin{aligned}
&{S_{y,i,i}(\mD;\pi(\mI_{\mathrm{in}}))}>
{\left(S_{y,i,i}(\mD;\pi(\mI_{\mathrm{in}}))\right)_{(k_y:\gK_\pi)}}+2{B_m}\\
&\quad\Longrightarrow\ {S_{y,i,i}(\mD;\mI_{\mathrm{in}})}>
{\left(S_{y,i,i}(\mD;\pi(\mI_{\mathrm{in}}))\right)_{(k_y:\gK_\pi)}}+{B_m}\\
&\quad\Longrightarrow\ {S_{y,i,i}(\mD;\pi(\mI_{\mathrm{in}}))}>
{\left(S_{y,i,i}(\mD;\pi(\mI_{\mathrm{in}}))\right)_{(k_y:\gK_\pi)}}.
\end{aligned}
\end{equation}
}
Both comparisons use the level {${\left(S_{y,i,i}(\mD;\pi(\mI_{\mathrm{in}}))\right)_{(k_y:\gK_\pi)}}$} from the same training sample.
No stability of fitted levels is used.

On {$\gE_{\pi(\mI_{\mathrm{in}}),\mI_{\mathrm{in}}}^m$}, the exceptional count in
\eqref{eq:comparison-monitor-stability} gives
{\begin{align}
&\sum_{i\in\gK_\pi}
\1\{{S_{y,i,i}(\mD;\mI_{\mathrm{in}})}>{\left(S_{y,i,i}(\mD;\pi(\mI_{\mathrm{in}}))\right)_{(k_y:\gK_\pi)}}+{B_m}\}
\nonumber\\
&\leby{eq:swap-score-comparison}
\sum_{i\in\gK_\pi}
\1\{{S_{y,i,i}(\mD;\pi(\mI_{\mathrm{in}}))}>{\left(S_{y,i,i}(\mD;\pi(\mI_{\mathrm{in}}))\right)_{(k_y:\gK_\pi)}}\}
+\lfloor(n_y+m_y)\theta_y\rfloor
\nonumber\\
&\leby{eq:symmetrized-training-budget}
n_y-k_y+\lfloor(n_y+m_y)\theta_y\rfloor.
\label{eq:comparison-training-budget}
\end{align}
}
Put $a:=n_y-k_y+\lfloor(n_y+m_y)\theta_y\rfloor<n_y$.
By \eqref{eq:comparison-training-budget}, at most $a$ training values
exceed {${\left(S_{y,i,i}(\mD;\pi(\mI_{\mathrm{in}}))\right)_{(k_y:\gK_\pi)}}+{B_m}$}.
Every reference value exceeding this threshold therefore precedes
the $(a+1)$st training occurrence in \eqref{eq:fixed-monitor-score-order}.
Consequently,
{\begin{equation}
\label{eq:fixed-monitor-rank-count}
\sum_{i\in[n_y+m_y]\setminus\gK_\pi}
\1\{{S_{y,i,i}(\mD;\mI_{\mathrm{in}})}>{\left(S_{y,i,i}(\mD;\pi(\mI_{\mathrm{in}}))\right)_{(k_y:\gK_\pi)}}+{B_m}\}
\leby{eq:permuted-training-ranks}
J_{\gK_\pi}^a.
\end{equation}
}
Using \eqref{eq:swap-score-comparison} and the exceptional count in \eqref{eq:comparison-monitor-stability} again,
{\begin{align*}
&C_y(\mD;\pi(\mI_{\mathrm{in}}))\\
&\eqby{eq:reindexed-extension-error}
\sum_{i\in[n_y+m_y]\setminus\gK_\pi}
\1\{{S_{y,i,i}(\mD;\pi(\mI_{\mathrm{in}}))}>{\left(S_{y,i,i}(\mD;\pi(\mI_{\mathrm{in}}))\right)_{(k_y:\gK_\pi)}}+2{B_m}\}\\
&\le
\sum_{i\in[n_y+m_y]\setminus\gK_\pi}
\1\{{S_{y,i,i}(\mD;\mI_{\mathrm{in}})}>{\bigl(S_{y,i,i}(\mD;\pi(\mI_{\mathrm{in}}))\bigr)_{(k_y:\gK_\pi)}}+{B_m}\}
+\lfloor(n_y+m_y)\theta_y\rfloor
\\
&\leby{eq:fixed-monitor-rank-count}
J_{\gK_\pi}^{a}
+\lfloor(n_y+m_y)\theta_y\rfloor.
\end{align*}
}
Strict exceedances preserve the bound at ties.
\end{proof}

\subsubsection{Proof of Lemma~\ref{lem:stable-extension-error}}

For each fixed $\mD$, \eqref{eq:future-error-rank-domination} and
\eqref{eq:uniform-rank-tail} bound the intersection of the extension tail event
and the pairwise stability event.
Lemma~\ref{lem:class-permutation-average} and \eqref{eq:class-conditional-stability-failure}
then give the result under $\sP_y$.

\begin{proof}
Fix $\mD$ and $s>0$.
On {$\gE_{\pi(\mI_{\mathrm{in}}),\mI_{\mathrm{in}}}^m$}, if $\lfloor(n_y+m_y)\theta_y\rfloor<k_y$, then
{\begin{align}
\frac{C_y(\mD;\pi(\mI_{\mathrm{in}}))}{m_y}&\leby{eq:future-error-rank-domination}
\frac{J_{\gK_\pi}^{n_y-k_y+\lfloor(n_y+m_y)\theta_y\rfloor}}{m_y}
+\frac{\lfloor(n_y+m_y)\theta_y\rfloor}{m_y},
\nonumber\\
\frac{n_y-k_y+\lfloor(n_y+m_y)\theta_y\rfloor}{n_y}
+\frac{\lfloor(n_y+m_y)\theta_y\rfloor}{m_y}
&\le
\epsilon_y+\frac{\lfloor(n_y+m_y)\theta_y\rfloor}{n_y}
+\frac{\lfloor(n_y+m_y)\theta_y\rfloor}{m_y}
\nonumber\\
&\le\epsilon_y+\theta_y\frac{(n_y+m_y)^2}{n_ym_y}.
\label{eq:rank-error-offset}
\end{align}
}
Using the uniform law in \eqref{eq:uniform-training-ranks}, for every fixed
$\mD$ and $s>0$,
{\begin{align}
&\sQ_\pi\left(\frac{C_y(\mD;\pi(\mI_{\mathrm{in}}))}{m_y}>\epsilon_y+\theta_y\frac{(n_y+m_y)^2}{n_ym_y}+s,\
\gE_{\pi(\mI_{\mathrm{in}}),\mI_{\mathrm{in}}}^m\right)
\nonumber\\
&\leby{eq:rank-error-offset}
\sQ_\pi\left(
J_{\gK_\pi}^{n_y-k_y+\lfloor(n_y+m_y)\theta_y\rfloor}>
m_y\left(\frac{n_y-k_y+\lfloor(n_y+m_y)\theta_y\rfloor}{n_y}+s\right)
\right)
\nonumber\\
&\leby{eq:uniform-rank-tail}{e^{-2n_ym_ys^2/(n_y+m_y+2)}}.
\label{eq:fixed-sample-stable-tail}
\end{align}
}
If $\lfloor(n_y+m_y)\theta_y\rfloor\ge k_y$, then
\[
1\le\frac{n_y-k_y+\lfloor(n_y+m_y)\theta_y\rfloor}{n_y}
\le\epsilon_y+\frac{\lfloor(n_y+m_y)\theta_y\rfloor}{n_y}
\le\epsilon_y+\theta_y\frac{(n_y+m_y)^2}{n_ym_y}.
\]
Since {$C_y(\mD;\pi(\mI_{\mathrm{in}}))\le m_y$}, the event on the left of
\eqref{eq:fixed-sample-stable-tail} is empty, so its final bound also holds.
{Taking expectations and using \eqref{eq:class-permutation-average},
\begin{align*}
&\sP_y\!\left(\frac{C_y(\mD;\mI_{\mathrm{in}})}{m_y}>
\epsilon_y+\theta_y\frac{(n_y+m_y)^2}{n_ym_y}+s\right)\\
&\leby{eq:fixed-sample-stable-tail}
e^{-2n_ym_ys^2/(n_y+m_y+2)}
+\sP_{y,\pi}((\gE_{\pi(\mI_{\mathrm{in}}),\mI_{\mathrm{in}}}^m)^c)
\leby{eq:class-conditional-stability-failure}
e^{-2n_ym_ys^2/(n_y+m_y+2)}+\frac{\kappa_m}{\sP(Y_\star=y)}.
\end{align*}}
\end{proof}

{\subsection{Balanced Extensions and Selected Quantiles}
\label{app:balanced-selected-thresholds}

In this subsection, let $m_y=n_y$ and allow the quantile rank to depend on the training data. Let $k_y(\mD_{\mI})\in[n_y]$ be measurable and permutation-invariant, and use it in $\gV_{y,i}(\mD;\mI)$ and $C_y(\mD;\mI)$. For $\mI=(i_1,\ldots,i_n)$, put
\begin{equation}
\label{eq:selected-training-error}
\widehat\epsilon_y(\mD;\mI):=\frac1{n_y}\sum_{r\in[n]:\,Y_{i_r}=y}
\1\!\left\{{S_{y,i_r,i_r}(\mD;\mI)}>
{\left(S_{y,i,i}(\mD;\mI)\right)_{(k_y(\mD_{\mI}):(\mD_{\mI})_y)}}\right\}.
\end{equation}
The sum is over training occurrences, including repetitions. The original fixed rank gives $\widehat\epsilon_y(\mD;\mI)\le\epsilon_y$.

\begin{lemma}[Balanced extension error]
\label{lem:balanced-extension-error}
For $s>0$,
\begin{equation}
\label{eq:balanced-extension-error}
\sP_y\!\left(\frac{C_y(\mD;\mI_{\mathrm{in}})}{n_y}>
\widehat\epsilon_y(\mD;\mI_{\mathrm{in}})+2\theta_y+s\right)
\le e^{-2n_y^2s^2/(2n_y+1)}+\frac{\kappa_m}{\sP(Y_\star=y)}.
\end{equation}
\end{lemma}

\begin{proof}
Fix the complete sample and use its ordering in \eqref{eq:fixed-monitor-score-order}.
For a training permutation $\pi$, abbreviate its fitted threshold by
$c_\pi:={\left(S_{y,i,i}(\mD;\pi(\mI_{\mathrm{in}}))\right)_{(k_y(\mD_{\pi(\mI_{\mathrm{in}})}):\gK_\pi)}}$.
The indices with ${S_{y,i,i}(\mD;\mI_{\mathrm{in}})}>c_\pi+{B_m}$ form a prefix, say $[j]$.
Let $b_{\mathrm{in}}$ and $b_{\mathrm{out}}$ count the exceptional training and extension occurrences in \eqref{eq:comparison-monitor-stability}.
On $\gE_{\pi(\mI_{\mathrm{in}}),\mI_{\mathrm{in}}}^m$, the comparisons in \eqref{eq:swap-score-comparison} give
\begin{align*}
|\gK_\pi\cap[j]|\le n_y\widehat\epsilon_y(\mD;\pi(\mI_{\mathrm{in}}))+b_{\mathrm{in}},\
C_y(\mD;\pi(\mI_{\mathrm{in}}))\le|\gK_\pi^c\cap[j]|+b_{\mathrm{out}},\
b_{\mathrm{in}}+b_{\mathrm{out}}\le b_y.
\end{align*}
Consequently,
\begin{equation}
\label{eq:shared-exception-budget}
\frac{C_y(\mD;\pi(\mI_{\mathrm{in}}))}{n_y}
-\widehat\epsilon_y(\mD;\pi(\mI_{\mathrm{in}}))
\le D_{\gK_\pi}+\frac{b_y}{n_y}
\le D_{\gK_\pi}+2\theta_y.
\end{equation}
Apply Lemma~\ref{lem:balanced-prefix-difference} and finite averaging as in Lemma~\ref{lem:class-permutation-average}, now to the difference on the left of \eqref{eq:shared-exception-budget}.
Add the failure probability in \eqref{eq:class-conditional-stability-failure}.
The threshold may depend on the training sample because $D_{\gK_\pi}$ controls every prefix.
Strict exceedances make the same argument valid at ties.
\end{proof}

For ranks selected from the training data, put
\begin{equation}
\label{eq:balanced-selected-error}
\delta_y^{\mathrm{bal}}(\mD_{\mathrm{in}}):=
\widehat\epsilon_y(\mD;\mI_{\mathrm{in}})+2\theta_y
+\frac{\kappa_m}{\beta\sP(Y_\star=y)}
+\sqrt{\frac{2n_y+1}{2n_y^2}\log\frac e\beta}.
\end{equation}
Then
\begin{equation}
\label{eq:balanced-selected-coverage}
\sP_y\!\left(\sP_y(\gV_{y,\star}(\mD;\mI_{\mathrm{in}})\mid\mD_{\mathrm{in}})
>\delta_y^{\mathrm{bal}}(\mD_{\mathrm{in}})\right)\le\beta.
\end{equation}
The selector, including any rank-dependent rule selection, must satisfy the same stability assumption. Formula~\eqref{eq:total-statistical-errors} remains a fixed-rank guarantee. The training values in this subsection are latent; the intervention guarantees below additionally estimate them.

\begin{proof}
The conditional-error identity still holds because the selected threshold is training-measurable.
Put $c:=2n_y^2/(2n_y+1)$ and
$R:=[C_y(\mD;\pi(\mI_{\mathrm{in}}))/n_y-\widehat\epsilon_y(\mD;\pi(\mI_{\mathrm{in}}))-2\theta_y]_+\1\{\gE_{\pi(\mI_{\mathrm{in}}),\mI_{\mathrm{in}}}^m\}$.
For any event $A\in\sigma(\mD_{\mathrm{in}})$ with $q:=\sP_y(A)>0$, write $A_\pi:=\pi^{-1}A$.
By \eqref{eq:training-index-exchangeability}, $\sP_{y,\pi}(A_\pi)=q$.
Lemma~\ref{lem:balanced-prefix-difference} and \eqref{eq:shared-exception-budget} give
$\sP_{y,\pi}(R>s)\le e^{-cs^2}$.
Apply \eqref{eq:weighted-tail-bound} with $h=\1\{A_\pi\}$:
\begin{equation}
\E^{\sP_{y,\pi}}[R\mid A_\pi]
=\frac{\E^{\sP_{y,\pi}}[\1\{A_\pi\}R]}q
\leby{eq:weighted-tail-bound}\sqrt{\frac{\log(e/q)}c}.
\label{eq:balanced-selected-moment}
\end{equation}
Write $\rho_y:=\sP_y(\gV_{y,\star}(\mD;\mI_{\mathrm{in}})\mid\mD_{\mathrm{in}})$ and suppress the arguments of $\widehat\epsilon_y$.
Since $A,\widehat\epsilon_y$ are training-measurable,
\begin{align}
\E_y[\rho_y-\widehat\epsilon_y\mid A]
&\eqby{eq:conditional-future-error}
\E_y[C_y(\mD;\mI_{\mathrm{in}})/n_y-\widehat\epsilon_y\mid A]
\nonumber\\
&\leby{eq:class-permutation-average}
2\theta_y+\E^{\sP_{y,\pi}}[R\mid A_\pi]
+\frac{\sP_{y,\pi}((\gE_{\pi(\mI_{\mathrm{in}}),\mI_{\mathrm{in}}}^m)^c)}q
\nonumber\\
&\leby{eq:balanced-selected-moment}
2\theta_y+\sqrt{\frac{\log(e/q)}c}
+\frac{\sP_{y,\pi}((\gE_{\pi(\mI_{\mathrm{in}}),\mI_{\mathrm{in}}}^m)^c)}q
\nonumber\\
&\leby{eq:class-conditional-stability-failure}
2\theta_y+\sqrt{\frac{\log(e/q)}c}+\frac{\kappa_m}{q\sP(Y_\star=y)}.
\label{eq:balanced-selected-conditional-moment}
\end{align}
Take $A:=\{\rho_y>\delta_y^{\mathrm{bal}}(\mD_{\mathrm{in}})\}$.
If $q>\beta$, then
\begin{align*}
0
&<\E_y[\rho_y-\delta_y^{\mathrm{bal}}(\mD_{\mathrm{in}})\mid A]\\
&\eqby{eq:balanced-selected-error}
\E_y[\rho_y-\widehat\epsilon_y\mid A]-2\theta_y
-\frac{\kappa_m}{\beta\sP(Y_\star=y)}-\sqrt{\frac{\log(e/\beta)}c}\\
&\leby{eq:balanced-selected-conditional-moment}
\sqrt{\frac{\log(e/q)}c}-\sqrt{\frac{\log(e/\beta)}c}
+\frac{\kappa_m}{\sP(Y_\star=y)}\left(\frac1q-\frac1\beta\right)
\le0.
\end{align*}
Consequently,
\[
\sP_y\!\left(\sP_y(\gV_{y,\star}(\mD;\mI_{\mathrm{in}})\mid\mD_{\mathrm{in}})
>\delta_y^{\mathrm{bal}}(\mD_{\mathrm{in}})\right)
=\sP_y(A)=q\le\beta.
\]
\end{proof}

\subsection{Training-Conditional Coverage}

We now prove item~\textnormal{(3)}:
\begin{equation}
\label{eq:training-conditional-generalization}
\sP_y\left(\sP_y(\gV_{y,\star}(\mD;\mI_{\mathrm{in}})\mid\mD_{\mathrm{in}})>\delta_y(\beta)\right)\le\beta.
\end{equation}
We use the CVaR formula of \citet[Theorem~10, Eqs.~(27)--(28)]{rockafellar_conditional_2002}, with upper-tail mass $\beta$, and account separately for stability failure.

\begin{proof}
Put $H:=C_y(\mD;\mI_{\mathrm{in}})/m_y$ and $\rho_y:=\E_y[H\mid\mD_{\mathrm{in}}]$.
Write $W_y(\gK_\pi)$ for $W_y$ evaluated at the training ranks $\gK_\pi$.
For $0\le u\le1$, \eqref{eq:future-error-rank-domination} and $H\le1$ give, including $b_y\ge k_y$ where $W_y=1$,
\begin{equation}
\label{eq:stable-rank-positive-part}
\left(\frac{C_y(\mD;\pi(\mI_{\mathrm{in}}))}{m_y}-u\right)_+
\le (W_y(\gK_\pi)-u)_+
+(1-u)\1\{(\gE_{\pi(\mI_{\mathrm{in}}),\mI_{\mathrm{in}}}^m)^c\}.
\end{equation}
Conditional Jensen's inequality and finite averaging now give
\begin{align}
\label{eq:conditional-rank-positive-part}
\E_y[(\rho_y-u)_+]
&\stackrel{\mathrm{Jensen}}{\le}\E_y[(H-u)_+]
\eqby{eq:class-permutation-average}
\E^{\sP_{y,\pi}}\!\left[\left(\frac{C_y(\mD;\pi(\mI_{\mathrm{in}}))}{m_y}-u\right)_+\right]
\nonumber\\
&\leby{eq:stable-rank-positive-part}
\E^{\sP_{y,\pi}}[(W_y(\gK_\pi)-u)_+]
+(1-u)\sP_{y,\pi}((\gE_{\pi(\mI_{\mathrm{in}}),\mI_{\mathrm{in}}}^m)^c)
\nonumber\\
&\eqby{eq:uniform-training-ranks}
\E^{\sQ_{n_y,m_y}}[(W_y-u)_+]
+(1-u)\sP_{y,\pi}((\gE_{\pi(\mI_{\mathrm{in}}),\mI_{\mathrm{in}}}^m)^c)
\nonumber\\
&\leby{eq:class-conditional-stability-failure}
\E^{\sQ_{n_y,m_y}}[(W_y-u)_+]+(1-u)\frac{\kappa_m}{\sP(Y_\star=y)}.
\end{align}
The objective in \eqref{eq:total-statistical-errors} is continuous on $[0,1]$, so choose a minimizer $u$.
If $\delta_y(\beta)>u$, Markov's inequality gives
\begin{align*}
\sP_y(\rho_y>\delta_y(\beta))
&\le\frac{\E_y[(\rho_y-u)_+]}{\delta_y(\beta)-u}
\leby{eq:conditional-rank-positive-part}
\frac{\E^{\sQ_{n_y,m_y}}[(W_y-u)_+]+(1-u)\kappa_m/\sP(Y_\star=y)}
{\delta_y(\beta)-u}
\eqby{eq:total-statistical-errors}\beta.
\end{align*}
If $\delta_y(\beta)=u$, then
\[
\E_y[(\rho_y-u)_+]
\leby{eq:conditional-rank-positive-part}
\E^{\sQ_{n_y,m_y}}[(W_y-u)_+]+(1-u)\frac{\kappa_m}{\sP(Y_\star=y)}
\eqby{eq:total-statistical-errors}
\beta(\delta_y(\beta)-u)=0,
\]
so $\sP_y(\rho_y>\delta_y(\beta))=0$. In both cases,
\[
\sP_y\!\left(\sP_y(\gV_{y,\star}(\mD;\mI_{\mathrm{in}})\mid\mD_{\mathrm{in}})
>\delta_y(\beta)\right)
\eqby{eq:conditional-future-error}
\sP_y(\rho_y>\delta_y(\beta))\le\beta.
\]

When $m_y=1$ and $\theta_y=\kappa_m=0$, $W_y$ is Bernoulli with mean
\begin{equation}
\label{eq:single-extension-mean-error}
\E^{\sQ_{n_y,m_y}}W_y
\eqby{eq:one-future-rank}
\frac{n_y-k_y+1}{n_y+1}
=\frac{\lfloor n_y\epsilon_y\rfloor+1}{n_y+1}.
\end{equation}
Thus \eqref{eq:total-statistical-errors} equals
$\min\{1,(\lfloor n_y\epsilon_y\rfloor+1)/(\beta(n_y+1))\}$, including the single-extension case without a separate definition.
\end{proof}

The same rank law also gives the closed-form bound
\begin{equation}
\label{eq:general-coverage-threshold}
\delta_y(\beta)\le
\epsilon_y+\theta_y\frac{(n_y+m_y)^2}{n_ym_y}
+\frac{\kappa_m}{\beta\sP(Y_\star=y)}
+\sqrt{\frac{n_y+m_y+2}{2n_ym_y}\log\frac e\beta}.
\end{equation}
To see this, put $b:=\epsilon_y+\theta_y(n_y+m_y)^2/(n_ym_y)$ and $c:=2n_ym_y/(n_y+m_y+2)$.
If $b_y\ge k_y$, then $b\ge1$ and the bound is immediate.
Otherwise, $R:=[W_y-b]_+$ satisfies $\sQ_{n_y,m_y}(R>s)\le e^{-cs^2}$ by Lemma~\ref{lem:uniform-subset-rank}.
Choose weights $h=h(W_y)\in[0,1]$ selecting the upper $\beta$ fraction of the rank law, with $\E^{\sQ_{n_y,m_y}}h=\beta$ and fractional weight at the boundary atom.
The variational formula gives
\begin{equation}
\label{eq:weighted-rank-bound}
\delta_y(\beta)\le\frac{\E^{\sQ_{n_y,m_y}}[hW_y]}\beta
+\frac{\kappa_m}{\beta\sP(Y_\star=y)}
\le b+\frac{\E^{\sQ_{n_y,m_y}}[hR]}\beta
+\frac{\kappa_m}{\beta\sP(Y_\star=y)}.
\end{equation}
Since $\E^{\sQ_{n_y,m_y}}h=\beta$,
\begin{equation}
\frac{\E^{\sQ_{n_y,m_y}}[hR]}\beta
\leby{eq:weighted-tail-bound}\sqrt{\frac{\log(e/\beta)}c}.
\label{eq:stable-extension-moment}
\end{equation}
\begin{align*}
\delta_y(\beta)
&\leby{eq:weighted-rank-bound}
b+\frac{\E^{\sQ_{n_y,m_y}}[hR]}\beta+\frac{\kappa_m}{\beta\sP(Y_\star=y)}\\
&\leby{eq:stable-extension-moment}
\epsilon_y+\theta_y\frac{(n_y+m_y)^2}{n_ym_y}
+\sqrt{\frac{n_y+m_y+2}{2n_ym_y}\log\frac e\beta}
+\frac{\kappa_m}{\beta\sP(Y_\star=y)}.
\end{align*}
For fixed ranks and $m_y=n_y$, \eqref{eq:balanced-selected-coverage} also permits the deterministic bound obtained from \eqref{eq:balanced-selected-error} by replacing $\widehat\epsilon_y$ with $\epsilon_y$.
One may choose the smaller of that bound and \eqref{eq:total-statistical-errors} before observing the data, with the same confidence $1-\beta$.
}

\subsection{Intervention Guarantees}
\label{app:proof-learned-conditional-means}

{\begin{proof}
For training occurrence $i$, the monitor $\widehat X(\mD_{\mathrm{in}}^{-i})$ is measurable with respect to $\mD_{\mathrm{in}}^{-i}$.
Use $\gF=\gF_i(\mD_{\mathrm{in}}^{-i})$ in Proposition~\ref{prop:pathwise-calibration}(2), with sign $(-1)^{y+1}$. Then
\begin{align}
&\sP_y\!\left(
S_{y,i,i}(\mD;\mI_{\mathrm{in}})>\widehat s_{y,i,i}(\mD;\mI_{\mathrm{in}},r(\beta/(2n_y)))
\,\middle|\,\gF_0\right)
\nonumber\\
&\leby{eq:leave-one-out-conditional-means}
\sP_y\!\left(
\exists t:\,
(-1)^{y+1}\!\left[\widehat\mu_{t,i}(\mD_{\mathrm{in}}^{-i})
-\mu_{t,y}^{\widehat X(\mD_{\mathrm{in}}^{-i})}\right]
>r_t(\beta/(2n_y))
\,\middle|\,\gF_0\right)
\nonumber\\
&\le\frac{\beta}{2n_y}.
\label{eq:loo-conditional-estimation}
\end{align}
Define the training-measurable event
\begin{equation}
\label{eq:loo-training-estimation-event}
F_y:=\bigcap_{i\in\gI_{\mathrm{in},y}}
\left\{S_{y,i,i}(\mD;\mI_{\mathrm{in}})\le\widehat s_{y,i,i}(\mD;\mI_{\mathrm{in}},r(\beta/(2n_y)))\right\}.
\end{equation}
Taking expectations in \eqref{eq:loo-conditional-estimation} and applying the union bound,
\begin{equation}
\sP_y(F_y^c)
\le\sum_{i\in\gI_{\mathrm{in},y}}
\sP_y\!\left(S_{y,i,i}(\mD;\mI_{\mathrm{in}})
>\widehat s_{y,i,i}(\mD;\mI_{\mathrm{in}},r(\beta/(2n_y)))\right)
\leby{eq:loo-conditional-estimation}\sum_{i\in\gI_{\mathrm{in},y}}\frac{\beta}{2n_y}
=\frac\beta2.
\label{eq:loo-training-estimation-failure}
\end{equation}
On $F_y$, \eqref{eq:loo-training-estimation-event} gives
\begin{align}
&\min\left\{1-y,\left(S_{y,i,i}(\mD;\mI_{\mathrm{in}})\right)_{(k_y:\mD_{\mathrm{in},y})}+2{B_m}\right\}
\nonumber\\
&\leby{eq:order-statistic-monotonicity}
\min\left\{1-y,
\left(\widehat s_{y,i,i}(\mD;\mI_{\mathrm{in}},r(\beta/(2n_y)))\right)_{(k_y:\mD_{\mathrm{in},y})}
+2B_m\right\}
\nonumber\\
&\eqby{eq:learned-predictable-levels}\widehat q_y.
\label{eq:loo-training-order-bound}
\end{align}
Since $S_{y,\star,\star}(\mD;\mI_{\mathrm{in}})\le1-y$, \eqref{eq:loo-training-order-bound} gives
\begin{equation}
\label{eq:loo-future-error-inclusion}
\{S_{y,\star,\star}(\mD;\mI_{\mathrm{in}})>\widehat q_y\}\cap F_y
\subseteq
\gV_{y,\star}(\mD;\mI_{\mathrm{in}})\cap F_y.
\end{equation}
Put
\begin{equation}
\label{eq:loo-calibration-good-event}
G_y:=F_y\cap\left\{
\sP_y(\gV_{y,\star}(\mD;\mI_{\mathrm{in}})\mid\mD_{\mathrm{in}})
\le\delta_y(\beta/2)\right\}.
\end{equation}
By item~\textnormal{(3)},
\begin{align}
\sP_y(G_y^c)
&\le \sP_y(F_y^c)
+\sP_y\!\left(\sP_y(\gV_{y,\star}(\mD;\mI_{\mathrm{in}})\mid\mD_{\mathrm{in}})
>\delta_y(\beta/2)\right)
\nonumber\\
&\leby{eq:loo-training-estimation-failure}
\frac\beta2+\sP_y\!\left(\sP_y(\gV_{y,\star}(\mD;\mI_{\mathrm{in}})\mid\mD_{\mathrm{in}})
>\delta_y(\beta/2)\right)
\nonumber\\
&\leby{eq:training-conditional-generalization}
\frac\beta2+\frac\beta2=\beta.
\label{eq:loo-calibration-failure}
\end{align}
On $G_y$, training measurability gives
\begin{equation}
\label{eq:learned-predictable-coverage}
\sP_y\!\left(S_{y,\star,\star}(\mD;\mI_{\mathrm{in}})>\widehat q_y
\,\middle|\,\mD_{\mathrm{in}}\right)
\leby{eq:loo-future-error-inclusion}
\sP_y(\gV_{y,\star}(\mD;\mI_{\mathrm{in}})\mid\mD_{\mathrm{in}})
\leby{eq:loo-calibration-good-event}\delta_y(\beta/2).
\end{equation}

For the next trajectory, $\mD_{\mathrm{in}}^{-\star}=\mD_{\mathrm{in}}$, so use $\gF=\gF_\star(\mD_{\mathrm{in}})$ and the deployed monitor $\widehat X=\widehat X(\mD_{\mathrm{in}})$. Then
\begin{equation}
\label{eq:learned-conditional-means}
S_{y,\star,\star}(\mD;\mI_{\mathrm{in}})
=\max_{t\in\gW}(-1)^y\mu_{t,y}^{\widehat X}(Z_{t,\star};\gF).
\end{equation}
The bounds satisfy $\widehat q_0,-\widehat q_1\in[0,1]$.
All fitted parameters are $\gH:=\sigma(\mD_{\mathrm{in}})$-measurable, with $\gH\subseteq\gF_0$.
Apply Lemma~\ref{lem:conditional-intervention-bounds} on the training-measurable events $G_y$.

\emph{(a) False intervention.}
On $G_0$, take $p_0=\widehat q_0$, $q_0=\widehat p$, and $\lambda=\widehat\lambda$:
\begin{align}
\label{eq:learned-ville-validity}
\sP_0(\widehat\tau<\infty\mid\mD_{\mathrm{in}})
&\leby{eq:conditional-false-intervention}
\sP_0(\max_{s\in\gW}\mu_{s,0}^{\widehat X}>\widehat q_0\mid\mD_{\mathrm{in}})
+e_0(\widehat q_0;\widehat\lambda,\widehat p,\alpha)
\nonumber\\
&\eqby{eq:learned-conditional-means}
\sP_0(S_{0,\star,\star}(\mD;\mI_{\mathrm{in}})>\widehat q_0\mid\mD_{\mathrm{in}})
+e_0(\widehat q_0;\widehat\lambda,\widehat p,\alpha)
\nonumber\\
&\leby{eq:learned-predictable-coverage}
\delta_0(\beta/2)+e_0(\widehat q_0;\widehat\lambda,\widehat p,\alpha).
\end{align}
Consequently,
\[
\sP_0\!\left(\sP_0(\widehat\tau<\infty\mid\mD_{\mathrm{in}})
>\delta_0(\beta/2)+e_0(\widehat q_0;\widehat\lambda,\widehat p,\alpha)\right)
\leby{eq:learned-ville-validity}\sP_0(G_0^c)
\leby{eq:loo-calibration-failure}\beta.
\]

\emph{(b) Detection.}
On $G_1$, take $p_1=-\widehat q_1$, $q_0=\widehat p$, and $\lambda=\widehat\lambda$.
For every $t\in\gW$,
\begin{align}
\label{eq:learned-mean-detection}
\sP_1(\widehat\tau>t\mid\mD_{\mathrm{in}})
&\leby{eq:conditional-missed-detection}
\sP_1(\mu_{t,1}^{\widehat X}<-\widehat q_1\mid\mD_{\mathrm{in}})
+e_{1,t}(-\widehat q_1;\widehat\lambda,\widehat p,\alpha)
\nonumber\\
&\leby{eq:learned-conditional-means}
\sP_1(S_{1,\star,\star}(\mD;\mI_{\mathrm{in}})>\widehat q_1\mid\mD_{\mathrm{in}})
+e_{1,t}(-\widehat q_1;\widehat\lambda,\widehat p,\alpha)
\nonumber\\
&\leby{eq:learned-predictable-coverage}
\delta_1(\beta/2)+e_{1,t}(-\widehat q_1;\widehat\lambda,\widehat p,\alpha).
\end{align}
\begingroup
\makeatletter
\def\@currentlabel{\theequation}
\label{eq:learned-mean-detection-condition}
\makeatother
\endgroup
The same event $G_1$ works for every $t$, so
\[
\sP_1\!\left(\exists t\in\gW:\sP_1(\widehat\tau>t\mid\mD_{\mathrm{in}})
>\delta_1(\beta/2)+e_{1,t}(-\widehat q_1;\widehat\lambda,\widehat p,\alpha)\right)
\leby{eq:learned-mean-detection}\sP_1(G_1^c)
\leby{eq:loo-calibration-failure}\beta.
\]
On $\gB_0^c$, the fallback $\widehat p=\widehat\lambda=1$ gives $\widehat\tau=\infty$ and $e_{1,t}(-\widehat q_1;\widehat\lambda,\widehat p,\alpha)=1$.

The outer probabilities range over all training samples. If
$\sP_y(\gB_0)>0$, conditioning further gives only
\begin{equation}
\label{eq:separation-conditional-confidence}
\sP_y(G_y^c\mid\gB_0)\le\frac{\beta}{\sP_y(\gB_0)}.
\end{equation}
A known lower bound $0<\pi_y\le\sP_y(\gB_0)$ permits conditional confidence
$1-\beta$ by using $\beta\pi_y$ in the corresponding error bound.
\end{proof}

}
     % Appendix J
\section{Proof of Lemma~\ref{lem:rule-edit-structure}}
\label{app:proof-rule-edit-structure}

\begin{proof}
Fix $\gS\in\gC$. The Boolean identities below hold for every
$i\in[n]$ and $t\in[T_1]$.

For part~\textnormal{(1)}, the definition gives
\[
X_{t,i}(\gS)=1
\ \Longleftrightarrow\
\bigvee_{\gR\in\gS}\xi_{t,i}^{\gR}=1
\ \Longleftrightarrow\
\{\gR\in\gS:\xi_{t,i}^{\gR}=1\}\ne\emptyset.
\]
The empty disjunction is zero, so this includes $\gS=\emptyset$.

For part~\textnormal{(2)}, let $\gR\in\gS$, $\gR'\subseteq[d]$ with
$1\le|\gR'|\le K_{\gR}$, and
$\gS'=(\gS\setminus\{\gR\})\cup\{\gR'\}$.
For $a,b\in\{0,1\}$, applying $a\vee b=a+(1-a)b$ to the two disjunctions gives
\begin{align*}
X_{t,i}(\gS)
&=X_{t,i}(\gS\setminus\{\gR\})
+\bigl(1-X_{t,i}(\gS\setminus\{\gR\})\bigr)\xi_{t,i}^{\gR},\\
X_{t,i}(\gS')
&=X_{t,i}(\gS\setminus\{\gR\})
+\bigl(1-X_{t,i}(\gS\setminus\{\gR\})\bigr)\xi_{t,i}^{\gR'}.
\end{align*}
Subtracting these identities gives
\begin{equation}
\label{eq:rule-replacement-difference}
X_{t,i}(\gS')-X_{t,i}(\gS)
=\bigl(1-X_{t,i}(\gS\setminus\{\gR\})\bigr)
\bigl(\xi_{t,i}^{\gR'}-\xi_{t,i}^{\gR}\bigr).
\end{equation}
This includes $\gR'\in\gS\setminus\{\gR\}$, since $a\vee a=a$.

For part~\textnormal{(3)}, conjunction is multiplication and
$a^2=a$ for $a\in\{0,1\}$, so
\begin{equation}
\label{eq:rule-predicate-addition}
\xi_{t,i}^{\gR\cup\{\ell\}}
=\prod_{j\in\gR\cup\{\ell\}}\xi_{t,i}^{j}
=\left(\prod_{j\in\gR}\xi_{t,i}^{j}\right)\xi_{t,i}^{\ell}
=\xi_{t,i}^{\gR}\xi_{t,i}^{\ell}.
\end{equation}
Also,
\begin{align*}
X_{t,i}(\gS\cup\{\{\ell\}\})-X_{t,i}(\gS)
&=\bigl(X_{t,i}(\gS)\vee\xi_{t,i}^{\ell}\bigr)-X_{t,i}(\gS)
=
\bigl(1-X_{t,i}(\gS)\bigr)\xi_{t,i}^{\ell}.
\end{align*}

For part~\textnormal{(4)}, suppose
$X_{s,i}(\gS')\le X_{s,i}(\gS)$ for every $s\in[T_1]$.
For every $t\in\gW$,
\begin{equation}
\label{eq:rule-edit-average-order}
\widehat\mu_{t,i}^{X(\gS')}
\eqby{eq:trajectory-confidence-bounds}
\frac1t\sum_{s=1}^tX_{s,i}(\gS')
\le\frac1t\sum_{s=1}^tX_{s,i}(\gS)
=\widehat\mu_{t,i}^{X(\gS)}.
\end{equation}
Taking maxima and minima preserves this order. Thus
\begin{equation}
\label{eq:rule-edit-extrema-order}
\widehat U_i^{X(\gS')}\le\widehat U_i^{X(\gS)},
\qquad
\widehat L_i^{X(\gS')}\le\widehat L_i^{X(\gS)}.
\end{equation}
If the pointwise ordering holds for every $i\in[n]$, then
\[
\widehat U_{(k:\gI_{\mathrm{in},0})}^{X(\gS')}
\leby{eq:order-statistic-monotonicity}\widehat U_{(k:\gI_{\mathrm{in},0})}^{X(\gS)},\quad k\in[n_0],
\qquad
\widehat L_{(k:\gI_{\mathrm{in},1})}^{X(\gS')}
\leby{eq:order-statistic-monotonicity}\widehat L_{(k:\gI_{\mathrm{in},1})}^{X(\gS)},\quad k\in[n_1].
\]

For predicate addition,
\begin{equation*}
\xi_{t,i}^{\gR\cup\{\ell\}}
\eqby{eq:rule-predicate-addition}
\xi_{t,i}^{\gR}\xi_{t,i}^{\ell}
\le\xi_{t,i}^{\gR},
\end{equation*}
so \eqref{eq:rule-replacement-difference} gives the required pointwise
ordering. Deleting a rule also cannot increase the disjunction.
Adding a rule cannot decrease the disjunction, and removing a predicate
cannot decrease its conjunction, provided the conjunction remains nonempty.
Interchanging $\gS$ and $\gS'$ in the preceding monotonicity argument
gives the reversed inequalities.

For part~\textnormal{(5)}, let $\gR',\gR\in\gS$ with
$\gR'\subsetneq\gR$. Then
\begin{equation*}
\xi_{t,i}^{\gR}
=\xi_{t,i}^{\gR'}\prod_{j\in\gR\setminus\gR'}\xi_{t,i}^{j}
\le\xi_{t,i}^{\gR'}
\le X_{t,i}(\gS\setminus\{\gR\}).
\end{equation*}
Hence
\begin{equation*}
X_{t,i}(\gS)
=X_{t,i}(\gS\setminus\{\gR\})\vee\xi_{t,i}^{\gR}
=X_{t,i}(\gS\setminus\{\gR\}).
\end{equation*}
\end{proof}
                % Appendix K
% \input{appendix/proof-finite-class-generalization}          % Temporarily excluded
% Compile the paper and Lean supplement together for arXiv.
\clearpage

\raggedbottom
\setcounter{section}{0}
\setcounter{equation}{0}
\setcounter{page}{1}
\renewcommand{\thesection}{\arabic{section}}
\renewcommand{\presectionnum}{}
\renewcommand{\theHsection}{supp.\arabic{section}}
\renewcommand{\theHequation}{supp.\arabic{equation}}
\gdef\thepage{S-\arabic{page}}
\makeatletter
\gdef\@shorttitle{Lean 4 Supplement}
\renewcommand{\@pnumwidth}{3em}
\renewcommand{\@tocrmarg}{4em}
\makeatother
\etocdepthtag.toc{supplement}
\thispagestyle{plain}
\phantomsection
\pdfbookmark[0]{Lean 4 Supplement}{lean-supplement}
\label{supp:start}
\jmlrpretitle Learn then Bet: Lean 4 Supplement\jmlrposttitle
\setcounter{tocdepth}{2}
\etocsettagdepth{paper}{none}
\etocsettagdepth{supplement}{subsection}
\tableofcontents

\section{Lean 4 setup}
\label{supp:setup}

Section~\ref{supp:challenge}, \emph{Challenge}, pairs the paper's assumptions
and guarantees with their Lean statements.
Section~\ref{supp:solution}, \emph{Solution}, presents the corresponding
proofs and their supporting results.

A \emph{Challenge} states the required proposition with \leanname{sorry}
in place of its proof. The corresponding Solution supplies a proof checked
by Lean. Comparator checks that the Solution has the same statement and
definitions. The mathematical review therefore concerns the displayed
assumptions, definitions, and conclusions.

The listings use the common Lean context:
\begin{lstlisting}[style=leansupplement]
open MeasureTheory ProbabilityTheory Filter Finset
open scoped Classical
open LeBet.StratifiedSample LeBet.SampleVector
\end{lstlisting}

For $S\subseteq\Omega$, \leanname{Set.Elem S} is the type of elements of $S$.
It records an element $\omega$ together with $\omega\in S$.

The proofs may use the following three axioms.
Propositional extensionality identifies equivalent propositions:
\[
(P\leftrightarrow Q)\ \Longrightarrow\ P=Q.
\]
\begin{lstlisting}[style=leansupplement]
#check propext
\end{lstlisting}
Classical choice selects an element of a nonempty type:
\[
\operatorname{choice}_\alpha:\operatorname{Nonempty}(\alpha)\to\alpha.
\]
\begin{lstlisting}[style=leansupplement]
#check Classical.choice
\end{lstlisting}
Quotient soundness identifies related representatives in the quotient:
\[
r(a,b)\ \Longrightarrow\ [a]_r=[b]_r.
\]
\begin{lstlisting}[style=leansupplement]
#check Quot.sound
\end{lstlisting}
These logical axioms are separate from the statistical assumptions stated in Challenge.

\section{Challenge}
\label{supp:challenge}

The specification follows the main text. Each mathematical definition or
statement is followed by its Lean formulation. Each Lean block states its
namespace and variable context; proof lemmas are given in
Section~\ref{supp:solution}.

\subsection{\paperlink{app:notation}{Notation}}

The Lean names \leanname{sP}, \leanname{mD}, \leanname{gW}, and \leanname{gS}
represent $\sP$, $\mD$, $\gW$, and $\gS$.
Subscripts use underscores, and hats use quoted identifiers such as
\lstinline[style=leansupplement]!«X̂»!.
\lstinline[style=leansupplement]!Vector Ω N! represents $\Omega^N$;
\leanname{Fin N} indexes its entries by $0,\ldots,N-1$.
\leanname{sP.real A} is $\sP(A)$, $=^m[\sP]$ is almost-sure equality,
and $\top$ denotes $\infty$.

\subsection{\paperlink{sec:problem-statement}{Sample space and observed data}}
\label{supp:data}

Let $(\Omega,\gF_\infty)$ be a measurable space and
$Y:\Omega\to\{0,1\}$ measurable. Fix $n_y,m_y\ge1$ and set
\[
\Omega_y:=\{\omega:Y(\omega)=y\},\qquad
\mOmega:=\Omega_0^{n_0+m_0}\times\Omega_1^{n_1+m_1}\times\Omega.
\]
\reviewcode{sample-space}

Put $n=n_0+n_1$ and $m=m_0+m_1$.
For $\mD=(\omega_1,\ldots,\omega_{n+m+1})$, let $\mD_{\mI}$ denote its indexed subvector.
Write $\mI_{\mathrm{in}}=(1,\ldots,n)$ and
\[
\gI_{\mathrm{in},y}:=\{i\in[n]:Y_i=y\},\quad
\gI_{\mathrm{out},y}:=\{i\in[n+m]\setminus[n]:Y_i=y\},\quad
\gI_{\mathrm{in/out},y}:=\gI_{\mathrm{in},y}\cup\gI_{\mathrm{out},y}.
\]
Place the observed trajectories first and $\omega_\star$ last:
\[
\begin{aligned}
\mD_{\mathrm{in}}&:=(\omega_1,\ldots,\omega_n),\\
\mD_{\mathrm{out}}&:=(\omega_{n+1},\ldots,\omega_{n+m}),\\
\mD&:=(\mD_{\mathrm{in}},\mD_{\mathrm{out}},\omega_\star).
\end{aligned}
\]
\reviewcode{data}

Equip $\mOmega$ with its product $\sigma$-algebra and probability law $\sP$.
For $\sP(Y_\star=y)>0$, write $\sP_y:=\sP(\,\cdot\mid Y_\star=y)$.

\subsection{\paperheading{assump:finite-exchangeable-extension}{Assumption}{Finite exchangeable extension}}
\label{supp:exchangeability}

\paragraph{Assumption.}
For each $y$ with $\sP(Y_\star=y)>0$,
\[
\pi(\mD)\overset{d}{=}\mD\quad\text{under }\sP_y,
\qquad
\pi\in\Pi_{\gI_{\mathrm{in/out},y}\cup\{\star\}}\cup\Pi_{\gI_{\mathrm{in/out},1-y}},
\]
where $\Pi_I$ fixes every index outside $I$.
Here \leanname{cond sP (classEvent y)} is $\sP_y$, and \leanname{map}
is the pushforward by the indicated permutation.
\reviewcode{exchangeability-assumption}

\subsection{\paperlink{sec:learned-intervention-guarantees}{Monitor and empirical levels}}
\label{supp:scores}

Let ${\gF=}(\gF_t)_{t\in\sN}$ be a filtration on $\Omega$.
Fix $1<T_0<T_1$ and $\gW=\{T_0,\ldots,T_1\}$.
A binary monitor is a process $(X_t)_{t\ge1}$ with
$X_t:\Omega\to\{0,1\}$ measurable with respect to $\gF_t$.
Lean factor index $s$ represents paper time $s+1$.
In the monitor blocks, the local variable \lstinline[style=leansupplement]!mΩ!
has type \lstinline[style=leansupplement]!MeasurableSpace Ω!.
It is the measurable structure on a single trajectory.
\reviewcode{monitor}

Fix the predicate pool $\Xi=(\xi^j)_{j\in[d]}$ before observing
$\mD_{\mathrm{in}}$. Each $\xi^j$ is an adapted binary process.
Lean indexes the predicates by $0,\ldots,d-1$;
\lstinline[style=leansupplement]!(Ξ j).X s ω! is $\xi^{j+1}_{s+1}(\omega)$.
\begin{lstlisting}[style=leansupplement]
section PredicateNotation
variable {Ω : Type*} {mΩ : MeasurableSpace Ω}
  {gF : Filtration ℕ mΩ} {d : ℕ}
  (Ξ : Fin d → LeBet.BinaryMonitor gF)

local notation "ξ" => fun (j : Fin d) => LeBet.BinaryMonitor.X (Ξ j)

end PredicateNotation
\end{lstlisting}

For a nonempty rule $\gR\subseteq[d]$, define
\[
\xi_t^{\gR}(\omega):=\bigwedge_{j\in\gR}\xi_t^j(\omega)
=\1\{\forall j\in\gR,\ \xi_t^j(\omega)=1\}.
\]
\reviewcode{conjunction}

For $K_{\gR},K_{\gS}\ge1$, the feasible rule sets form
\[
\gC:=\{\gS\subseteq2^{[d]}:|\gS|\le K_{\gS},\
1\le|\gR|\le K_{\gR}\ \forall\gR\in\gS\}.
\]
\reviewcode{rule-class}

For $\gS\in\gC$, combine the rules by disjunction:
\[
X_t(\gS)(\omega):=\bigvee_{\gR\in\gS}\xi_t^{\gR}(\omega)
=\1\{\exists\gR\in\gS,\ \forall j\in\gR,\ \xi_t^j(\omega)=1\}.
\]
The empty rule set gives $X_t(\emptyset)=0$.
\reviewcode{rule-monitor}

Order statistics count repetitions. For an indexed sample $\mD$, write
$v_{(k:\mD)}$ for its $k$th smallest value.

\subsection{Leave-one-out fitting and conditional means}
\label{supp:loo}

Let $\widehat{\gS}(\mD)\in\gC$ be measurable and invariant under permutations
of the training sample, and write $\widehat X(\mD):=X(\widehat{\gS}(\mD))$.
The same procedure applies to the remaining class sizes after deletion,
returning $\emptyset$ if either class is absent.
Write $\mathfrak D_{\mathrm{in/out}}$ for the indexed samples drawn from
$\omega_1,\ldots,\omega_{n+m}$ with $n_y$ occurrences in each class $y$.
Let $\mD^{-j}$ delete occurrence $j$ if present, and equal $\mD$ otherwise.
Write
\[
\widehat\mu_{t,i}(\mD):=\widehat\mu_{t,i}^{\widehat X(\mD)},\qquad
\gF_i(\mD):=(\sigma(\mD,Z_{u,i}))_{u\ge0}.
\]
The average conditional mean used in \eqref{eq:leave-one-out-conditional-means} is
\[
\mu_{t,y}^{\widehat X(\mD_{\mI})}(Z_{t,i};\gF_i(\mD_{\mI}))
=\frac1t\sum_{s=1}^t
\E_y[\widehat X_s(\mD_{\mI})(Z_{s,i})\mid\mD_{\mI},Z_{s-1,i}].
\]
\reviewcode{loo-mean}

For $r=(r_t)_{t\in\gW}\ge0$, define
\[
\begin{aligned}
s_{y,i,j}(\mD;\mI,r)&:=\max_{t\in\gW}[(-1)^y
\mu_{t,y}^{\widehat X(\mD_{\mI}^{-j})}(Z_{t,i};\gF_i(\mD_{\mI}^{-j}))+r_t],\\
\widehat s_{y,i,j}(\mD;\mI,r)&:=\max_{t\in\gW}
[(-1)^y\widehat\mu_{t,i}(\mD_{\mI}^{-j})+r_t].
\end{aligned}
\]
\reviewcode{loo-values}

\subsection{Leave-one-out stability and rank calibration}
\label{supp:stability}

For integers $k\ge1$ and $h\ge0$, \eqref{def:sampling-laws} defines
\[
\sQ_{k,h}:=\operatorname{Uniform}\bigl(\{K\subseteq[k+h]:|K|=k\}\bigr).
\]
In Lean, this is \leanname{subsetLaw} on \leanname{Fin (k + h)} with sample size $k$.
\reviewcode{subset-sampling}
Let $\sQ_\pi$ be uniform on the group generated by
$\Pi_{\gI_{\mathrm{in/out},0}}\cup\Pi_{\gI_{\mathrm{in/out},1}}$, and write
$\sP_{y,\pi}:=\sP_y\otimes\sQ_\pi$ for the independent joint law of $(\mD,\pi)$.
The Lean product uses the coordinate order $(\pi,\mD)$.
\reviewcode{exchange-law}

Write $S_{y,i,j}(\mD;\mI):=s_{y,i,j}(\mD;\mI,0)$.
Let $\epsilon_y,\theta_y,\kappa_m\in[0,1)$ and $\alpha,\beta\in(0,1)$.
For $\mD_{\mI},\mD_{\mI'}\in\mathfrak D_{\mathrm{in/out}}$, define
\[
\gE_{\mI,\mI'}^m:=\bigcap_{y\in\{0,1\}}
\left\{\frac1{n_y+m_y}\sum_{i\in\gI_{\mathrm{in/out},y}}
\1\{|S_{y,i,i}(\mD;\mI)-S_{y,i,i}(\mD;\mI')|>B_m\}\le\theta_y\right\}.
\]
Assumption~\getrefnumber{assump:score-m-stability} requires
$\E^{\sQ_\pi}\sP((\gE_{\pi(\mI_{\mathrm{in}}),\mI_{\mathrm{in}}}^m)^c)\le\kappa_m$.
Section~\ref{supp:solution-setup} gives its classwise Lean formulation and
\eqref{eq:class-conditional-stability-failure}.

Put $k_y=n_y-\lfloor n_y\epsilon_y\rfloor$ and $b_y=\lfloor(n_y+m_y)\theta_y\rfloor$.
For $\gK_y=\{i_1<\cdots<i_{n_y}\}\sim\sQ_{n_y,m_y}$, set
\[
J_{\gK_y}^a=i_{a+1}-a-1,\qquad
W_y=\begin{cases}
\min\{1,(J_{\gK_y}^{n_y-k_y+b_y}+b_y)/m_y\},&b_y<k_y,\\
1,&b_y\ge k_y.
\end{cases}
\]
\reviewcode{rank-variables}
For $\beta\in(0,1)$, define
\[
\delta_y(\beta):=\min_{0\le u\le1}
\left\{u+\frac{\E^{\sQ_{n_y,m_y}}[(W_y-u)_+]
+(1-u)\kappa_m/\sP(Y_\star=y)}{\beta}\right\}.
\]
\reviewcode{rank-objective}

Define the error events and counts by \eqref{eq:learned-detection-condition}:
\[
\begin{aligned}
\gV_{y,i}(\mD;\mI)&:=\{S_{y,i,i}(\mD;\mI)>
(S_{y,i,i}(\mD;\mI))_{(k_y:(\mD_{\mI})_y)}+2B_m\},\\
C_y(\mD;\mI)&:=\sum_{i\in\gI_{\mathrm{in/out},y}\setminus\gI}\1\{\gV_{y,i}(\mD;\mI)\}.
\end{aligned}
\]

\subsection{Observable levels and betting parameters}
\label{supp:parameters}

Write $a_t(\epsilon):=\sqrt{2\log(1/\epsilon)/t}$ and
$r(\epsilon):=(r_t(\epsilon))_{t\in\gW}$, where
\[
r_t(\epsilon)=\min\left\{
\frac{1+T_0/t}{4}a_{T_0}(\epsilon/2),
\frac12a_t(\epsilon/(2|\gW|))\right\}.
\]
\reviewcode{mean-radius}
The observable bound in \eqref{eq:learned-predictable-levels} is
\[
\widehat q_y:=\min\left\{1-y,
\left(\widehat s_{y,i,i}\left(\mD;\mI_{\mathrm{in}},
r\left(\frac{\beta}{2n_y}\right)\right)\right)_{(k_y:\mD_{\mathrm{in},y})}+2B_m\right\}.
\]
\reviewcode{loo-level}

The deployment parameters use the full-fit empirical extrema
\[
\widehat U_i^{\widehat X(\mD)}=\max_{t\in\gW}\widehat\mu_{t,i}(\mD),\qquad
\widehat L_i^{\widehat X(\mD)}=\min_{t\in\gW}\widehat\mu_{t,i}(\mD).
\]
\reviewcode{raw-scores}
Write $\widehat\Delta(\mD):=
[\widehat L_{(n_1-k_1+1:\mD_1)}^{\widehat X(\mD)}
-\widehat U_{(k_0:\mD_0)}^{\widehat X(\mD)}]_+$.
If $\widehat\Delta(\mD)>0$, set
\[
\widehat p_0:=\widehat U_{(k_0:\mD_0)}^{\widehat X(\mD)},\qquad
\widehat p_1:=\widehat L_{(n_1-k_1+1:\mD_1)}^{\widehat X(\mD)},\qquad
\widehat p:=(\widehat p_0+\widehat p_1)/2.
\]
\reviewcode{midpoint-parameters}
With $v(p):=1/4-[p-1/2]_+^2$, put $\widehat\lambda:=a_{T_1}(\alpha)/\sqrt{v(\widehat p)}$.
Otherwise, set $(\widehat p,\widehat\lambda):=(1,1)$.
\reviewcode{bernoulli-curvature}
\reviewcode{calibrated-stake}

Write $\psi_\lambda(p):=\log(1-p+pe^\lambda)$ and use the Bernoulli KL divergence.
\reviewcode{bernoulli-functions}
For the deployed monitor $\widehat X(\mD_{\mathrm{in}})$, set
\[
E_t(\lambda,p):=\exp\!\left\{\lambda\sum_{s=1}^tX_s-t\psi_\lambda(p)\right\},\qquad
\tau(\lambda,p):=\inf\{t\in\gW:E_t(\lambda,p)\ge1/\alpha\}.
\]
\reviewcode{monitor-evidence}
The error terms are those of \eqref{eq:mean-false-intervention}:
\[
\begin{aligned}
e_0(p;\lambda,q,\alpha)&:=\alpha\exp\!\left(T_1[\psi_\lambda(p)-\psi_\lambda(q)]_+
-T_0[\psi_\lambda(q)-\psi_\lambda(p)]_+\right),\\
e_{1,t}(p;\lambda,q,\alpha)&:=\exp\!\left(-t\,\KL\!\left(
\Ber\!\left(\min\{p,\lambda^{-1}\{\psi_\lambda(q)+a_t(\alpha)^2/2\}\}\right)
\middle\|\Ber(p)\right)\right).
\end{aligned}
\]
\reviewcode{mean-errors}

\subsection{Learned intervention guarantees}
\label{supp:challenge-learned}

Assume finite exchangeable extension, leave-one-out stability,
$\sP(Y_\star=y)>0$, and a measurable, permutation-invariant selector.
Let $\widehat\tau:=\tau(\widehat\lambda,\widehat p)$.
The following Lean context specifies the variables and hypotheses.
Section~\ref{supp:solution-setup} gives the coordinate construction and notation correspondence.
\reviewcode{paper-context}

Part~(1) of Theorem~\getrefnumber{thm:learned-intervention-guarantees} states
\[
\sP_y(\gV_{y,\star}(\mD;\mI_{\mathrm{in}})\mid\mD_{\mathrm{in}})
=\E_y[C_y(\mD;\mI_{\mathrm{in}})/m_y\mid\mD_{\mathrm{in}}]
\quad\sP_y\text{-almost surely}.
\]
\reviewcode{challenge_conditional_error_identity}

Part~(2), for $s>0$, states
\[
\sP_y\!\left(\frac{C_y(\mD;\mI_{\mathrm{in}})}{m_y}>
\epsilon_y+\theta_y\frac{(n_y+m_y)^2}{n_ym_y}+s\right)
\le e^{-2n_ym_ys^2/(n_y+m_y+2)}+\frac{\kappa_m}{\sP(Y_\star=y)}.
\]
\reviewcode{challenge_extension_tail}

Part~(3) states
$\sP_y(\sP_y(\gV_{y,\star}(\mD;\mI_{\mathrm{in}})\mid\mD_{\mathrm{in}})>\delta_y(\beta))\le\beta$.
\reviewcode{challenge_error_coverage}

Part~(4) states
\[
\sP_0\big(\sP_0(\widehat\tau<\infty\mid\mD_{\mathrm{in}})>
\delta_0(\beta/2)+e_0(\widehat q_0;\widehat\lambda,\widehat p,\alpha)\big)\le\beta.
\]
\reviewcode{challenge_false_intervention}

Part~(5) states
\[
\sP_1\big(\exists t\in\gW:\sP_1(\widehat\tau>t\mid\mD_{\mathrm{in}})>
\delta_1(\beta/2)+e_{1,t}(-\widehat q_1;\widehat\lambda,\widehat p,\alpha)\big)\le\beta.
\]
\reviewcode{challenge_detection}

\section{Solution}

\label{supp:solution}

The setup collects the coordinate definitions and interface conversions.
The remaining proofs follow the main appendix.

\subsection{Sample coordinates and leave-one-out setup}

\label{supp:solution-setup}

Coordinates carry the product measurable structure.

\reviewcode{sample-coordinates}

The first class entries are observed; the remaining entries form the extension.

\reviewcode{sample-indices}

The embeddings preserve training indices and offset extension indices.

\begin{lstlisting}[style=leansupplement]
section SampleIndexNotation
variable (n_y m_y : ℕ)

local notation "I_in_y" => gI_in_y n_y m_y
local notation "I_out_y" => I_in_yᶜ
local notation "I_y" => I_in_y ∪ I_out_y

end SampleIndexNotation
\end{lstlisting}

Extract the observed and extension vectors in their original order.

\reviewcode{sample-slices}

\begin{reviewunit}

The next-label event is encoded by \leanname{JointSample.classEvent}.

\reviewcode{class-event}

\end{reviewunit}

The class-conditional law and training sigma-algebra are represented below.

\begin{lstlisting}[style=leansupplement]
section SampleLawNotation
variable {Ω : Type*} [MeasurableSpace Ω] {Y : Ω → Fin 2}
  {n_0 m_0 n_1 m_1 : ℕ}
  (sP : Measure (mΩ Y n_0 m_0 n_1 m_1)) [IsProbabilityMeasure sP]
  (y : Fin 2)

local notation "π_y" => sP.real (JointSample.classEvent y)
local notation "sP_y" => cond sP (JointSample.classEvent y)
local notation "gF_in" =>
  MeasurableSpace.comap
    (mD_in (Y := Y) (n_0 := n_0) (m_0 := m_0)
      (n_1 := n_1) (m_1 := m_1)) inferInstance

variable (hπ : 0 < π_y)
end SampleLawNotation
\end{lstlisting}

Permute one class and keep the other class and next trajectory fixed.

\reviewcode{class-permutations}

The next-trajectory permutation acts on the corresponding label event.

\reviewcode{next-permutations}

Sum the monitor observations up to the specified time.

\reviewcode{monitor-sum}

The Lean field \lstinline[style=leansupplement]!BinaryMonitor.μ! is a bounded
one-step conditional mean. The manuscript uses its time average.

\reviewcode{mean-steps}

The compensator sums these one-step conditional means.

\reviewcode{mean-compensator}

Quantiles count repetitions and use the integer violation budget.

\reviewcode{quantiles}

The floor of the empirical error budget gives the permitted number of violations.

\reviewcode{rank-budget}

\paragraph{Classwise representation.}

Lean treats one class at a time. The types \leanname{O} and \leanname{R}
store the other class and its training data. The map \leanname{other} extracts
that training data. The learner is measurable and permutation-invariant at
each reduced sample size.

\reviewcode{loo-sample}

Permute one class and keep the other class fixed.

\reviewcode{loo-permute}

Append the next trajectory as the last coordinate of the evaluated class.

\reviewcode{joint-append}

Separate the two class blocks and the next trajectory.

\reviewcode{joint-class-raw-coordinates}

The composed maps give the classwise sample used by the LOO proofs.

\reviewcode{joint-class-coordinates}

The set \leanname{K} selects training occurrences in the evaluated class.
Together with \leanname{other}, it determines the fitting sample and training information.

\reviewcode{loo-training}

The filtration combines the fitting sample with the evaluated trajectory's observed history.

\reviewcode{loo-filtration}

Fit on the selected training sample and evaluate the specified trajectory.

\reviewcode{loo-monitor}

Finite averaging constructs bounded conditional-mean versions with exact permutation symmetry.

\reviewcode{loo-conditional-version}

Take the temporal maximum of signed means with the supplied radius.

\reviewcode{signed-maximum}

The training and extension pool excludes the last, next-trajectory coordinate.

\reviewcode{loo-pool}

Relabel the selected training indices.

\reviewcode{rank-relabel}

The array \leanname{u} contains the diagonal LOO conditional-mean extrema;
\leanname{v} is the next-trajectory value under the deployed fit.

\reviewcode{loo-rank-values}

The training quantile uses the integer violation budget \leanname{r}.

\reviewcode{occurrence-level}

Permute the other class while fixing the evaluated class.

\reviewcode{other-permutations}

Average stability failure over training subsets and permutations of the other class.
The class-conditioning lemma supplies the failure allowance.

\reviewcode{average-stability}

Average the strict extension violations above the adjusted training quantile.

\reviewcode{extension-rate}

Conditional probabilities are conditional expectations of indicators.

\reviewcode{conditional-probability}

The LOO radius allocates the mean-estimation error across training occurrences.
Its agreement with the manuscript radius is proved by
\leanname{leaveOneOutRadius_eq_meanRadius}.

\reviewcode{average-radius}

Training monitors exclude the evaluated occurrence. The deployed monitor uses the full training sample.

\reviewcode{loo-monitors}

Add the stability allowance after taking the training quantile.

\reviewcode{upper-level}

Apply the confidence radius before temporal maximization and clip after taking the quantile.

\reviewcode{mean-level}

The following proofs establish the sample operations, measurability, and coordinate conversions.

\proofcode{LeBet/SampleVector}

\proofcode{LeBet/StratifiedSample}

\proofcode{LeBet/JointSample}

\proofcode{LeBet/SampleActions}

\proofcode{LeBet/PaperAssumptions}

\proofcode{LeBet/RuleMonitor}

\proofcode{LeBet/JointExperiment}

\proofcode{LeBet/ExchangeableExtension}

\proofcode{LeBet/ConditionalPullback}

Condition the average stability failure on the next label.

\proofpart{class-conditioning}

\proofcode{LeBet/LearnedMonitor}

The fitted monitor is adapted to the sample-augmented filtration.

\proofcode{LeBet/LeaveOneOutModel}

Finite averaging gives conditional means with exact permutation symmetry.

\proofcode{LeBet/EquivariantConditionalMeans}

\proofcode{LeBet/LeaveOneOutSymmetry}

\proofcode{LeBet/LeaveOneOutValues}

Coordinate transport preserves the training information, conditional probabilities, and PAC events.

\proofcode{LeBet/JointLeaveOneOut}

\proofpart{occurrence-setup}

\proofpart{loo-calibration-setup}

\proofpart{average-stability-setup}

Uniformly averaging law-preserving exchanges preserves the sample law.

\proofpart{average-balanced-setup}

The default parameters are measurable, and the stake is positive.

\proofpart{default-parameters}

\subsection{\paperheading{app:betting-interpretation}{Appendix}{Betting Interpretation of the Monitor Evidence}}

The log-growth decomposition gives the Bernoulli KL maximum and its optimizer.

\proofcode{LeBet/KL}

\subsection{\paperheading{app:auxiliary-lemmas}{Appendix}{Auxiliary Lemmas}}

Log-normalizer bounds give the evidence comparisons.

\proofpart{evidence-bounds}

\subsubsection{\paperlink{app:auxiliary-hoeffding}{Hoeffding Bounds}}

The finite-sampling proof uses Bernoulli populations. The general convex
comparison and arbitrary bounded-population versions are not separately formalized.

\proofcode{LeBet/Hoeffding}

\proofcode{LeBet/FiniteSampling}

\proofcode{LeBet/SamplingTails}

\proofcode{LeBet/RankProbability}

Tail integration gives the weighted bound and the conditional transfer.

\proofcode{LeBet/TruncatedCoverage}

\subsubsection{\paperlink{app:auxiliary-order-statistics}{Order Statistics and Finite Averaging}}

Order-statistic identities preserve strict violations and repeated values.

\proofcode{LeBet/OrderStatistics}

\proofcode{LeBet/FiniteSymmetry}

\proofcode{LeBet/UniformRank}

\proofcode{LeBet/OneFuture}

The exact rank law also gives the single-extension case.

\proofcode{LeBet/RankDistribution}

Reflection at the first crossing gives the balanced finite-sampling bound.

\proofcode{LeBet/BalancedReflection}

\proofcode{LeBet/BalancedPrefix}

\subsubsection{\paperlink{app:auxiliary-conditional-means}{Conditional Means and Stability}}

\paragraph{\paperlink{lem:class-permutation-average}{Training-sample averaging}.}

Reindex the extension errors under training-sample permutations.

\proofpart{finite-average-reindexing}

\subsubsection{\paperlink{app:auxiliary-concentration}{Martingales and Concentration}}

The Bernoulli exponential martingale gives the concentration bounds.

\proofcode{LeBet/Probability}

\proofpart{bernoulli-martingale}

\proofcode{LeBet/Concentration}

Stop at the first crossing and apply conditional optional stopping.

\proofcode{LeBet/ConditionalVille}

Localize bounded tilts to obtain the conditional crossing bound for finite measurable tilts.

\proofcode{LeBet/ConditionalBernoulli}

The same process gives the linear and fixed-time Hoeffding bounds.

\proofcode{LeBet/ConditionalHoeffding}

\subsubsection{\paperlink{app:auxiliary-intervention}{Intervention Events}}

These identities characterize the evidence and stopping events.

\proofpart{fixed-monitor-intervention}

\proofpart{detection-events}

A signed deviation bound controls the corresponding temporal maximum.

\proofpart{mean-support}

\proofcode{LeBet/SignedMeanValues}

\paragraph{\paperlink{lem:conditional-intervention-bounds}{Conditional intervention bounds}.}

Concavity bounds the compensator by the average conditional mean.

\proofpart{predictable-compensator}

The conditional Bernoulli bounds give false-intervention and detection control.

\proofcode{LeBet/MeanIntervention}

\subsection{\paperheading{app:proof-pathwise-calibration}{Appendix}{Proof of Proposition~\getrefnumber{prop:pathwise-calibration}}}

Apply the log-normalizer bounds to the evidence process.

\proofpart{pathwise-implications}

Combine the linear and fixed-time boundaries, then take a finite union over training occurrences.

\proofpart{confidence-setup}

\proofcode{LeBet/ConditionalMeanConfidence}

\subsection{\paperheading{app:proof-statistical-guarantees}{Appendix}{Proof of Proposition~\getrefnumber{prop:statistical-guarantees}}}

\label{supp:solution-statistical}

Apply the conditional intervention bounds with the trivial conditioning sigma-algebra.

\subsection{\paperheading{app:proof-learned-intervention-guarantees}{Appendix}{Proof of Theorem~\getrefnumber{thm:learned-intervention-guarantees}}}

\label{supp:solution-learned}

\subsubsection{\paperlink{lem:conditional-error-identity}{Conditional Error}}

Extension--next swaps fix the training sample. Average the resulting conditional identities.

\proofpart{occurrence-identity}

\proofpart{loo-identity}

\proofpart{solution-conditional-error-identity}

\subsubsection{\paperlink{lem:stable-extension-error}{Extension Error under Stability}}

\paragraph{\paperlink{lem:stable-rank-domination}{Stability and Rank Domination}.}

\proofcode{LeBet/StableScores}

\proofcode{LeBet/RankTransfer}

\proofcode{LeBet/StratifiedLearning}

\proofpart{rank-moment-support}

\proofpart{average-rank-comparison}

Apply the finite-sampling tail to the diagonal LOO array.

\proofpart{occurrence-tail}

\proofpart{loo-tail}

\proofpart{solution-extension-tail}

\subsubsection{\paperlink{app:balanced-selected-thresholds}{Balanced Extensions and Selected Quantiles}}

Split the exceptional occurrences into training and extension parts.
The prefix maximum permits a data-dependent quantile.

\proofcode{LeBet/BalancedScores}

\proofpart{average-balanced-tail}

The conditional-error identity and truncated-moment transfer give the selected-quantile bound.

\proofpart{average-balanced-coverage}

\subsubsection{\paperlink{eq:training-conditional-generalization}{Training-Conditional Coverage}}

\label{supp:solution-coverage}

Rank domination and conditional Jensen give the positive-part bound.
Minimizing the objective gives training-conditional coverage.

\proofcode{LeBet/FiniteRankCoverage}

\proofpart{rank-closed-radius}

\proofpart{occurrence-coverage}

\proofpart{loo-coverage}

\proofpart{solution-error-coverage}

The integrated tail and exact single-extension law bound the rank objective.

\proofcode{LeBet/RankBounds}

\proofcode{LeBet/OneFutureScores}

\proofcode{LeBet/OneFutureLearning}

\subsubsection{\paperlink{app:proof-learned-conditional-means}{Intervention Guarantees}}

\label{supp:solution-intervention}

Transfer the training mean estimates through the quantile, then apply rank coverage.

\proofpart{estimated-coverage}

\proofpart{rank-estimation-transfer}

\proofcode{LeBet/MeanLearning}

Substitute the learned mean bounds into the conditional intervention inequalities.

\proofpart{loo-type-i}

\proofpart{solution-false-intervention}

\proofpart{loo-detection}

\proofpart{solution-detection}

\subsection{\paperheading{app:proof-rule-edit-structure}{Appendix}{Proof of Lemma~\getrefnumber{lem:rule-edit-structure}}}

Boolean identities give the rule-edit relations. Pointwise order propagates
through averages, extrema, and empirical quantiles.

\proofcode{LeBet/RuleStructure}

\end{document}